\documentclass[11pt]{article}
\usepackage[margin=1in]{geometry}
\usepackage[linesnumbered,ruled,vlined]{algorithm2e}
\usepackage{amsmath}
\usepackage{amssymb}
\usepackage{amsthm}
\usepackage{centernot}
\usepackage{epsfig}
\usepackage{fancyhdr}
\usepackage{fontawesome}
\usepackage{footmisc}
\usepackage{graphicx}
\usepackage{hyperref}
\usepackage{mathalfa}
\usepackage{mathtools}
\usepackage{multirow}
\usepackage{pifont}
\usepackage{pgfplots}
\pgfplotsset{width=10cm,compat=1.10}
\usepackage{tikz}
\usepackage{txfonts}
\usepackage{colortbl}
\usepackage{setspace}
\usepackage{subfig}
\usepackage{mathbbol}
\usepackage{natbib}
\setcitestyle{square}
\SetKwRepeat{Do}{do}{while}
\usepackage[toc]{appendix}
\usepackage{multicol}
\usepackage{accents}
\let\mb\mathbf
\let\mc\mathcal
\let\mt\mathtt
\let\mf\mathfrak
\let\bs\boldsymbol

\let\headcirc\mathring
\newcommand{\Prob}[0]{\mc{P}}
\newcommand{\G}[0]{\mc{G}}

\newcommand{\I}[0]{\mt{I}}
\newcommand{\E}[0]{\mt{E}}
\newcommand{\param}[0]{\mt{param}}

\newcommand{\D}[0]{\mc{D}}
\newcommand{\Pa}[0]{\mt{pa}}
\newcommand{\An}[0]{\mt{an}}
\newcommand{\Ch}[0]{\mt{ch}}
\newcommand{\De}[0]{\mt{de}}
\newcommand{\Nd}[0]{\mt{nd}}
\newcommand{\Ne}[0]{\mt{ne}}

\newcommand{\SK}[0]{\mt{sk}}
\newcommand{\MEC}[0]{\mathbb{MEC}}
\newcommand{\CMC}[0]{\mathbb{CMC}}
\newcommand{\CFC}[0]{\mathbb{CFC}}
\newcommand{\adjF}[0]{\mathbb{AdjF}}
\newcommand{\oriF}[0]{\mathbb{OriF}}
\newcommand{\resF}[0]{\mathbb{ResF}}
\newcommand{\Fr}[0]{\mathbb{Fr}}
\newcommand{\uFr}[0]{\mathbb{uFr}}
\newcommand{\SGS}[0]{\mathbb{SGSM}}
\newcommand{\Pm}[0]{\mathbb{PM}}
\newcommand{\uPm}[0]{\mathbb{uPM}}
\newcommand{\triF}[0]{\mathbb{TriF}}
\newcommand{\ParamM}[0]{\mathbb{ParamM}}
\newcommand{\uParamM}[0]{\mathbb{uParamM}}

\newcommand{\DAG}[0]{\mathbb{DAG}}
\newcommand{\la}[0]{\langle}
\newcommand{\ra}[0]{\rangle}
\newcommand{\CI}[0]{\perp\!\!\!\perp}

\newcommand{\ot}[0]{\leftarrow}

\newcommand{\ib}[0]{\item[$\bullet$]}
\newcommand{\cellblk}[0]{\cellcolor{black}}
\newcommand{\cellcyan}[0]{\cellcolor{cyan}}
\newcommand{\cellmag}[0]{\cellcolor{magenta}}
\newcommand{\cellgray}[0]{\cellcolor{lightgray}}

\newcommand{\cupdot}{\mathbin{\mathaccent\cdot\cup}}
\newtheorem{definition}{Definition}[section]
\newtheorem{corollary}[definition]{Corollary}
\newtheorem{lemma}[definition]{Lemma}
\newtheorem{example}[definition]{Example}
\newtheorem{theorem}[definition] {Theorem}

\newtheorem{fact}[definition]{Fact}
\newenvironment{pf}{\noindent \textit{Proof.\hspace{0.1cm}}}{\hfill$\square$}
\begin{document}
\fancypagestyle{plain}{
\fancyhead{}
\fancyhead[L]{\textbf{Causal Razors}}
\fancyhead[R]{\textbf{Wayne Lam} \href{mailto:wylam055@gmail.com}{\faEnvelopeO}}}
\renewcommand{\footrulewidth}{1pt}
%%%%%%%%%%%%%%%%%%%%%%%%%%%%%%%%%%%%%%%%%%%%%%%%%%%%%%%%%%%%%%%%%%%%%%

\title{Causal Razors: A Comparative Study of Simplicity Assumptions in Causal Discovery\footnote{This paper is part of the author's dissertation titled \textit{Causal Razors and Causal Search Algorithms} \citep{MyPhD}.}}
\author{{Wai-yin (Wayne) Lam} \\
  \multicolumn{1}{p{.7\textwidth}}{\centering\small\emph{Department of Philosophy\\ Carnegie Mellon University\\ Pittsburgh, Pennsylvania, USA}}}
% \author{Wai-yin (Wayne) Lam}
% \affil{Department of Philosophy\\ Carnegie Mellon University\\ Pittsburgh, Pennsylvania, USA}
\date{}
\maketitle

%%%%%%%%%%%%%%%%%%%%%%%%%%%%%%%%%%%%%%%%%%%%%%%%%%%%%%%%%%%%%%%%%%%%%%
%%% Abstract %%%%%%%%%%%%%%%%%%%%%%%%%%%%%%%%%%%%%%%%%%%%%%%%%%%%%%%%%
%%%%%%%%%%%%%%%%%%%%%%%%%%%%%%%%%%%%%%%%%%%%%%%%%%%%%%%%%%%%%%%%%%%%%%
\begin{abstract}
\noindent When performing causal discovery, assumptions have to be made on how the true causal mechanism corresponds to the underlying joint probability distribution. These assumptions are labeled as \textit{causal razors} in this work. We review numerous causal razors appeared in the literature, and offer a comprehensive logical comparison over them. In particular, we scrutinize an unpopular causal razor, namely \textit{parameter minimality}, in multinomial causal models and its logical relations with other well-studied causal razors. Our logical result poses a dilemma in selecting a reasonable scoring criterion for score-based casual search algorithms.
\end{abstract}

\fancyhead[L]{\,} 
%%%%%%%%%%%%%%%%%%%%%%%%%%%%%%%%%%%%%%%%%%%%%%%%%%%%%%%%%%%%%%%%%%%%%%
%%%%%%%%%%%%%%%%%%%%%%%%%%%%%%%%%%%%%%%%%%%%%%%%%%%%%%%%%%%%%%%%%%%%%%
%%%%%%%%%%%%%%%%%%%%%%%%%%%%%%%%%%%%%%%%%%%%%%%%%%%%%%%%%%%%%%%%%%%%%%
\section{Introduction}
\label{sec:intro}
%%%%%%%%%%%%%%%%%%%%%%%%%%%%%%%%%%%%%%%%%%%%%%%%%%%%%%%%%%%%%%%%%%%%%%
%%%%%%%%%%%%%%%%%%%%%%%%%%%%%%%%%%%%%%%%%%%%%%%%%%%%%%%%%%%%%%%%%%%%%%
%%%%%%%%%%%%%%%%%%%%%%%%%%%%%%%%%%%%%%%%%%%%%%%%%%%%%%%%%%%%%%%%%%%%%%
A fundamental goal of scientific inference is to discover causal structures. How to attain this goal based on non-experimental data is a striking question, especially in the era of abundant data. In recent decades, the research of \textit{causal discovery} has been flourishing, particularly due to the explosion of data and the rise of many machine learning techniques that ease computational complexity. While traditional statistical tools are designed to recover non-causal associations, causal discovery aims at the process of inferring causal relationships between variables in a given system from observational data. 

Causal discovery procedures, also known as \textit{causal search algorithms}, generally connect statistical information from the observational data with a graphical representation that depicts causal relations. One popularly used graphical device is causal directed acyclic graphs (DAGs) where each directed edge depicts the direct causal influence flowing from one variable to another. By postulating that the true causal mechanism can be represented by a causal DAG, causal search algorithms primarily purport to recover the causal information from the observational data generated by the ground truth.

Given an observational dataset over a set of measured variables $\mb{V}$ with a joint probability distribution $\Prob$, each DAG $\G$ over $\mb{V}$ is a \textit{hypothesis} with an objective to \textit{explain} $\Prob$ in terms of its graphical features, particularly through \textit{d-separations} which was famously introduced by \cite{Pearl_10.5555/534975}. The explanation of $\Prob$ is predominately referring to the set of \textit{conditional independencies} (CIs) entailed by $\Prob$, which is a focal point for researchers to mine possible causal information. Accordingly, one intuitive task in causal discovery is to shrink the space of hypotheses to an extent that substantial causal information of the true causal mechanism can be retrieved. Nonetheless, the size of the hypothesis space grows exponentially with the size of $\mb{V}$. So, a crucial research question concerns how \textit{poorer} hypotheses can be eliminated from consideration. 

Structure discovery can hardly be feasible without making any \textit{assumptions}. Causal discovery makes no exception. The elimination of poorer hypotheses can be performed only by assuming that the true causal mechanism, denoted as $\G^*$, satisfies certain desired features such that DAGs without these features are deemed inferior. As inspired by the \textit{principle of common cause} from \citet{Reichenbach}, one well-known assumption is the \textit{Causal Markov Condition} (CMC) which requires that the CIs entailed through d-separations by $\G^*$ is a subset of the CIs held in the joint probability distribution $\Prob$. Along these lines, DAGs violating CMC (i.e., \textit{non-Markovian} DAGs) are impermissible hypotheses and should be rejected.

Despite its broad acceptance, CMC alone still leaves us with an ample amount of hypotheses to consider. For instance, every complete DAG (i.e., all vertices are pairwise connected by a directed edge) trivially satisfies CMC in that it entails no CI by d-separation. Stronger assumptions, thus, are required to retrieve more substantial information about $\G^*$. One widely discussed candidate in the literature is the \textit{Causal Faithfulness Condition} (CFC), or \textit{faithfulness} for short. By adopting the converse of CMC, faithfulness assumes that the set of CIs held in $\Prob$ is a subset of those entailed graphically by $\G^*$. Albeit its strong logical strength which enables a considerable cut-down of the hypothesis space, $\G^*$ might not be identifiable even if CFC is satisfied. DAGs different in orientations can graphically entail the same set of CIs. In standard notations, these DAGs are in the same \textit{Markov equivalence class} (MEC). One notable theoretical result is that the MEC of the $\G^*$ (i.e., the true MEC) is identifiable when the true \textit{causal model} $(\G^*, \Prob)$ meets CFC.

On a positive note, the violation of CFC, as shown by \cite{spirtes2000causation}, has a Lebesgue measure of zero at the level of population for a wide range of probability distributions that are commonly studied. Nevertheless, unfaithfulness does not vanish in small probability in practice. Sometimes scientific models have paramter values evolved through a long-term dynamic equilibrium and tended to elicit unfaithfulness \citep{Andersen_CFC}. In the case of finite samples, violation of CFC can be \textit{almost} (in contrast to \textit{exact}) in the sense that dependencies are mistakenly judged as independencies (i.e., almost independencies) by statistical tests \citep{zhang2008detection}. To avoid, so, one might be tempted to adopt the \textit{strong causal faithfulness condition} which assumes that no CI not entailed by CMC almost holds in the joint distribution \citep{Kalisch_strong_CFC}. However, \citep{uhler2013geometry} convincingly show that almost unfaithfulness occurs frequently. This observation prompts researchers to reflect on the legitimacy of assuming CFC at the sample level despite its nice mathematical property possessed at the population level.

Below is an example illustrating this phenomenon. Suppose we want to analyze the causal relations over the following five variables: \textit{taking contraception pills} ($C$), \textit{diabetes} ($D$), \textit{smoking} ($K$), \textit{pregnancy} ($P$), and \textit{stroke} ($T$). Say $\G^*$ in Figure \ref{fig:intro_ex} depicts the true data-generating causal mechanism where we collect finite samples from. We observe that $T$ is directly caused by each of the other four variables. For example, there is a positive causal influence from $C$ to $T$ due to a slight increase in probability of having stroke after taking contraceptive pills. On the other hand, there is also an indirect but negative causal influence from $C$ to $T$ mediated by $P$. This can be interpreted as how the probability of pregnancy is lowered by taking contraceptive pills while pregnancy raises the probability of stroke. Suppose that the direct (and positive) and indirect (and negative) causal influences from $C$ to $T$ are relatively close to even out, that is, $C$ makes a negligible change to the probability of $T$ though they are probabilistically dependent. Due to sampling error, our finite sample informed us the marginal independence between $C$ and $T$ mistakenly. The almost independence between $C$ and $T$ is not graphically represented by $\G^*$ (due to the adjacency between them). Thus, $\G^*$ is not faithful to the sampling distribution and it exhibits an almost violation of CFC.

\begin{figure}[h]
    \centering
    \begin{tikzpicture}
    \node (X1) at (0.0,1.0) {$C$};
    \node (X2) at (0.9510565162951535,0.30901699437494745) {$D$};
    \node (X3) at (0.5877852522924732,-0.8090169943749473) {$K$};
    \node (X4) at (-0.587785252292473,-0.8090169943749476) {$P$};
    \node (X5) at (-0.9510565162951536,0.30901699437494723) {$T$};
    \node (label) at (2, -0.4) {$\G^*$};
    \path [->,line width=0.4mm] (X1) edge (X5);
    \path [->,line width=0.4mm] (X2) edge (X5);
    \path [->,line width=0.4mm] (X3) edge (X5);
    \path [->,line width=0.4mm] (X4) edge (X5);
    \path [->,line width=0.4mm] (X1) edge (X4);
    \end{tikzpicture}
    \caption{A DAG hypothesis explaining the causal relations over \textit{taking contraception pills} ($C$), \textit{diabetes} ($D$), \textit{smoking} ($K$), \textit{pregnancy} ($P$), and \textit{stroke} ($T$).}
    \label{fig:intro_ex}
\end{figure}

Historically, the \textit{PC} algorithm \citep{PC})  and the \textit{GES} algorithm \citep{chickering2002optimal} have been proven to identify the true MEC under CFC. But they can fail this task under unfaithfulness. For instance, the PC algorithm will mistakenly judge $C$ and $T$ as non-adjacent due to their almost independence. Thus, the problem of almost unfaithfulness motivates the research direction of studying assumptions strictly weaker than CFC while the true MEC is still identifiable. These include the restricted form of faithfulness by \citet{ramsey2006}, also the \textit{Sparsest Markov Representation} (SMR) assuption introduced by \citet{raskutti2018learning}. Numerous theoretical assumptions of a similar sort have been scrutinized in the literature. For instance, \cite{zhang2013comparison} discussed two minimality conditions, namely \textit{Pearl-minimality} and \textit{SGS-minimality}, that the true causal model is expected to satisfy. With CFC included, he referred to these three conditions as \textit{Occam's razors} for Markovian causal models. Obviously, this notation dates back to William of Occam's law of parsimony which gives precedence to \textit{simplicity}: of the competing hypotheses, the simpler one is to be preferred. Given the many simplicity assumptions pertinent to causal discovery have been surveyed in recent decades, we borrow a similar terminology to subsume all of these assumptions under the generic term \textit{causal razors}.

More than a dozen causal razors have been proposed But a comprehensive logical comparison of them is still found wanting. One possible exception is \citep{forster2020frugal} which compares a handful of causal razors with \textit{frugality}: a weaker variant of SMR that requires that Markovian DAGs which are not the \textit{sparsest} should be rejected. For example, consider the DAG hypothesis $\G'$ in Figure \ref{fig:intro_ex_6_edge} which is supposedly also Markovian like $\G^*$ in Figure \ref{fig:intro_ex}. Following frugality, $\G'$ should be rejected because $\G^*$ contains strictly fewer edges than $\G'$.

%%%%%%%%%%%%%%%%%%%%
\begin{figure}[h]
    \centering
    \begin{tikzpicture}
    \node (X1) at (0.0,1.0) {$C$};
    \node (X2) at (0.9510565162951535,0.30901699437494745) {$D$};
    \node (X3) at (0.5877852522924732,-0.8090169943749473) {$K$};
    \node (X4) at (-0.587785252292473,-0.8090169943749476) {$P$};
    \node (X5) at (-0.9510565162951536,0.30901699437494723) {$T$};
    \node (label) at (2, -0.4) {$\G'$};
    \path [->,line width=0.4mm] (X1) edge (X4);
    \path [->,line width=0.4mm] (X2) edge (X4);
    \path [->,line width=0.4mm] (X3) edge (X4);
    \path [->,line width=0.4mm] (X5) edge (X4);
    \path [->,line width=0.4mm] (X2) edge (X5);
    \path [->,line width=0.4mm] (X3) edge (X5);
    \end{tikzpicture}
    \caption{An alternative DAG hypothesis which satisfies CMC.}
    \label{fig:intro_ex_6_edge}
\end{figure}
%%%%%%%%%%%%%%%%%%%%

Ostensibly, frugality has an intuitive appeal in terms of simplicity. Analogous to counting edges in DAGs, experimental research in psychology (e.g., \citet{Children_node_simplicity}) find positive evidence that people identify simplicity among causal hypotheses as the number of causes invoked in a causal explanation. Also, certain causal search algorithms (e.g., \citet{raskutti2018learning}, \citet{solus2021consistency}) purport to identify the true MEC by favoring sparser Markovian DAGs. 

A theoretical motivation for embracing causal razors defined by graphical sparsity is due to the commonly studied \textit{linear Gaussian} causal models. They license an elegant bijection between directed edges in a DAG and linear coefficients that are \textit{parameters} used to define a causal model. Thus, the intuitive appeal of frugality is arguably grounded by a more fundamental concept: \textit{number of parameters} in a causal model. 

Counting parameters is a ubiquitous practice in both statistics and computer science. It helps to measure the statistical complexity of a statistical/learning model, and hypotheses with fewer parameters are less prone to the problem of \textit{overfitting} in general \citep{MacKay_1992}. However, no serious attempt has been devoted to the analysis of \textit{parameter minimality} in the causal discovery literature, not to mention its relation to other causal razors such as frugality. 

The points made above incentivize a more encyclopedic analysis of causal razors. By laying out a spectrum of causal razors defined inclusively by CMC and CFC, they will be scrutinized in terms of their logical strength. On the other hand, we demonstrate that a comprehensive logical analysis of causal razors, though can easily be glossed over, yields salient implications to complement the increasingly influential development of causal search algorithms. One instance is the recent result of \citep{GRaSP} which shows that CFC is logically equivalent to a specific form of Pearl-minimality. This logical discovery can be utilized to show the limitation of a causal search algorithm that was not aware of by its developers. For instance, \citet{solus2021consistency} thought that their \textit{GSP algorithm} can be correct under unfaithfulness. Yet, when CFC fails, the logical equivalence demonstrates how their algorithm can be trapped by a Pearl-minimal DAG which is deemed as sub-optimal in their framework. This is an instance verifying the practical significance of a careful comparative study of causal razors.

On the other hand, we focus on the comparison between frugality and parameter minimality in \textit{multinomial} causal models. An example will be showcased to prove the logical independence between the two causal razors, that is, a sparser DAG (e.g., $\G^*$ in Figure \ref{fig:intro_ex}) can have more parameters than a denser one (e.g., $\G'$ in Figure \ref{fig:intro_ex_6_edge}). Though $\G'$ appears to be implausible when we have background knowledge over the variables, it is not evident whether $\G^*$ or $\G'$ is a more preferable hypothesis when no background or contextual information is given. This logical independence, furthermore, is closely intertwined with actual implementations of causal search algorithms, particularly the \textit{scored-based} species such as GES and GSP. In the face of almost unfaithfulness, should one adopt \textit{edge count} as a scoring criterion that prefers sparser Markovian DAGs, or a metric (e.g., \textit{Bayesian scoring criterion} by \citet{schwarz1978estimating}) that favors those with fewer parameters? This dilemma evokes algorithm users to reflect on which causal razor they have implicitly committed to when performing robust causal discovery.

Additional remarks about causal razors should be made explicit before delving into further details. First, though the term `razor' encodes the spirit of Occam's razor, not every causal razor to be studied conveys a clear bearing of simplicity. For instance, \textit{orientation faithfulness} \citep{ramsey2006} requires that triplets of vertices in a DAG are not associated with a particular kind of CIs. It is not obvious how this assumption can be affiliated with a sense of theoretical parsimony. Thus, causal razors in this work primarily refer to the behavior of shrinking the hypothesis space. 

Second, the shrinking behavior should be understood \textit{functionally}. Following the functional account of causation espoused by \citet{woodward2021causation}, causal reasoning focuses on what agents \textit{want to do} with the causal notions with certain goals and purposes. So the shrinking behavior of causal razors needs to be studied with a well-defined \textit{desideratum} that informs us how much shrinking is necessary and sufficient. Following the common practice, this work will concentrate on the identification of the true MEC. That being said, different formal desiderata prevail in the literature. The \textit{LiNGAM algorithm} \citep{shimizu2006linear} does not seek to identify the true MEC, and the \textit{conservative PC algorithm} \citep{ramsey2006} aims to identify a broader class of DAGs than the true MEC. Following the learning-theoretic approach as in \citep{LinZhang2020}, one should inquire what the highest achievable desideratum is when assuming a particular causal razor. The study of desiderata can be interpreted in a practical sense as well. An algorithm assuming a strong causal razor can run much faster than its competitor which assumes weaker. When there is a lack of computational resources, researchers may justifiably choose to sacrifice accuracy for efficiency by embracing the latter as a desideratum. Instead of insisting on an objective study of desiderata, this work leans to open up a menu of possibilities that researchers are free to choose from.

Lastly, \textit{parametric assumptions} can also be used to shrink the space of possible hypotheses, and should be qualified as causal razors accordingly. Nevertheless, parametric assumptions are generally disconnected from \textit{structural} causal razors such as CFC and frugality that are defined non-parametrically. Still, parameter minimality is a notable exception which demands a parametric assumption to effectively enumerate parameters. In this work, the discussion of parametric assumptions will only be studied in association with parameter minimality. But interested readers are suggested to explore the interplay between parametric assumptions and structural causal razors.

This paper is organized as follows. \textbf{Section \ref{sec:basic}} begins with some basic terminologies. In \textbf{Section \ref{sec:common_rz}}, we review a list of eleven structural causal razors studied in the literature, followed by a comprehensive logical analysis of them in \textbf{Section \ref{sec:rz_hier}}. \textbf{Section \ref{sec:paramM}} discusses parameter minimality and its stronger variant, also their logical relations to the structural causal razors in the multinomial context. \textbf{Section \ref{sec:alg_imp}} initiates an algorithmic dilemma resulted from our logical analysis. Concluding remarks will be given in \textbf{Section \ref{sec:discuss}}. Proofs and construction details of examples will be left in the \textbf{Appendix}.

%%%%%%%%%%%%%%%%%%%%%%%%%%%%%%%%%%%%%%%%%%%%%%%%%%%%%%%%%%%%%%%%%%%%%%
%%%%%%%%%%%%%%%%%%%%%%%%%%%%%%%%%%%%%%%%%%%%%%%%%%%%%%%%%%%%%%%%%%%%%%
%%%%%%%%%%%%%%%%%%%%%%%%%%%%%%%%%%%%%%%%%%%%%%%%%%%%%%%%%%%%%%%%%%%%%%
\section{Basic Terminologies}
\label{sec:basic}
%%%%%%%%%%%%%%%%%%%%%%%%%%%%%%%%%%%%%%%%%%%%%%%%%%%%%%%%%%%%%%%%%%%%%%
%%%%%%%%%%%%%%%%%%%%%%%%%%%%%%%%%%%%%%%%%%%%%%%%%%%%%%%%%%%%%%%%%%%%%%
%%%%%%%%%%%%%%%%%%%%%%%%%%%%%%%%%%%%%%%%%%%%%%%%%%%%%%%%%%%%%%%%%%%%%%
We use italicized letters for variables (e.g., $X_1, Y$) and boldfaced letters for sets of variables (e.g., $\mb{X}$) throughout this paper. In the following, we review some standard notations and definitions related to directed acyclic graphs and joint probability distributions. 

A \textit{directed graph} $\mc{G}$ over a set of measured variables $\mb{V} = \{X_1,..., X_m\}$ consists of $m$ vertices $\mb{v} = \{1,...,m\}$ where each vertex $i \in \mb{v}$ associates to the variable $X_i \in \mb{V}$, and each edge in $\G$ is directed with the form $j \to k$. We focus on simple graphs where no vertex has an edge to itself, and there is at most one edge for every pair of vertices. For any vertices $j, k \in \mb{v}$, we say that $j$ and $k$ are \textit{adjacent} in $\G$ if there exists a directed edge between $j$ and $k$ (i.e., $j \to k$ or $k \to j$) in $\G$. Let $\E(\G)$ be the set of edges in $\G$.

Given a directed graph $\G$, a \textit{path} $\mb{p}$ is a sequence of vertices $\la \mb{p}_1, ..., \mb{p}_k\ra$ for some $k \geq 2$ where $\mb{p}_i$ and $\mb{p}_{i+1}$ are adjacent in $\G$ for each $1 \leq i < k$. Such a path $\mb{p}$ is said to be a \textit{directed path} (from $\mb{p}_1$ to $\mb{p}_k$) in $\G$ if $(\mb{p}_{i} \to \mb{p}_{i+1}) \in \E(\G)$ holds for each $1 \leq i < k$. A directed acyclic graph (DAG) is a directed graph where no vertex has a directed path to itself.

Given a set of variables $\mb{V}$, denote $\DAG(\mb{v})$ as the set of all possible DAGs defined over $\mb{V}$. Below are some standard graphical notations. Given a DAG $\G \in \DAG(\mb{v})$ and a vertex $i \in \mb{v}$:
\begin{itemize}
    \ib $\Pa(i, \G) := \{j \in \mb{v}: (j \to i) \in \E(\G)\}$ : the set of \textit{parents} of $i$ in $\G$;
    \ib $\Ch(i, \G) := \{j \in \mb{v}: (i \to j) \in \E(\G)\}$ : the set of \textit{children} of $i$ in $\G$;
    \ib $\Ne(i, \G) := \Pa(i, \G) \cup \Ch(i, \G)$ : the set of \textit{neighbors} of $i$ in $\G$;
    \ib $\An(i, \G) := $ transitive closure of $\Pa(i, \G)$ union with $\{i\}$: the set of \textit{ancestors} of $i$ in $\G$; 
    \ib $\De(i, \G) := $ transitive closure of $\Ch(i, \G)$ union with $\{i\}$ : the set of \textit{descendants} of $i$ in $\G$;
    \ib $\Nd(i, \G) := \mb{v} \setminus \De(i, \G)$ : the set of \textit{non-descendants} of $i$ in $\G$.   
\end{itemize}
The \textit{skeleton} of $\G$, denoted as $\SK(\G)$, is the set of (unordered pairs of) adjacencies in $\G$. Pictorially, when $\G$ is a DAG, $\SK(\G)$ represents the \textit{undirected} graph of $\G$ obtained by removing all the arrowheads in $\G$.

For any triple of pairwise distinct vertices $i, j, k \in \mb{v}$, we say that $(i, j, k)$ is \textit{unshielded} if $(i, j), (j, k) \in \SK(\G)$ but $(i, k) \notin \SK(\G)$, and $(i, j, k)$ is shielded (or a \textit{triangle}) if they are pairwise adjacent in $\G$. Given a path $\mb{p} = \la \mb{p}_1,...,\mb{p}_k\ra$ in a DAG $\G$, $\mb{p}_i$ (for any $1 < i < k$) is a \textit{collider} on $\mb{p}$ if $(\mb{p}_{i-1} \to \mb{p}_{i}), (\mb{p}_{i+1} \to \mb{p}_i) \in \E(\G)$, and a \textit{non-collider} otherwise.

For any $i, j \in \mb{v}$ and any $\mb{k} \subseteq \mb{v} \setminus \{i, j\}$, $i$ and $j$ are \textit{d-connected} given $\mb{k}$ in $\G$ if there exists a path $\mb{p}$ between $i$ and $j$ in $\G$ such that no non-collider on $\mb{p}$ is in $\mb{k}$, and each collider $l$ on $\mb{p}$ or a descendant of $l$ is in $\mb{k}$. $i$ and $j$ are \textit{d-separated} given $\mb{k}$ in $\G$ if $i$ and $j$ are not d-connected given $\mb{k}$. For any disjoint subsets of vertices $\mb{i}, \mb{j}, \mb{k} \subseteq \mb{v}$, $\mb{i}$ and $\mb{j}$ are d-separated given $\mb{k}$ in $\G$, written as  $\mb{i} \perp_\G \mb{j}\,|\,\mb{k}$, if $i$ and $j$ are d-separated by $\mb{k}$ in $\G$ for every $i \in \mb{i}$ and every $j \in \mb{j}$. 

On the other hand, consider a joint probability distribution $\Prob$ over $\mb{V}$. For any subsets $\mb{X}, \mb{Y} \subseteq \mb{V}$, the \textit{conditional probability distribution} of $\mb{X}$ given $\mb{Y}$ is denoted by $\Prob(\mb{X}\mid \mb{Y})$. For any pairwise disjoint subsets of variables $\mb{X}, \mb{Y}, \mb{Z} \subseteq \mb{V}$, $\mb{X}$ and $\mb{Y}$ are said to be \textit{conditionally independent} given $\mb{Z}$ if $\Prob(\mb{X}\mid \mb{Y}, \mb{Z}) = \Prob(\mb{X} \mid \mb{Z})$, and we write $\mb{X} \CI_\Prob \mb{Y}\,|\,\mb{Z}$ to denote the conditional independence (CI) between $\mb{X}$ and $\mb{Y}$ given $\mb{Z}$ in $\Prob$. When $\mb{X} = \{X\}$ and $\mb{Y} = \{Y\}$ are singleton sets, we write $X \CI_\Prob Y\mid \mb{Z}$ (instead of $\{X\} \CI_\Prob \{Y\}\mid \mb{Z}$). When $\mb{Z} = \varnothing$, we write $\mb{X}\CI_\Prob \mb{Y}$ (instead of $\mb{X}\CI_\Prob \mb{Y}\mid\mb{Z}$).\footnote{$\mb{X}\CI_\Prob \mb{Y}$ is generally referred to as a \textit{marginal independence} in the literature. But, for the sake of convenience, all marginal independencies are labeled as CIs.} 

Following \citep{Studeny_book}, we use \textit{independence models} as machinery to juxtapose the CI relations that hold in a joint distribution with those that entailed by a DAG through d-separation. First, we define the set of all possible CIs as: 
\begin{align*}
    \I(\mb{V}) := \{\la \mb{X}, \mb{Y}\,|\,\mb{Z}\ra: \mb{X}, \mb{Y}, \mb{Z} \text{ are pairwise disjoint subsets of } \mb{V}\}.
\end{align*}
The independence model of a joint distribution $\Prob$ (over $\mb{V}$) is
\begin{align*}
    \I(\Prob) = \{\la \mb{X}, \mb{Y}\,|\,\mb{Z}\ra \in \I(\mb{V}): \mb{X} \CI_\Prob \mb{Y}\,|\,\mb{Z}\}
\end{align*}
and the independence model of a DAG $\G$ (over $\mb{V}$) is
\begin{align*}
    \I(\G) = \{\la\mb{X}_\mb{j}, \mb{X}_\mb{k}\,|\,\mb{X}_\mb{l}\ra  \in \I(\mb{V}): \mb{j} \perp_\G \mb{k}\,|\,\mb{l}\}
\end{align*}
where $\mb{X}_\mb{i} = \{X_j\}_{j\,\in\,\mb{i}}$ for every $\mb{i} \subseteq \mb{v}$. Similar to the shorthands introduced above, we write $\la X, Y \mid \mb{Z}\ra$ when $\mb{X}$ and $\mb{Y}$ are singleton sets $\{X\}$ and $\{Y\}$, and write $\la \mb{X}, \mb{Y}\ra$ when $\mb{Z} = \varnothing$.

Two DAGs $\G_1$ and $\G_2$ are \textit{Markov equivalent} if $\I(\G_1) = \I(\G_2)$. The \textit{Markov equivalent class} of a DAG $\G$, denoted $\MEC(\G)$, is defined by the set of DAGs that are Markov equivalent to $\G$. Two DAGs can be different but Markov equivalent. One simple case is the 2-vertex case where $\E(\G_1) = \{1 \to 2\}$ and $\E(\G_2) = \{2 \to 1\}$ while $\I(\G_1) = \I(\G_2) = \varnothing$. As proven by \citet{verma1988causal}, two DAGs are Markov equivalent if and only if they share the same set of adjacencies and unshielded colliders. Each MEC can be graphically represented by a unique \textit{completed partially directed acyclic graph} (CPDAG) which is a mixed graph composed of directed and undirected edges. \citep{Meek1995} showed that a set of orientation rules is complete to obtain a unique CPDAG from the skeleton and the unshielded colliders of a DAG.

% for a more detailed characterization of CPDAGs and the procedures of converting a DAG into its unique CPDAG. By relabeling the variables of $\G^*$ in Figure \ref{fig:intro_ex} in an alphabetical order, Figure \ref{fig:CPDAG} shows the CPDAG of $\G^*$ where the adjacency between 1 and 4 is an undirected edge.

% \begin{figure}[h]
%     \centering
%     \begin{tikzpicture}
%     \node (X1) at (0.0,1.0) {$1$};
%     \node (X2) at (0.9510565162951535,0.30901699437494745) {$2$};
%     \node (X3) at (0.5877852522924732,-0.8090169943749473) {$3$};
%     \node (X4) at (-0.587785252292473,-0.8090169943749476) {$4$};
%     \node (X5) at (-0.9510565162951536,0.30901699437494723) {$5$};
%     \path [->,line width=0.4mm] (X1) edge (X5);
%     \path [->,line width=0.4mm] (X2) edge (X5);
%     \path [->,line width=0.4mm] (X3) edge (X5);
%     \path [->,line width=0.4mm] (X4) edge (X5);
%     \path [line width=0.4mm] (X1) edge (X4);
%     \end{tikzpicture}
%     \caption{CPDAG of $\G^*$ in Figure \ref{fig:intro_ex} after relabeling the variables by an alphabetical order.}
%     \label{fig:CPDAG}
% \end{figure}

A \textit{causal model} is a pair $(\G, \Prob)$ where $\G$ is a DAG and $\Prob$ is a joint probability distribution over the same set of variables $\mb{V}$. We denote $\G^*$ as the \textit{true} data-generating DAG and so $(\G^*, \Prob)$ is the true causal model assumed to always exist. 

Next, we use $\param(\G, \Prob)$ to define the set of \textit{parameters} of a causal model $(\G, \Prob)$. A \textit{parametric assumption} on $\Prob$ has to be made in order to identify the set of parameters of a causal model. So, for a \textit{non-parametric} model where we do not assume that the joint distribution $\Prob$ belongs to any particular parametric family of distributions, $\param(\G, \Prob)$ becomes an undefined notion. In the following, we write $\param(\G)$ for short when the underlying joint distribution $\Prob$ is clear from context. We will focus on two kinds of causal models in this work: \textit{structural equation models} (SEMs) and \textit{multinomial causal models}.

A structural equation model $(\G, \bs{\epsilon}, \mc{F})$ over $\mb{V} = \{X_1,..., X_m\}$ where $\bs{\epsilon} = \{\epsilon_1,...,\epsilon_m\}$ is a set of \textit{noise terms}, $\mc{F} = \{f_1,..., f_m\}$ is a set of \textit{structural equations} where the following is satisfied for each $1 \leq i \leq m$.
\begin{align}
    X_i = f_i(\Pa(i, \G), \epsilon_i)
\end{align}
Let $\Prob_{\bs\epsilon}$ be the joint probability distribution over the noise terms. Given the structural equations in $\mc{F}$, a joint distribution $\Prob$ over $\mb{V}$ is induced provided that $X_i$'s can be uniquely solved in terms of $\epsilon_i$'s. One commonly studied species of SEMs in the literature is the \textit{linear Gaussian causal model} where $\bs\epsilon$ follows a multivariate Gaussian distribution, and each $f_i \in \mc{F}$ is a linear equation. To be precise, each variable $X_i \in \mb{V}$ is governed by the following equation:
\begin{align}
\label{linear_Gaussian_param_eq}
    X_i = \Bigg(\sum_{X_j\,\in\,\mb{X}_{\Pa(i, \G)}} \theta_{ji}\,X_j\Bigg) + \epsilon_i  
\end{align}
\noindent where $0 \neq \theta_{ji} \in \bs\theta_\G$ only if $j \in \Pa(i, \G)$, and each $\epsilon_i \in \bs\epsilon$ follows a Gaussian distribution. One widely made assumption is that noise terms in $\bs\epsilon$ are assumed to jointly independent.\footnote{By requiring joint independence of $\bs\epsilon$, the mean and variance of each $\epsilon_i \in \bs\epsilon$ can be estimated by the conditional mean and partial variance of $X_i$ given its parents $\mb{X}_{\Pa(i, \G)}$.} Under this assumption, we say that $\param(\G) = \bs\theta_\G$ (without involving $\bs\epsilon$). Given the bijection between $\bs\theta_\G$ and $\E(\G)$, we have a nice equality $|\param(\G)| = |\E(\G)|$ for every linear causal Gaussian model $(\G, \Prob)$.

On the other hand, in a multinomial causal model $(\G, \Prob)$, 
$\Prob$ is a joint multinomial distribution where each $X_i \in \mb{V}$ is a discrete $k$-valued variable with its range of values specified by \textit{range}$(X_i) = \{0,1,...,k-1\}$ where $k \geq 2$. Let $\mathtt{r}(i)$ be the \textit{cardinality} of $X_i$'s range of values (i.e., $\mt{r}(i) := |\textit{range}(X_i)|$). So, we say that $\mb{V}$ is \textit{ranged from} $\la \mt{r}(1), \mt{r}(2),..., \mt{r}(|\mb{v})|\ra$. We further define $range(\mb{X}_\mb{i})$ for any subset of variables $\mb{X_i} \subseteq \mb{V}$ as the Cartesian product of $\{range(X_j)\}_{j \in \mb{i}}$. Similarly, we extend $\mb{r}(\cdot)$ to any subset of variables $\mb{X}_\mb{i} \subseteq \mb{V}$ as follows: 
\begin{align*}
    \mt{r}(\mb{i}) := 
    \begin{cases}
    \prod_{i \in \mb{i}} \mt{r}(i) & \text{if } \mb{i} \neq \varnothing\\
    1 &  \text{otherwise.}
    \end{cases}
\end{align*}
For each $i \in \mb{v}$, let $\bs\theta_{i, \G}$ be the set of parameters of $i$ in $\G$ needed to compute $\Prob(X_i\,|\,\mb{X}_{\Pa(i, \G)})$. Each parameter in $\bs\theta_{i, \G}$ corresponds to the probability of $X_i$ taking a particular value given a specific value configuration $\mb{s} \in \textit{range}(\mb{X}_{\Pa(i, \G)})$. Given that $\sum_{j \in range(X_i)} \Prob(X_i = j \mid \mb{X}_{\Pa(i, \G)} = \mb{s}) = 1$ for each $\mb{s} \in \textit{range}(\mb{X}_{\Pa(i, \G)})$, the \textit{degree of freedom} (i.e., the maximum number of logically independent values) of computing $\Prob(X_i\,|\,\mb{X}_{\Pa(i, \G)} = \mb{s})$ is $\mt{r}(i)-1$.\footnote{Since $\sum_{j \in range(X_j)} \Prob(X_i = j \mid \mb{X}_{\Pa(i, \G)} = \mb{s}) = 1$, $\Prob(X_i = k \mid \mb{X}_{\Pa(i, \G)} = \mb{s})$ is determined when $\Prob(X_i = j \mid \mb{X}_{\Pa(i, \G)} = \mb{s})$ for every $j \in range(X_i)\setminus \{k\}$ is known.} Accordingly, we have 
\begin{align}
\label{eq:BN_param_eq_vertex}
|\bs\theta_{i, \G}| := (\mt{r}(i) - 1) \times \mt{r}(\Pa(i, \G)).
\end{align}
Consequently, the set of parameters of a multinomial causal model $(\G, \Prob)$ is $\param(\G) = \cup_{i\in \mb{v}}\,\bs\theta_{i, \G} = \bs{\theta}_\G$ such that
\begin{align}
\label{eq:BN_param_eq_DAG}
    |\param(\G)| = \sum_{i \in \mb{v}} |\bs\theta_{i, \G}| = \sum_{i \in \mb{v}} (\mt{r}(i) - 1) \times \mt{r}(\Pa(i, \G)).
\end{align}
Consider the $\G^*$ and $\G'$ in Figure \ref{fig:intro_ex} and \ref{fig:intro_ex_6_edge} again. By supposing that all variables are binary, readers can verify that $|\param(\G^*)| = 21$ and $|\param(\G')| = 23$. Unlike the linear Gaussian case, $|\param(\G)| > |\E(\G)|$ holds for every multinomial causal model $(\G, \Prob)$. 

So much for the basic notations. In the next section, we will utilize them to articulate the crux of this work, namely, \textit{causal razors}.

%%%%%%%%%%%%%%%%%%%%%%%%%%%%%%%%%%%%%%%%%%%%%%%%%%%%%%%%%%%%%%%%%%%%%%
%%%%%%%%%%%%%%%%%%%%%%%%%%%%%%%%%%%%%%%%%%%%%%%%%%%%%%%%%%%%%%%%%%%%%%
%%%%%%%%%%%%%%%%%%%%%%%%%%%%%%%%%%%%%%%%%%%%%%%%%%%%%%%%%%%%%%%%%%%%%%
\section{Structural Causal Razors}
\label{sec:common_rz}
%%%%%%%%%%%%%%%%%%%%%%%%%%%%%%%%%%%%%%%%%%%%%%%%%%%%%%%%%%%%%%%%%%%%%%
%%%%%%%%%%%%%%%%%%%%%%%%%%%%%%%%%%%%%%%%%%%%%%%%%%%%%%%%%%%%%%%%%%%%%%
%%%%%%%%%%%%%%%%%%%%%%%%%%%%%%%%%%%%%%%%%%%%%%%%%%%%%%%%%%%%%%%%%%%%%%
The general goal of causal discovery, as outlined in \textbf{Section \ref{sec:intro}}, is the retrieval of causal information of the true DAG $\G^*$ from the joint probability distribution $\Prob$. This can be rephrased as the behavior of shrinking the hypothesis space $\DAG(\mb{v})$ to the extent that only a desired class of DAGs remains. This shrinking behavior inevitably requires assumptions to be made of the true causal model $(\G^*, \Prob)$ such that DAGs failing the assumptions will be removed from consideration. These assumptions are generically labeled as \textit{causal razors} in this work.  

Roughly speaking, causal razors can be classified into three different groups, though they can possibly overlap. First, a \textit{parametric assumption} can be imposed on the joint probability distribution $\Prob$ of the true causal model such that certain DAGs incompatible with $\Prob$ should be eliminated from the hypothesis space. For example, by assuming that $(\G^*, \Prob)$ is a linear Gaussian causal model, CIs held in $\Prob$ can only be explained by the d-separation relations in $\G^*$ or the coincidental cancellation of linear coefficients induced by \textit{path-cancellation} in $\G^*$.\footnote{See \citep{Weinberger_faithfulness} for a in-depth theoretical discussion of path-cancellation.} For instance, by representing the example in Figure \ref{fig:intro_ex} as a linear Gaussian model, the unfaithful independence between taking contraception pills ($C$) and stroke ($T$) is resulted from the cancellation effect between two directed paths from $C$ to $T$. Another sort of causal razors are \textit{algorithmic} in nature. The set of DAGs prescribed by the causal razor is determined by the output of a specific causal search algorithm. For example, the assumption of \textit{Edge Sparsest Permutation} (ESP) in \citep{solus2021consistency} and \textit{GRaSP-razors} in \citep{GRaSP} are of this sort. 

Nonetheless, the kind of causal razors that draws more attention in the literature is of a different category. Generally speaking, their definitions are expressed without appealing to a specific parametric or algorithmic feature. Their focus is more on a well-defined graphical feature (e.g., edges), or a feature that entailed by a graphical feature (e.g., CIs entailed by d-separation). We label these causal razors as \textit{structural} causal razors. Admittedly, this heuristic categorization is imprecise. But this deliberate vagueness can simplify the exposition in our upcoming comparative analysis.

We first review a list of structural causal razors studied in the literature. Each definition will be unpacked into three parts: \textit{[class]} specifies the class of DAGs satisfying the intended condition, \textit{[property]} describes each DAG in \textit{[class]}, and \textit{[razor]} states the causal razor supposed to be satisfied by the true causal model $(\G^*, \Prob)$. When we speak of a causal razor, we generally refer to the \textit{[razor]} part.\footnote{Following the terminologies in \citep{zhang2013comparison}, \textit{[property]} and \textit{[razor]} of a causal razor can be interpreted as a \textit{methodological assumption} and an \textit{ontological assumption} respectively. To be more precise, thaking the causal razor (e.g., Markov) as a methodological assumption (or principle) means that DAGs not satifying \textit{[property]} (e.g., non-Markovian) are hypotheses that should be rejected according to the assumption. On the other hand, taking the causal razor as an ontological assumption means that the true causal model is assumed to satisfy the \textit{[property]} in question (e.g., causal Markov condition).} Definitions of causal razors are stated within a surrounding box for a more glossarial view for readers.

The first candidate is the \textit{causal Markov condition} (CMC) which serves as the most fundamental pillar of causal discovery.\\
\vspace{0.1cm}

%%%%%%%%%%%%%%%%%%%%
\noindent\fbox{\parbox{\textwidth}{
\begin{definition}
\label{def:Markov}
(Markov) For any joint probability distribution $\Prob$ over $\mb{V}$, 
\begin{enumerate}
    \ib [class] $\CMC(\Prob) := \{\G \in \DAG(\mb{v}): \I(\G) \subseteq \I(\Prob)\}$;
    \ib [property] a DAG $\G$ is Markovian (to $\Prob$) if $\G \in \CMC(\Prob)$;
    \ib [razor] $(\G^*, \Prob)$ satisfies the causal Markov condition if $\G^* \in \CMC(\Prob)$.
\end{enumerate}
\end{definition}}}
\,\\
%%%%%%%%%%%%%%%%%%%%

\noindent CMC requires that the CIs entailed by the true DAG (through d-separation) is a subset of those held in the joint distribution. The definition of CMC is obtained from the \textit{global Markov condition} (i.e., $\I(\G) \subseteq \I(\Prob)$) which is equivalent to each of the following two conditions.\footnote{The proof to the equivalence of the three Markov conditions can be found in \citep{lauritzenlectures}.} First, $\G$ satisfies the \textit{local Markov assumption} relative to $\Prob$ if
\begin{align}
\label{eq:local_Markov}
    X_i \CI_\Prob \mb{X}_{\Nd(i, \G)} \setminus \mb{X}_{\Pa(i, \G)}\,|\, \mb{X}_{\Pa(i, \G)} & \text{\,\,\,\,\,for every $X_i \in \mb{V}$.} 
\end{align}
In words, every variable is independent of its non-descendants by conditioning on all of its parents. This condition helps rewriting the joint probability distribution $\Prob$ by the famous \textit{Markov factorization}:
\begin{align}
\label{eq:Markov_factorization}
    \Prob(\mb{V}) = \prod_{i \in \mb{v}} \Prob(X_i\,|\,\mb{X}_{\Pa(i, \G)}).
\end{align}
Notice that CMC alone cannot attain the desideratum of identifying the true Markov equivalence class (except the trivial case where $\I(\Prob) = \varnothing$). This is because if $\G$ is Markovian, every \textit{supergraph} $\G'$ of $\G$ (i.e., $\E(\G') \subseteq \E(\G)$) is also Markovian. Also, every complete DAG is Markovian trivially since it entails no CI by d-separation. Thus, $\CMC(\Prob)$ still leaves us with a huge hypothesis space to perform causal search. To further contract the search space, a widely discussed causal razor is the \textit{causal faithfulness condition} (CFC) from \citep{spirtes2000causation}.\\

%%%%%%%%%%%%%%%%%%%%
\noindent\fbox{\parbox{\textwidth}{
\begin{definition}
\label{def:faithful}
(Faithfulness) For any joint probability distribution $\Prob$, 
\begin{enumerate}
    \ib [class] $\CFC(\Prob) := \{\G \in \CMC(\Prob): \I(\Prob) \subseteq \I(\G)\}$;
    \ib [property] a DAG $\G$ is faithful (to $\Prob$) if $\G \in \CFC(\Prob)$;
    \ib [razor] $(\G^*, \Prob)$ satisfies the causal faithfulness condition if $\G^* \in \CFC(\Prob)$.
\end{enumerate}
\end{definition}}}
\,\\
%%%%%%%%%%%%%%%%%%%%

By assuming the converse of CMC, CFC further requires that $\I(\G^*) = \I(\Prob)$.\footnote{Instead of following the common practice in defining CFC as the converse of CMC (or $\CFC(\Prob):= \{\G \in \DAG(\mb{v}): \I(\Prob) \subseteq \I(\G)\}$), defining CFC as the conjunction of CMC and its converse (or $\CFC(\Prob):= \{\G \in \CMC(\mb{v}): \I(\Prob) \subseteq \I(\G)\}$ as in \textbf{Definition \ref{def:faithful}}) helps simplifying the exposition in the sense that a faithful but non-Markovian DAG does not exist. The same convention will be used for the \textit{[class]} of each causal razor defined below.} In other words, all CIs held in $\Prob$ can be perfectly \textit{explained} by $\G^*$ through d-separation. Denote $\Psi_{\G, \Prob} = \I(\Prob) \setminus \I(\G)$ as the set of \textit{unfaithful} CIs in $\G \in \CMC(\Prob)$. Sometimes we write $\I(\Prob) = \I(\G) \cupdot \Psi_{\G, \Prob}$ to highlight the disjoint union between $\I(\G)$ and $\Psi_{\G, \Prob}$. Methodologically speaking, when $\G \in \CFC(\Prob)$ while $\G'$ has a non-empty $\Psi_{\G', \Prob}$, then $\G$ is strictly preferred to $\G'$ in terms of explanatory power, and thus $\G'$ should be rejected. 

When CFC is satisfied, the identification of the true MEC is an attainable desideratum because $\MEC(\G^*) = \CFC(\Prob)$ holds. On a positive note, the violation of CFC has a Lebesgue measure of zero in the large sample limit.\footnote{See \citet[pp.41-2]{spirtes2000causation}.} Nevertheless, as argued by \citet{Andersen_CFC}, unfaithfulness does not vanish in small probability because certain parameter values of a scientific model can be obtained through a long-term dynamic equilibrium which can give rise to a violation of CFC. Moreover, as shown by \citet{uhler2013geometry}, learning CIs from observational data by hypothesis testing is error-prone, and so \textit{almost violations} of faithfulness are frequent in finite samples. Back to the example in Figure \ref{fig:intro_ex}, $\G^* \notin \CFC(\Prob)$ because of the almost independence between $C$ and $T$ which is not entailed by $\G^*$ though d-separation. 

By upholding the same desideratum while admitting the frequent almost violations of CFC, researchers are motivated to resort to causal razors strictly weaker than CFC. One earliest approach is the assumption of \textit{restricted faithfulness} offered by \citet{ramsey2006}, which can be decomposed into \textit{adjacency faithfulness} and \textit{orientation faithfulness}.\footnote{Following the definitions of CMC and CFC, we label each \textit{[razor]} starting with the term \textit{causal} to indicate that it is the true causal model that satisfies the associated \textit{[property]}. But this naming custom will often be neglected henceforth for brevity.}\\

%%%%%%%%%%%%%%%%%%%%
\noindent\fbox{\parbox{\textwidth}{
\begin{definition}
\label{def:adj_faith}
(Adjacency faithfulness) For any joint probability distribution $\Prob$ over $\mb{V}$, 
\begin{enumerate}
    \ib [property] a DAG $\G$ is adjacency-faithful (to $\Prob$) if $\la X_i, X_j\,|\,\mb{X}_\mb{s}\ra \notin \I(\Prob)$ for every $(i, j) \in \SK(\G)$ and every $\mb{s} \subseteq \mb{v} \setminus \{i, j\}$;
    \ib [class] $\adjF(\Prob) := \{\G \in \CMC(\Prob): \G \text{ is adjacency-faithful to } \Prob\}$;
    \ib [razor] $(\G^*, \Prob)$ satisfies the causal adj-faithfulness assumption if $\G^* \in \adjF(\Prob)$.
\end{enumerate}
\end{definition}}}
\,\\
%%%%%%%%%%%%%%%%%%%%

%%%%%%%%%%%%%%%%%%%%
\noindent\fbox{\parbox{\textwidth}{
\begin{definition}
\label{def:ori_faith}
(Orientation faithfulness) For any joint probability distribution $\Prob$ over $\mb{V}$, 
\begin{enumerate}
    \ib [property] a DAG $\G$ is orientation-faithful (to $\Prob$) if for every unshielded triple $(i, j, k)$ in $\G$,
    \begin{itemize}
    \item[(a)] if $i \to j \ot k$ holds in $\G$, then $\la X_i, X_k\,|\,\mb{X}_\mb{s}\ra \notin \I(\Prob)$ for every $\mb{s} \subseteq \mb{v} \setminus \{i, k\}$ where $j \in \mb{s}$;
    \item[(b)] otherwise, $\la X_i, X_k\,|\,\mb{X}_\mb{s}\ra \notin \I(\Prob)$ for every $\mb{s} \subseteq \mb{v} \setminus \{i, k\}$ where $j \notin \mb{s}$;
    \end{itemize}
    \ib [class] $\oriF(\Prob) := \{\G \in \CMC(\Prob): \G \text{ is orientation-faithful to } \Prob\}$;
    \ib [razor] $(\G^*, \Prob)$ satisfies the causal ori-faithfulness assumption if $\G^* \in \oriF(\Prob)$.
\end{enumerate}
\end{definition}}}
\,\\
%%%%%%%%%%%%%%%%%%%%

%%%%%%%%%%%%%%%%%%%%
\noindent\fbox{\parbox{\textwidth}{
\begin{definition}
\label{def:res_faith}
(Restricted faithfulness) For any joint probability distribution $\Prob$, 
\begin{enumerate}
    \ib [class] $\resF(\Prob) := \adjF(\Prob) \cap \oriF(\Prob)$;
    \ib [property] a DAG $\G$ is restricted-faithful (to $\Prob$) if $\G \in \resF(\Prob)$;
    \ib [razor] $(\G^*, \Prob)$ satisfies the causal res-faithfulness assumption if $\G^* \in \resF(\Prob)$.
\end{enumerate}
\end{definition}}}
\,\\
%%%%%%%%%%%%%%%%%%%%

According to the classic result by \citet{Verma1990EquivalenceAS}, $\MEC(\G)$ for any given DAG $\G$ can be uniquely identified by its skeleton $\SK(\G)$ and the set of unshielded colliders in $\G$. Note that res-faithfulness assumption ensures that both adjacencies and unshielded colliders in $\G^*$ are correctly identified by $\I(\Prob)$ (through adj-faithfulness and ori-faithfulness respectively). Hence, $\MEC(\G^*)$ is identifiable under res-faithfulness. As proven by \citet{ramsey2006}, the famous PC algorithm (originated by \citet{PC}) can correctly identify $\MEC(\G^*)$ when res-faithfulness holds. For our earlier 5-node example in Figure \ref{fig:intro_ex}, $\G^*$ is not adjacency-faithful since the adjacency between $C$ and $T$ is associated with an (almost) unfaithful independence. Thus, $\G^*$ is not restricted-faithful either. 

A different route to identify $\MEC(\G^*)$ is the \textit{sparsest Markov representation} (SMR) assumption proposed by \citet{raskutti2018learning}. This assumption is defined in terms of the sparsity of $\G^*$, or equivalently, \textit{frugality} in \citep{forster2020frugal}.\\

%%%%%%%%%%%%%%%%%%%%
\noindent\fbox{\parbox{\textwidth}{
\begin{definition}
\label{def:frugal}
(Frugality) For any joint probability distribution $\Prob$,
\begin{enumerate}
    \ib [class] $\Fr(\Prob) := \{\G \in \CMC(\Prob): \neg \exists \G' \in \CMC(\Prob)$ s.t. $|\E(\G')| < |\E(\G)|\}$;
    \ib [property] a DAG $\G$ is frugal (to $\Prob$) if $\G \in \Fr(\Prob)$;
    \ib [razor] $(\G^*, \Prob)$ satisfies the causal frugality assumption if $\G^* \in \Fr(\Prob)$.
\end{enumerate}
\end{definition}}}
\,\\
%%%%%%%%%%%%%%%%%%%%

%%%%%%%%%%%%%%%%%%%%
\noindent\fbox{\parbox{\textwidth}{
\begin{definition}
\label{def:u-frugal}
(Unique frugality / SMR) For any joint probability distribution $\Prob$,
\begin{enumerate}
    \ib [class] $\uFr(\Prob) := $
    $\begin{cases}
    \Fr(\Prob) & \text{if all DAGs in $\Fr(\Prob)$ belong to the same MEC,}\\
    \varnothing & \text{otherwise; }
    \end{cases}$
    \ib [property] a DAG $\G$ is u-frugal (to $\Prob$) if $\G \in \uFr(\Prob)$;
    \ib [razor] $(\G^*, \Prob)$ satisfies the causal u-frugality assumption if $\G^* \in \uFr(\Prob)$.
\end{enumerate}
\end{definition}}}
\,\\
%%%%%%%%%%%%%%%%%%%%

From an ontological point of view, frugality hypothesizes that $\G^*$ is one of the sparsest Markovian DAGs. Methodologically, it states that only the sparsest Markovian DAGs are acceptable hypotheses. Back to the two DAGs in Figure \ref{fig:intro_ex} and \ref{fig:intro_ex_6_edge}, $\G^*$ is a frugal DAG while $\G'$ is not.\footnote{See \textbf{Example \ref{ex:uFr_not_triF}} for the details of $\I(\Prob)$ to verify that no Markovian DAG is sparser than $\G^*$.} On the other hand, u-frugality requires not only frugality but also the \textit{uniqueness} of MEC (i.e., all frugal DAGs are Markov equivalent). Thus, u-frugality is stronger than its non-unique variant by definition. Accordingly, u-frugality imposes an equivalence between $\MEC(\G^*)$ and $\Fr(\Prob)$ and hence the identification of $\MEC(\G^*)$ is straightforward when u-frugality is correctly assumed. In practice, the \textit{Sparsest Permutation} (SP) algorithm in \citep{raskutti2018learning} returns $\Fr(\Prob)$ by enumerating all possible permutations and locating the sparsest induced DAGs. The correctness of SP, hence, is evident when u-frugality is satisfied.\footnote{To be precise, by supposing that $\Prob$ is a \textit{graphoid} (see \textbf{Appendix \ref{app:graphoid}}), SP enumerates all permutations over $\mb{v}$ where each permutation induces a SGS-minimal DAG.} 

Now we turn to some other causal razors studied in the literature. One candidate is the minimality condition sometimes referred to as the \textit{minimal I-map} (e.g., \citet{verma1988causal}). We follow \citet{zhang2013comparison} and label this causal razor as \textit{SGS-minimality} as in \citep{spirtes2000causation}.\\

%%%%%%%%%%%%%%%%%%%%
\noindent\fbox{\parbox{\textwidth}{
\begin{definition}
\label{def:SGS-minimal}
(SGS-minimality) For any joint probability distribution $\Prob$,
\begin{enumerate}
    \ib [class] $\SGS(\Prob) := \{\G \in \CMC(\Prob): \neg \exists \G' \in \CMC(\Prob)$ s.t. $\E(\G') \subset \E(\G)\}$;
    \ib [property] a DAG $\G$ is SGS-minimal (to $\Prob$) if $\G \in \SGS(\Prob)$;
    \ib [razor] $(\G^*, \Prob)$ satisfies the causal SGS-minimality assumption if $\G^* \in \SGS(\Prob)$.
\end{enumerate}
\end{definition}}}
\,\\
%%%%%%%%%%%%%%%%%%%%

In plain words, SGS-minimality requires that no subgraph is Markovian. Similar to frugality, SGS-minimality is an \textit{edge-minimality} condition but is defined in terms of set containment instead of cardinality. This causal razor plays a crucial role in \textit{permutation-based} causal search algorithms (e.g., SP). As proven by \cite{verma1988causal}, a DAG constructed from a topological ordering of $\mb{v}$ and a \textit{semigraphoid} (e.g., a joint probability distribution) must be SGS-minimal.\footnote{See \textbf{Appendix \ref{app:graphoid}} for the definition of a semigraphoid.} In addition, as argued by \cite{ZhangSpirtes2011}, SGS-minimality is a fairly safe assumption to be made, particularly under an interventionist interpretation of causality.\footnote{Readers might suspect why the unique variant of SGS-minimality is not introduced above. See footnote \normalfont{\ref{u-SGS-min}} on why it is an uninteresting causal razor.}

Another causal razor that has drawn wide attention is the minimality principle studied in \cite{Pearl2009Causality}. Following the naming convention in \citep{zhang2013comparison}, we denote this causal razor as \textit{Pearl-minimality}, or \textit{P-minimality} for short.\\

%%%%%%%%%%%%%%%%%%%%
\noindent\fbox{\parbox{\textwidth}{
\begin{definition}
\label{def:P-minimal}
(P-minimality) For any joint probability distribution $\Prob$,
\begin{enumerate}
    \ib [class] $\Pm(\Prob) := \{\G \in \CMC(\Prob): \neg \exists \G' \in \CMC(\Prob)$ s.t. $\I(\G) \subset \I(\G')\}$;
    \ib [property] a DAG $\G$ is P-minimal (to $\Prob$) if $\G \in \Pm(\Prob)$;
    \ib [razor] $(\G^*, \Prob)$ satisfies the causal P-minimality assumption if $\G^* \in \Pm(\Prob)$.
\end{enumerate}
\end{definition}}}
\,\\
%%%%%%%%%%%%%%%%%%%%

Given that a Markovian DAG is a hypothesis that purports to explain the CIs in $\I(\Prob)$, P-minimality requires the set of CIs \textit{unexplained} by $\G$ (i.e., $\Psi_{\G, \Prob}$) to be minimal in a set-theoretical sense. Methodologically speaking, if a Markovian DAG $\G$ can explain a superset of CIs compared to another Markovian DAG $\G'$ (i.e., $\I(\G') \subset \I(\G) \subseteq \I(\Prob)$), the latter should be rejected. \citet{zhang2013comparison} offered several arguments in support of P-minimal as an ontological assumption, particularly when $\G^*$ is relatively sparse with few triangles. Also, by pursuing a learning-theoretic approach, \citet{LinZhang2020} show that certain important convergence properties must be sacrificed when P-minimality is violated. 

Next, analogous to the relation between frugality and u-frugality, we introduce the strengthened form of P-minimality with a uniqueness condition.\\

%%%%%%%%%%%%%%%%%%%%
\noindent\fbox{\parbox{\textwidth}{
\begin{definition}
\label{def:u-P-minimal}
(Unique P-minimality) For any joint probability distribution $\Prob$,
\begin{enumerate}
    \ib [class] $\uPm(\Prob) := $
    $\begin{cases}
    % \Pm(\Prob) & \text{if $\G' \in \MEC(\G)$ for every $\G, \G' \in \Pm(\Prob)$}\\
    \Pm(\Prob) & \text{if all DAGs in $\Pm(\Prob)$ belong to the same MEC,}\\
    \varnothing & \text{otherwise; }
    \end{cases}$
    \ib [property] a DAG $\G$ is u-P-minimal (to $\Prob$) if $\G \in \uPm(\Prob)$;
    \ib [razor] $(\G^*, \Prob)$ satisfies the causal u-P-minimality assumption if $\G^* \in \uPm(\Prob)$.
\end{enumerate}
\end{definition}}}
\,\\
%%%%%%%%%%%%%%%%%%%%

\noindent Unlike the other discussed causal razors, u-P-minimality does not attract much attention except by \citet{LinZhang2020} and \citet{GRaSP}. Despite its unpopularity, the latter authors prove that u-P-minimality and CFC are logically equivalent. We will recast their simple proof in \textbf{Appendix \ref{app:CFC_uPM}}, and discuss an algorithmic implication of this logical equivalence in \textbf{Section \ref{sec:alg_imp}}.

We now turn to a causal razor that has been extensively studied in \citep{kTriangle}, \citep{zhang2008detection}, and \citep{zhang2013comparison}, namely, the assumption of \textit{triangle faithfulness}.\\

%%%%%%%%%%%%%%%%%%%%%%%%%%%%%%%%%%%%%%%%%%%%%%%%%%%%%%%%%%%%%%%%%%%%%%
%%%%%%%%%%%%%%%%%%%%%%%%%%%%%%%%%%%%%%%%%%%%%%%%%%%%%%%%%%%%%%%%%%%%%%
\noindent\fbox{\parbox{\textwidth}{
\begin{definition}
\label{def:tri_faith}
(Triangle faithfulness) Given a joint probability distribution $\Prob$ over $\mb{V}$, 
\begin{enumerate}
    \ib [property] a DAG $\G$ is triangle-faithful (to $\Prob$) if for every shielded triple $(i, j, k)$ in $\G$,
    \begin{itemize}
    \item[(a)] if $i \to j \ot k$ holds in $\G$, then $\la X_i, X_k\,|\,\mb{X}_\mb{s}\ra \notin \I(\Prob)$ for any $\mb{s} \subseteq \mb{v} \setminus \{i, k\}$ where $j \in \mb{s}$;
    \item[(b)] otherwise, $\la X_i, X_k\,|\,\mb{X}_\mb{s}\ra \notin \I(\Prob)$ for any $\mb{s} \subseteq \mb{v} \setminus \{i, k\}$ where $j \notin \mb{s}$;
    \end{itemize}
    \ib [class] $\triF(\Prob) := \{\G \in \CMC(\Prob): \G \text{ is triangle-faithful to } \Prob\}$;
    \ib [razor] $(\G^*, \Prob)$ satisfies the causal tri-faithfulness assumption if $\G^* \in \triF(\Prob)$.
\end{enumerate}
\end{definition}}}\\
%%%%%%%%%%%%%%%%%%%%%%%%%%%%%%%%%%%%%%%%%%%%%%%%%%%%%%%%%%%%%%%%%%%%%%
%%%%%%%%%%%%%%%%%%%%%%%%%%%%%%%%%%%%%%%%%%%%%%%%%%%%%%%%%%%%%%%%%%%%%%

Readers might find tri-faithfulness highly similar to ori-faithfulness where the former is defined over triangles and the latter over unshielded triples. Nevertheless, tri-faithfulness is a logical consequence of adj-faithfulness (but not of ori-faithfulness). There are two noteworthy remarks concerning tri-faithfulness. The first First, it relates to the \textit{detectability of unfaithfulness}. To be precise, the violation of any \textit{[razor]} can be binarized as \textit{detectable} and \textit{undetectable}; detectable if its associated \textit{[class]} is empty, and undetectable if \textit{[class]} is non-empty but not containing $\G^*$. As proven by \citet{zhang2008detection}, any undetectable violation of CFC must be a consequence of the violation of tri-faithfulness. On the other hand, as shown in \citep{zhang2013comparison}, the conjunction of SGS-minimality and tri-faithfulness entails P-minimality. This buttresses his claim that P-minimality is relatively safe to be assumed when $\G^*$ is relatively sparse with a small number of triangles.

The eleven causal razors defined in this section do not exhaust all that have been introduced in the literature. For example, \textit{NOI-minimality} in \citep{Zhalama2019asp} requires that the true DAG entails the \textit{greatest number} of CIs over all Markovian DAGs. \textit{Single shielded/unshielded-collider-faithfulness} in \citep{Ng_local_A_star} weakens the tri-/ori-faithfulness assumptions. Readers are recommended to relate other candidates in the literature to those defined above for a more thorough logical exposition. Our selective focus on the eleven causal razors above is primarily due to their tight logical hierarchies that we are going to explore in the coming section.\\

%%%%%%%%%%%%%%%%%%%%%%%%%%%%%%%%%%%%%%%%%%%%%%%%%%%%%%%%%%%%%%%%%%%%%%
%%%%%%%%%%%%%%%%%%%%%%%%%%%%%%%%%%%%%%%%%%%%%%%%%%%%%%%%%%%%%%%%%%%%%%
%%%%%%%%%%%%%%%%%%%%%%%%%%%%%%%%%%%%%%%%%%%%%%%%%%%%%%%%%%%%%%%%%%%%%%
\section{Logical Hierarchy of Structural Causal Razors}
\label{sec:rz_hier}
%%%%%%%%%%%%%%%%%%%%%%%%%%%%%%%%%%%%%%%%%%%%%%%%%%%%%%%%%%%%%%%%%%%%%%
%%%%%%%%%%%%%%%%%%%%%%%%%%%%%%%%%%%%%%%%%%%%%%%%%%%%%%%%%%%%%%%%%%%%%%
%%%%%%%%%%%%%%%%%%%%%%%%%%%%%%%%%%%%%%%%%%%%%%%%%%%%%%%%%%%%%%%%%%%%%%

In this section, the eleven causal razors defined above will be compared in terms of their logical strength. This can be achieved by comparing their \textit{[class]}'s in terms of possible pairwise subset relations. Most cases where two \textit{[class]}'s stand in a subset relation involve simple proofs (if not true by definition). We are not going to recast the specifics except for the relatively new result that CFC and u-P-minimality are logically equivalent in the \textbf{Appendix \ref{app:CFC_uPM}}. Contrarily, a counterexample will be covered for each case where the subset relation fails.

\begin{theorem}
\label{thm:rz_sb}
For any joint probability distribution $\Prob$, the following statements are true:
\begin{enumerate}
    \item[(a)] $\CFC(\Prob) \subseteq \resF(\Prob) = \adjF(\Prob) \cap \oriF(\Prob) \subseteq \uFr(\Prob) \subseteq \Fr(\Prob) \subseteq \Pm(\Prob) \subseteq \SGS(\Prob) \subseteq \CMC(\Prob)$;
    \item[(b)] $\adjF(\Prob) \subseteq \triF(\Prob)$;
    \item[(c)] $\adjF(\Prob) \subseteq \Fr(\Prob)$.
\end{enumerate}
\end{theorem}
\begin{pf}
For (a), \textbf{Definition \ref{def:res_faith}, \ref{def:u-frugal}, \ref{def:SGS-minimal}} give us $\resF(\Prob) = \adjF(\Prob) \cap \oriF(\Prob)$, $\uFr(\Prob) \subseteq \Fr(\Prob)$ and $\SGS(\Prob) \subseteq \CMC(\Prob)$ respectively. $\CFC(\Prob) \subseteq \resF(\Prob)$ is proven in \citep{ramsey2006}, $\resF(\Prob) \subseteq \uFr(\Prob)$ in \citep{raskutti2018learning}, $\Fr(\Prob) \subseteq \Pm(\Prob)$ in \citep{forster2020frugal}, and $\Pm(\Prob) \subseteq \SGS(\Prob)$ in \citep{zhang2013comparison}. (b) is true by \textbf{Definition \ref{def:tri_faith}}.

For (c), by reductio, suppose $\G \in \DAG(\mb{v})$ where $\G \in \adjF(\Prob) \setminus \Fr(\Prob)$. So, there exists $\G' \in \CMC(\Prob)$ where $|\E(\G)| > |\E(\G')|$. There must exist $(i, j) \in \SK(\G)$ such that $(i, j) \notin \SK(\G')$. Given that $(i, j) \notin \SK(\G')$, $\la X_i, X_j \mid \mb{X}_{\mb{s}}\ra \in \I(\G') \subseteq \I(\Prob)$ for some $\mb{s} \subseteq \mb{v} \setminus \{i, j\}$. However, since $(i, j) \in \SK(\G)$, it contradicts that $\G \in \adjF(\Prob)$.
\end{pf}\\

\begin{theorem}
\label{CFC-uPm}
\cite{GRaSP} For any joint probability distribution $\Prob$, $\CFC(\Prob) = \uPm(\Prob)$.
\end{theorem}
\begin{pf}
See \textbf{Appendix \ref{app:CFC_uPM}}.
\end{pf}\\

Next, we turn to the counterexamples where the subset relations fail. To simplify our exposition, only CIs held between two singleton sets in an independence model will be listed (e.g., $\la X, Y\,|\,\mb{Z}\ra$). Meticulous readers can verify that the independence model $\I(\Prob)$ of each example below induces a (\textit{compositional}) \textit{graphoid} (see \textbf{Appendix \ref{app:graphoid}}).\\

%%%%%%%%%%%%%%%%%%%%
\begin{example}
\label{ex:resF_not_CFC}
There exists a joint probability distribution $\Prob$ such that $\resF(\Prob) \setminus \CFC(\Prob) \neq \varnothing$.
\end{example}
\noindent Consider the DAG $\G^*$ in \text{Figure} \ref{fig:resF_not_CFC}. Let $\Prob$ be the joint probability distribution where $\I(\Prob) = \I(\G^*) \cupdot \{\la X_1, X_6\ra\}$. Obviously, $\G^* \notin \CFC(\Prob)$. But $X_1 \CI_\Prob X_6$ is not associated with any adjacency or unshielded triple in $\G^*$. Hence, $\G^* \in \resF(\Prob)$.\footnote{This example can be easily constructed as a linear Gaussian causal model using path-cancellation. Given that $\bs{\theta}_{\G^*} = \{\theta_{12}, \theta_{14}, \theta_{23}, \theta_{36},  \theta_{45}, \theta_{56}\}$, we set $\theta_{12} \theta_{23} \theta_{36} = -(\theta_{14} \theta_{45} \theta_{56})$. Accordingly, the two directed paths $1 \to 2 \to 3 \to 6$ and $1 \to 4 \to 5 \to 6$ in $\G^*$ cancel out and induce $\la X_1, X_6\ra \in \Psi_{\G^*, \Prob}$.}\hfill $\square$\\
%%%%%%%%%%%%%%%%%%%%

%%%%%%%%%%%%%%%%%%%%
\begin{figure}[h]
    \centering
    \begin{tikzpicture}
    \node(V1) at (0,0) {$1$};
    \node(V2) at (1.5,0.75) {$2$};
    \node(V3) at (3,0.75) {$3$};
    \node(V4) at (1.5,-0.75) {$4$};
    \node(V5) at (3,-0.75) {$5$};
    \node(V6) at (4.5, 0) {$6$};
    \node(G) at (5, -0.75) {$\G^*$};
    \path[->, line width=0.4mm] (V1) edge (V2);
    \path[->, line width=0.4mm] (V2) edge (V3);
    \path[->, line width=0.4mm] (V1) edge (V4);
    \path[->, line width=0.4mm] (V4) edge (V5);
    \path[->, line width=0.4mm] (V3) edge (V6);
    \path[->, line width=0.4mm] (V5) edge (V6);
    \end{tikzpicture}
    \caption{$\G^* \in \resF(\Prob) \setminus \CFC(\Prob)$ due to $\Psi_{\G^*, \Prob} = \{\la X_1, X_6\ra\}$}
    \label{fig:resF_not_CFC}
\end{figure}
%%%%%%%%%%%%%%%%%%%%

%%%%%%%%%%%%%%%%%%%%
\begin{example}
\label{ex:uFr_not_triF}
There exists a joint probability distribution $\Prob$ such that $\uFr(\Prob) \setminus \triF(\Prob) \neq \varnothing$.
\end{example}
\noindent This example is from \citep[supplementary materials]{GRaSP}.\footnote{To construct the example as a linear Gaussian causal model $(\G^*, \Prob)$, set $-\theta_{15} = \theta_{14} \theta_{45}$ such that the two directed paths $1 \to 5$ and $1 \to 4 \to 5 $ cancel out and induce the unfaithful $X_1 \CI_\Prob X_5$. The other unfaithful CIs (i.e., $\{\psi_2, ..., \psi_8\}$) are entailed by $\I(\G^*) \cupdot \{\la X_1, X_5\ra\}$ using graphoid axioms (including intersection and composition).} Consider the DAG $\G^*$ in Figure \ref{fig:uFr_not_triF}. Let $\Prob$ be the joint probability distribution with $\I(\Prob) = \I(\G^*) \cupdot \Psi_{\G^*, \Prob}$ where
\begin{align*}
    \Psi_{\G^*, \Prob} = &
\begin{Bmatrix}
    \psi_1: \la X_1, X_5\ra, &
    \psi_2: \la X_1, X_2\,|\,\{X_5\}\ra, \\ 
    \psi_3: \la X_1, X_3\,|\,\{X_5\}\ra, &
    \psi_4: \la X_1, X_5\,|\,\{X_2\}\ra, \\
    \psi_5: \la X_1, X_5\,|\,\{X_3\}\ra, &
    \psi_6: \la X_1, X_2\,|\,\{X_3, X_5\}\ra, \\
    \psi_7: \la X_1, X_3\,|\,\{X_2, X_5\}\ra, &
    \psi_8: \la X_1, X_5\,|\,\{X_2, X_3\}\ra
\end{Bmatrix}.
\end{align*}
Readers can verify that every Markovian DAG is either Markov equivalent to $\G^*$ or denser than $\G^*$.\footnote{To verify so, one simple way is to induce a DAG $\G_\pi$ from each \textit{permutation} $\pi$ over the set of vertices $\mb{v}$ by the DAG-inducing method discussed in \citep{Verma1990EquivalenceAS} or \citep{raskutti2018learning}.\label{ft:permutation_based}} Hence, $\G^* \in \uFr(\Prob)$. Next, $\G^* \notin \triF(\Prob)$ because of the triangle $(1, 4, 5)$ and $\psi_1$. Indeed, one can also show that $\triF(\Prob) = \varnothing$.\footnote{By enumerating all SGS-minimal DAGs (using the method discussed in footnote \ref{ft:permutation_based}), one can verify that none of them is tri-faithful. Next, observe that if $\G \notin \triF(\Prob)$, then $\G' \notin \triF(\Prob)$ for any $\G' \in \DAG(\mb{v})$ where $\E(\G) \subseteq \E(\G')$. Since each Markovian DAG is a supergraph of some SGS-minimal DAG, no Markovian DAG is tri-faithful. Hence, $\triF(\Prob) = \varnothing$.}\hfill $\square$\\
%%%%%%%%%%%%%%%%%%%%

%%%%%%%%%%%%%%%%%%%%
\begin{figure}[h]
    \centering
    \begin{tikzpicture}
    \node (X1) at (0.0,1.0) {$1$};
    \node (X2) at (0.9510565162951535,0.30901699437494745) {$2$};
    \node (X3) at (0.5877852522924732,-0.8090169943749473) {$3$};
    \node (X4) at (-0.587785252292473,-0.8090169943749476) {$4$};
    \node (X5) at (-0.9510565162951536,0.30901699437494723) {$5$};
    \node (label) at (2, -0.4) {$\G^*$};
    \path [->,line width=0.4mm] (X1) edge (X5);
    \path [->,line width=0.4mm] (X2) edge (X5);
    \path [->,line width=0.4mm] (X3) edge (X5);
    \path [->,line width=0.4mm] (X4) edge (X5);
    \path [->,line width=0.4mm] (X1) edge (X4);
    \end{tikzpicture}
    \caption{$\G^* \in \uFr(\Prob) \setminus \triF(\Prob)$ due to $\la X_1, X_5 \ra \in \Psi_{\G^*, \Prob}$.}
    \label{fig:uFr_not_triF}
\end{figure}
%%%%%%%%%%%%%%%%%%%%

%%%%%%%%%%%%%%%%%%%%
\begin{example}
\label{ex:Fr_not_uFr}
There exists a joint probability distribution $\Prob$ such that
\begin{enumerate}
    \item[(a)] $\Fr(\Prob) \setminus \uFr(\Prob) \neq \varnothing$;
    \item[(b)] $\triF(\Prob) \setminus \uFr(\Prob) \neq \varnothing$;
    \item[(c)] $\oriF(\Prob) \setminus \uFr(\Prob) \neq \varnothing$;
    \item[(d)] $\triF(\Prob) \setminus \adjF(\Prob) \neq \varnothing$;
    \item[(e)] $\oriF(\Prob) \setminus \triF(\Prob) \neq \varnothing$.
\end{enumerate}
\end{example}
\noindent This example is from \citet{forster2020frugal}.\footnote{To construct the example as a linear Gaussian model, set $-\theta_{14} = \theta_{12} \theta_{23} \theta_{34}$ such that $1 \to 2 \to 3 \to 4$ and $1 \to 4$ cancel out and induce the unfaithful $X_1 \CI_\Prob X_4$.} Consider $\G^*$ and $\G'$ in Figure \ref{fig:Fr_not_uFr}. Let $\Prob$ be the joint probability distribution where $\I(\Prob) = \I(\G^*) \cupdot \{\la X_1, X_4\ra\}$. Readers can easily verify that $\G^*, \G' \in \Fr(\Prob)$.\footnote{$\Fr(\Prob)$ can be obtained by the SP algorithm which enumerates all permutation-inducing DAGs as mentioned in footnote \ref{ft:permutation_based}.\label{ft:Fr_SP}} However, $\uFr(\Prob) = \varnothing$ because $\G' \notin \MEC(\G^*)$. So, (a) is proven. Also, we have $\G^* \in \oriF(\Prob) \cap \triF(\Prob)$ because the only unfaithful CI of $\G^*$ is $X_1 \CI_\Prob X_4$ which is not associated with any unshielded triple or triangle in $\G^*$. However, $\G^* \notin \adjF(\Prob)$ due to $(1, 4) \in \SK(\G^*)$ and the unfaithful CI. Thus, (b)-(d) are proven as well. Lastly, the only unfaithful CI $X_1 \CI_\Prob X_3\,|\,\{X_2\}$ of $\G'$ is not associated with any unshielded triple, but it instantiates a violation of tri-faithfulness at the triangle $(1, 3, 2)$. Hence, (e) follows immediately.\hfill $\square$\\
%%%%%%%%%%%%%%%%%%%%

%%%%%%%%%%%%%%%%%%%%
\begin{figure}[h]
    \centering
    \subfloat{
    \begin{tikzpicture}
    \node(V1) at (0,2) {$1$};
    \node(V2) at (2,2) {$2$};
    \node(V3) at (2,0) {$3$};
    \node(V4) at (0,0) {$4$};
    \node(G) at (3, 0) {$\G^*$};
    \path[->, line width=0.5mm] (V1) edge (V2);
    \path[->, line width=0.5mm] (V2) edge (V3);
    \path[->, line width=0.5mm] (V3) edge (V4);
    \path[->, line width=0.5mm] (V1) edge (V4);
    \end{tikzpicture}}
    \hspace{3cm}
    \subfloat{
    \begin{tikzpicture}
    \node(V1) at (0,2) {$1$};
    \node(V2) at (2,2) {$2$};
    \node(V3) at (2,0) {$3$};
    \node(V4) at (0,0) {$4$};
    \node(G) at (3, 0) {$\G'$};
    \path [->, line width=0.5mm] (V1) edge (V2);
    \path [->, line width=0.5mm] (V1) edge (V3);
    \path [->, line width=0.5mm] (V3) edge (V2);
    \path [->, line width=0.5mm] (V4) edge (V3);
    \end{tikzpicture}}
    \caption{$\G^*, \G' \in \Fr(\Prob)$ but $\uFr(\Prob) = \varnothing$.}
    \label{fig:Fr_not_uFr}
\end{figure}
%%%%%%%%%%%%%%%%%%%%

%%%%%%%%%%%%%%%%%%%%
\begin{example}
\label{ex:uFr_not_oriF}
There exists a joint probability distribution $\Prob$ such that
\begin{enumerate}
    \item[(a)] $\uFr(\Prob) \setminus \oriF(\Prob) \neq \varnothing$;
    \item[(b)] $\adjF(\Prob) \setminus \oriF(\Prob) \neq \varnothing$;
    \item[(c)] $\Pm(\Prob) \setminus \Fr(\Prob) \neq \varnothing$.
\end{enumerate}
\end{example}
\noindent Consider $\G^*$ in Figure \ref{fig:uFr_not_oriF}. Let $\Prob$ be the joint probability distribution where $\I(\Prob) = \I(\G^*) \cupdot \{\la X_1, X_4\ra\}$.\footnote{Like above, by setting $\theta_{12} \theta_{24} = -(\theta_{13} \theta_{34})$ in the linear Gaussian model $(\G^*, \Prob)$, the two paths $1 \to 2 \to 4$ and $1 \to 3 \to 4$ cancel out and induce the unfaithful $X_1 \CI_\Prob X_4$.}  We leave the readers to verify that $\G^* \in \uFr(\Prob)$.\footnote{Similar to footnote \ref{ft:Fr_SP}, we obtain $\Fr(\Prob)$ by the SP algorithm and verify that all DAGs in $\Fr(\Prob)$ belong to $\MEC(\G^*)$.} Next, notice that the only unfaithful CI $X_1 \CI_\Prob X_4$ of $\G^*$ is not associated with any adjacency in $\G^*$. Thus, we have $\G^* \in \adjF(\Prob)$ as in (a). However, it elicits a violation of ori-faithfulness which requires that $X_1$ and $X_4$ are conditionally dependent on $X_2$ and/or $X_3$. So, $\G^* \notin \oriF(\Prob)$ as in (b). Contrarily, consider $\G'$ in Figure \ref{fig:uFr_not_oriF} where $\I(\G') = \{\la X_1, X_4\ra\} \subset \I(\Prob)$. Obviously, $\G' \notin \Fr(\Prob)$ because $|\E(\G^*)| < |\E(\G')|$. Also, there is no way to extend $\I(\G')$ to obtain a Markovian DAG. Hence, $\G' \in \Pm(\Prob) \setminus \Fr(\Prob)$ as in (c).\hfill $\square$
%%%%%%%%%%%%%%%%%%%%

%%%%%%%%%%%%%%%%%%%%
\begin{figure}[h]
    \centering
    \subfloat{
    \begin{tikzpicture}
    \node(V1) at (0,2) {$1$};
    \node(V2) at (2,2) {$2$};
    \node(V3) at (2,0) {$4$};
    \node(V4) at (0,0) {$3$};
    \node(G) at (3, 0) {$\G^*$};
    \path[->, line width=0.5mm] (V1) edge (V2);
    \path[->, line width=0.5mm] (V2) edge (V3);
    \path[->, line width=0.5mm] (V4) edge (V3);
    \path[->, line width=0.5mm] (V1) edge (V4);
    \end{tikzpicture}}
    \hspace{3cm}
    \subfloat{
    \begin{tikzpicture}
    \node(V1) at (0,2) {$1$};
    \node(V2) at (2,2) {$2$};
    \node(V3) at (0,0) {$3$};
    \node(V4) at (2,0) {$4$};
    \node(G) at (3, 0) {$\G'$};
    \path[->, line width=0.5mm] (V1) edge (V2);
    \path[->, line width=0.5mm] (V1) edge (V3);
    \path[->, line width=0.5mm] (V4) edge (V2);
    \path[->, line width=0.5mm] (V4) edge (V3);
    \path[->, line width=0.5mm] (V2) edge (V3);
    \end{tikzpicture}}
    \caption{$\Psi_{\G^*, \Prob} = \{\la X_1, X_4\ra\}$ whereas $\G' \in \Pm(\Prob) \setminus \Fr(\Prob).$}
    \label{fig:uFr_not_oriF}
\end{figure}
%%%%%%%%%%%%%%%%%%%%

%%%%%%%%%%%%%%%%%%%%
\begin{example}
\label{ex:adjF_not_uFr}
There exists a joint probability distribution $\Prob$ such that $\adjF(\Prob) \setminus \uFr(\Prob) \neq \varnothing$.
\end{example}
\noindent Let $\Prob$ be the joint probability distribution where $\I(\Prob) = \{\la X_1, X_3\ra, \la X_1, X_3\,|\,\{X_2\}\ra\}$.\footnote{See \textbf{Appendix \ref{app:ex_adjF_not_paramM}} for the construction of the multinomial causal model $(\G_0, \Prob)$.} Consider the two Markovian DAGs $\G_0, \G_1$ in Figure \ref{fig:adjF_not_uFr}. We have $\G_0, \G_1 \in \adjF(\Prob)$ since each of their unfaithful CI is not associated with an adjacency. However, they are frugal DAGs that do not belong to the same MEC. Hence, $\uFr(\Prob) = \varnothing$.\hfill $\square$\\
%%%%%%%%%%%%%%%%%%%%

%%%%%%%%%%%%%%%%%%%%
\begin{figure}[h]
    \centering
    \subfloat{
    \begin{tikzpicture}
    \node(V1) at (0,0) {$1$};
    \node(V2) at (1.5,1.5) {$2$};
    \node(V3) at (3,0) {$3$};
    \node(G) at (3.75, 1.5) {$\G_0$};
    \path[->, line width=0.5mm] (V1) edge (V2);
    \path[->, line width=0.5mm] (V2) edge (V3);
    \end{tikzpicture}}
    \hspace{3cm}
    \subfloat{
    \begin{tikzpicture}
    \node(V1) at (0,0) {$1$};
    \node(V2) at (1.5,1.5) {$2$};
    \node(V3) at (3,0) {$3$};
    \node(G) at (3.75, 1.5) {$\G_1$};
    \path[->, line width=0.5mm] (V1) edge (V2);
    \path[->, line width=0.5mm] (V3) edge (V2);
    \end{tikzpicture}}
    \caption{Even though both $\G_0, \G_1 \in \adjF(\Prob)$, they do not belong to the same MEC and thus $\uFr(\Prob) = \varnothing$.}
    \label{fig:adjF_not_uFr}
\end{figure}
%%%%%%%%%%%%%%%%%%%%

%%%%%%%%%%%%%%%%%%%%
\begin{example}
\label{ex:CMC_not_SGSM}
There exists a joint probability distribution $\Prob$ such that 
\begin{enumerate}
    \item[(a)] $\CMC(\Prob) \setminus \SGS(\Prob) \neq \varnothing$;
    \item[(b)] $\SGS(\Prob) \setminus \Pm(\Prob) \neq \varnothing$.
\end{enumerate}
\end{example}
\noindent Consider the three DAGs in Figure \ref{fig:CMC_not_SGSM}. Let $\Prob$ be the joint probability distribution where $\I(\Prob) = \{\la X_1, X_3\ra\}$. Thus, $\G^* \in \CFC(\Prob)$. Note that $\G' \in \CMC(\Prob)$ is a supergraph of $\G^*$. It entails that $\G' \notin \SGS(\Prob)$. On the other hand, $\G'' \in \SGS(\Prob)$ because no subgraph of $\G''$ is Markovian. However, $\G'' \notin \Pm(\Prob)$ because $\varnothing = \I(\G'') \subset \I(\G^*) = \I(\Prob)$.\footnote{\label{u-SGS-min}The faithful $(\G^*, \Prob)$ can be easily constructed as a linear Gaussian or a multinomial causal model. On the other hand, this example tells us that $\G^*$ and $\G''$ are both SGS-minimal to $\Prob$ where they do not belong to the same MEC. This implies that the unique variant of SGS-minimality fails even if CFC is satisfied. This renders unique-SGS-minimality an overly strong and uninteresting causal razor.

Further, notice that $\G' \notin \SGS(\Prob)$ but $\G'' \in \SGS(\Prob)$. This indicates that SGS-minimality is not \textit{MEC-preserving}. To be more accurate, a causal razor is MEC-preserving if every DAG in the same MEC is contained in the \textit{[class]} specified by the causal razor, or none belongs to the \textit{[class]}. Similarly, this example shows that tri-faithfulness is not MEC-preserving either: $\G' \in \triF(\Prob)$ while $\G'' \notin \triF(\Prob)$. Readers can verify that all causal razors covered in this work, except SGS-minimality and tri-faithfulness, are MEC-preserving.}\hfill $\square$
%%%%%%%%%%%%%%%%%%%%

%%%%%%%%%%%%%%%%%%%%
\begin{figure}[h]
    \centering
    \subfloat{
    \begin{tikzpicture}
    \node(V1) at (0,0) {$1$};
    \node(V2) at (1.5,1.5) {$2$};
    \node(V3) at (3,0) {$3$};
    \node(G) at (3.2, 1.5) {$\G^*$};
    \path[->, line width=0.5mm] (V1) edge (V2);
    \path[->, line width=0.5mm] (V3) edge (V2);
    \end{tikzpicture}}
    \hspace{0.5cm}
    \subfloat{
    \begin{tikzpicture}
    \node(V1) at (0,0) {$1$};
    \node(V2) at (1.5,1.5) {$2$};
    \node(V3) at (3,0) {$3$};
    \node(G) at (3.2, 1.5) {$\G'$};
    \path[->, line width=0.5mm] (V1) edge (V2);
    \path[->, line width=0.5mm] (V3) edge (V2);
    \path[->, line width=0.5mm] (V1) edge (V3);
    \end{tikzpicture}}
    \hspace{0.5cm}
    \subfloat{
    \begin{tikzpicture}
    \node(V1) at (0,0) {$1$};
    \node(V2) at (1.5,1.5) {$2$};
    \node(V3) at (3,0) {$3$};
    \node(G) at (3.2, 1.5) {$\G''$};
    \path[->, line width=0.5mm] (V1) edge (V2);
    \path[->, line width=0.5mm] (V2) edge (V3);
    \path[->, line width=0.5mm] (V1) edge (V3);
    \end{tikzpicture}}
    \caption{$\G^*$ is the unique DAG faithful to $\Prob$ where $\I(\Prob) = \{\la X_1, X_3\ra\}$. Also, $\G', \G'' \in \CMC(\Prob) \setminus \Pm(\Prob)$.}
    \label{fig:CMC_not_SGSM}
\end{figure}
%%%%%%%%%%%%%%%%%%%%

%%%%%%%%%%%%%%%%%%%%
\begin{example}
\label{ex:oriF_triF_not_SGSM}
There exists a joint probability distribution $\Prob$ such that
\begin{enumerate}
    \item[(a)] $\oriF(\Prob) \setminus \SGS(\Prob) \neq \varnothing$;
    \item[(b)] $\triF(\Prob) \setminus \SGS(\Prob) \neq \varnothing$.
\end{enumerate}
\end{example}
\noindent Let $\Prob$ be the joint probability distribution over $\mb{V} = \{X_1, X_2\}$. Consider $\G^*, \G' \in \DAG(\mb{v})$ where $\G^*$ is an empty DAG and $\G'$ is a single-edge DAG. Given that $\I(\Prob) = \{\la X_1, X_2\ra\}$, we have $\G' \in \oriF(\Prob) \cap \triF(\Prob)$ vacuously. But $\G' \notin \SGS(\Prob)$ because $\G'$ is a supergraph of $\G^* \in \CFC(\Prob)$.\hfill $\square$\\
%%%%%%%%%%%%%%%%%%%%

We summarize all the pairwise possible subset relations among the \textit{[class]}'s of the eleven structural causal razors in Table \ref{tab:rz_pairwise_stru} with the applications of a simple set-theoretic fact below. 

%%%%%%%%%%%%%%%%%%%%
\begin{fact}
\label{fact:dagger}
Consider any two sets $\mb{A}$ and $\mb{B}$, and also any $\mb{A}^+ \supseteq \mb{A}$ and $\mb{B}^- \subseteq \mb{B}$. If $\mb{A} \setminus \mb{B} \neq \varnothing$, then $\mb{A}^+ \setminus \mb{B}^- \neq \varnothing$. 
\end{fact}
%%%%%%%%%%%%%%%%%%%%

Now we introduce a new concept to binarize the eleven causal razors. Recall that the violation of a causal razor is detectable if its associated \textit{[class]} is empty. Observe that certain causal razors \textit{can never be} detectably violated. For example, $\Fr(\Prob)$ is necessarily non-empty for any joint probability distribution $\Prob$ since there always exists a sparsest Markovian DAG. We say that a causal razor is \textit{always realizable} if its associated \textit{[class]} is never empty. Consequently, a causal razor is \textit{not} always realizable if there exists a joint probability distribution that renders the associated \textit{[class]} as empty. In other words, if a causal razor is not always realizable, it rules out certain joint probability distributions \textit{a priori} as no DAG hypothesis satisfies the \textit{[property]} relative to the distribution. Now we can demarcate the eleven causal razors regarding this property.\\ 

%%%%%%%%%%%%%%%%%%%%
\begin{theorem}
\label{thm:partial_testable}
The following statements are true:
\begin{enumerate}
    \item[(a)] CMC, ori-faithfulness, SGS-minimality, P-minimality, and frugality are always realizable. 
    \item[(b)] CFC, u-P-minimality, res-faithfulness, adj-faithfulness, tri-faithfulness, and u-frugality are not always realizable.
\end{enumerate}
\end{theorem}
\begin{pf}
For (a), $\oriF(\Prob)$ is necessarily non-empty for any joint probability distribution $\Prob$ because a complete DAG contains no unshielded triple. The always-realizability of other causal razors in (a) is obvious. For (b), \textbf{Example \ref{ex:uFr_not_triF}} is the case where $\triF(\Prob) = \varnothing$. \textbf{Example \ref{ex:Fr_not_uFr}} shows the case where $\uFr(\Prob) = \varnothing$. The non-always-realizability of other causal razors in (b) follows from \textbf{Theorem \ref{thm:rz_sb}}.
\end{pf}\\
%%%%%%%%%%%%%%%%%%%%

We now generalize the results from \textit{[class]}'s to \textit{[razor]}'s in Table \ref{tab:rz_pairwise_stru}. Consider any pair of causal razors. Let $\mathbb{r}_1(\cdot)$ and $\mathbb{r}_2(\cdot)$ be their respective \textit{[class]}, and $\textit{razor}_1$ and $\textit{razor}_2$ be their respective \textit{[razor]}. Logical strength of \textit{[razor]}'s can then be formulated as follows:
\begin{enumerate}
    \ib $\textit{razor}_1$ is \textit{logically equivalent} to $\textit{razor}_2$ if $\mathbb{r}_1(\Prob) = \mathbb{r}_2(\Prob)$ for any joint probability distribution $\Prob$.
    \ib $\textit{razor}_1$ is \textit{strictly stronger} than $\textit{razor}_2$ if $\mathbb{r}_1(\Prob) \subseteq \mathbb{r}_2(\Prob)$ holds for any joint probability distribution $\Prob$, but $\mathbb{r}_2(\Prob') \setminus \mathbb{r}_1(\Prob') \neq \varnothing$ for some joint probability distribution $\Prob'$.
    \ib $\textit{razor}_1$ is \textit{logically independent} with $\textit{razor}_2$ if there exist two joint probability distribution $\Prob$ and $\Prob'$ such that $\mathbb{r}_2(\Prob) \setminus \mathbb{r}_1(\Prob) \neq \varnothing$ and $\mathbb{r}_1(\Prob') \setminus \mathbb{r}_2(\Prob') \neq \varnothing$.
\end{enumerate}

We visualize the logical hierarchy over the eleven structural causal razors in Figure \ref{fig:Hier}. To simplify terms, we label each \textit{[razor]} without the word ``causal'', ``condition'', or ``assumption''. Logically equivalent \textit{[razor]}'s are collapsed into a single node. Each arrow indicates that the \textit{[razor]} at the tail-node is strictly stronger than the \textit{[razor]} at the head-node. Any two nodes not connected by a unidirectional path of arrows are logically independent.

%%%%%%%%%%%%%%%%%%%%
\begin{table}[p]
\centering
\resizebox{\textwidth}{!}{
\begin{tabular}{|c|c|c|c|c|c|c|c|c|c|c|}
\hline
& $\CFC$/$\uPm$ & $\resF$ & $\adjF$ & $\oriF$ & $\uFr$ & $\Fr$ & $\Pm$ & $\SGS$ & $\triF$ & $\CMC$\\
\hline
$\CFC$/$\uPm$
& \cellblk 
& \cellcyan \textbf{\ref{thm:rz_sb}} 
& \cellcyan \textbf{\ref{thm:rz_sb}}
& \cellcyan \textbf{\ref{thm:rz_sb}}
& \cellcyan \textbf{\ref{thm:rz_sb}}
& \cellcyan \textbf{\ref{thm:rz_sb}}
& \cellcyan \textbf{\ref{thm:rz_sb}}
& \cellcyan \textbf{\ref{thm:rz_sb}}
& \cellcyan \textbf{\ref{thm:rz_sb}}
& \cellcyan \textbf{\ref{def:faithful}}\\
\hline
$\resF$ 
& \cellmag \textbf{\ref{ex:resF_not_CFC}} 
& \cellblk 
& \cellcyan \textbf{\ref{thm:rz_sb}}
& \cellcyan \textbf{\ref{thm:rz_sb}}
& \cellcyan \textbf{\ref{thm:rz_sb}}
& \cellcyan \textbf{\ref{thm:rz_sb}}
& \cellcyan \textbf{\ref{thm:rz_sb}}
& \cellcyan \textbf{\ref{thm:rz_sb}}
& \cellcyan \textbf{\ref{thm:rz_sb}}
& \cellcyan \textbf{\ref{def:res_faith}}\\
\hline
$\adjF$ 
& \cellmag \textbf{\ref{ex:uFr_not_oriF}}(b)$^\dagger$ 
& \cellmag \textbf{\ref{ex:uFr_not_oriF}}(b)$^\dagger$ 
& \cellblk
& \cellmag \textbf{\ref{ex:uFr_not_oriF}}(b) 
& \cellmag \textbf{\ref{ex:adjF_not_uFr}} 
& \cellcyan \textbf{\ref{thm:rz_sb}}
& \cellcyan \textbf{\ref{thm:rz_sb}}
& \cellcyan \textbf{\ref{thm:rz_sb}}
& \cellcyan \textbf{\ref{thm:rz_sb}}
& \cellcyan \textbf{\ref{def:adj_faith}}\\
\hline
$\oriF$
& \cellmag \textbf{\ref{ex:Fr_not_uFr}}(e)$^\dagger$
& \cellmag \textbf{\ref{ex:Fr_not_uFr}}(e)$^\dagger$
& \cellmag \textbf{\ref{ex:Fr_not_uFr}}(e)$^\dagger$
& \cellblk 
& \cellmag \textbf{\ref{ex:Fr_not_uFr}}(c)
& \cellmag \textbf{\ref{ex:oriF_triF_not_SGSM}}(a)$^\dagger$
& \cellmag \textbf{\ref{ex:oriF_triF_not_SGSM}}(a)$^\dagger$
& \cellmag \textbf{\ref{ex:oriF_triF_not_SGSM}}(a)
& \cellmag \textbf{\ref{ex:Fr_not_uFr}}(e)
& \cellcyan \textbf{\ref{def:ori_faith}}\\
\hline
$\uFr$ 
& \cellmag \textbf{\ref{ex:uFr_not_triF}}(a)$^\dagger$ 
& \cellmag \textbf{\ref{ex:uFr_not_triF}}(a)$^\dagger$ 
& \cellmag \textbf{\ref{ex:uFr_not_triF}}(a)$^\dagger$ 
& \cellmag \textbf{\ref{ex:uFr_not_oriF}}(a) 
& \cellblk 
& \cellcyan \textbf{\ref{def:u-frugal}} 
& \cellcyan \textbf{\ref{thm:rz_sb}}
& \cellcyan \textbf{\ref{thm:rz_sb}}
& \cellmag \textbf{\ref{ex:uFr_not_triF}}(a) 
& \cellcyan \textbf{\ref{def:u-frugal}}\\
\hline
$\Fr$ 
& \cellmag \textbf{\ref{ex:Fr_not_uFr}}(a)$^\dagger$ 
& \cellmag \textbf{\ref{ex:Fr_not_uFr}}(a)$^\dagger$ 
& \cellmag \textbf{\ref{ex:uFr_not_triF}}(a)$^\dagger$ 
& \cellmag \textbf{\ref{ex:uFr_not_oriF}}(a)$^\dagger$ 
& \cellmag \textbf{\ref{ex:Fr_not_uFr}}(a) 
& \cellblk 
& \cellcyan \textbf{\ref{thm:rz_sb}} 
& \cellcyan \textbf{\ref{thm:rz_sb}}
& \cellmag \textbf{\ref{ex:uFr_not_triF}}(a)$^\dagger$ 
& \cellcyan \textbf{\ref{def:frugal}}\\
\hline
$\Pm$ 
& \cellmag \textbf{\ref{ex:Fr_not_uFr}}(a)$^\dagger$
& \cellmag \textbf{\ref{ex:Fr_not_uFr}}(a)$^\dagger$ 
& \cellmag \textbf{\ref{ex:uFr_not_triF}}(a)$^\dagger$ 
& \cellmag \textbf{\ref{ex:uFr_not_oriF}}(a)$^\dagger$ 
& \cellmag \textbf{\ref{ex:Fr_not_uFr}}(a)$^\dagger$ 
& \cellmag \textbf{\ref{ex:uFr_not_oriF}}(c) 
& \cellblk 
& \cellcyan \textbf{\ref{thm:rz_sb}}
& \cellmag \textbf{\ref{ex:uFr_not_triF}}(a)$^\dagger$ 
& \cellcyan \textbf{\ref{def:P-minimal}}\\
\hline
$\SGS$ 
& \cellmag \textbf{\ref{ex:Fr_not_uFr}}(a)$^\dagger$ 
& \cellmag \textbf{\ref{ex:Fr_not_uFr}}(a)$^\dagger$ 
& \cellmag \textbf{\ref{ex:uFr_not_triF}}(a)$^\dagger$ 
& \cellmag \textbf{\ref{ex:uFr_not_oriF}}(a)$^\dagger$ 
& \cellmag \textbf{\ref{ex:Fr_not_uFr}}(a)$^\dagger$ 
& \cellmag \textbf{\ref{ex:uFr_not_oriF}}(c)$^\dagger$ 
& \cellmag \textbf{\ref{ex:CMC_not_SGSM}}(b)
& \cellblk 
& \cellmag \textbf{\ref{ex:uFr_not_triF}}(a)$^\dagger$ 
& \cellcyan \textbf{\ref{def:SGS-minimal}}\\
\hline
$\triF$ 
& \cellmag \textbf{\ref{ex:Fr_not_uFr}}(b)$^\dagger$ 
& \cellmag \textbf{\ref{ex:Fr_not_uFr}}(b)$^\dagger$ 
& \cellmag \textbf{\ref{ex:Fr_not_uFr}}(d) 
& \cellmag \textbf{\ref{ex:uFr_not_oriF}}(b)$^\dagger$ 
& \cellmag \textbf{\ref{ex:Fr_not_uFr}}(b) 
& \cellmag \textbf{\ref{ex:oriF_triF_not_SGSM}}(b)$^\dagger$ 
& \cellmag \textbf{\ref{ex:oriF_triF_not_SGSM}}(b)$^\dagger$ 
& \cellmag \textbf{\ref{ex:oriF_triF_not_SGSM}}(b) 
& \cellblk 
& \cellcyan \textbf{\ref{def:tri_faith}}\\
\hline
$\CMC$ 
& \cellmag \textbf{\ref{ex:Fr_not_uFr}}(a)$^\dagger$ 
& \cellmag \textbf{\ref{ex:Fr_not_uFr}}(a)$^\dagger$ 
& \cellmag \textbf{\ref{ex:uFr_not_triF}}(a)$^\dagger$ 
& \cellmag \textbf{\ref{ex:uFr_not_oriF}}(a)$^\dagger$ & \cellmag \textbf{\ref{ex:Fr_not_uFr}}(a)$^\dagger$ 
& \cellmag \textbf{\ref{ex:uFr_not_oriF}}(c)$^\dagger$ 
& \cellmag \textbf{\ref{ex:uFr_not_oriF}}(a)$^\dagger$ 
& \cellmag \textbf{\ref{ex:CMC_not_SGSM}}(a) 
& \cellmag \textbf{\ref{ex:uFr_not_triF}}(a)$^\dagger$ 
& \cellblk\\
\hline
\end{tabular}}
    \caption{\textit{Pairwise comparisons of [class]'s from eleven structural causal razors.} Each cell concerns whether the \textit{[class]} in the row is a subset of the \textit{[class]} in the column. Cells on the diagonal are filled with black to indicate their trivial truth. Cells filled with {\color{cyan}{\text{cyan}}} indicate that the subset relation is true either by \textbf{Theorem \ref{thm:rz_sb}} or by definition. Cells filled with {\color{magenta}{\text{magenta}}} refer to the counterexample of the subset relation in \textbf{Example \ref{ex:resF_not_CFC}} - \textbf{\ref{ex:oriF_triF_not_SGSM}}. The superscript $\dagger$ indicates an application of \textbf{Fact \ref{fact:dagger}} with \textbf{Theorem \ref{thm:rz_sb}}.}
    \label{tab:rz_pairwise_stru}
\end{table}
%%%%%%%%%%%%%%%%%%%%

%%%%%%%%%%%%%%%%%%%%
\begin{figure}[p]
\centering
\begin{tikzpicture}
%%%%%%%%%%
\node (a) at (0,0) {$\bullet$};
\draw [decorate,
	decoration = {brace}] (-0.5,0.5) --  (-0.5,-0.5);
\node(a1) at (-1.7,0.25) {faithfulness};
\node(a2) at (-1.9,-0.25) {u-P-minimality};
%%%%%%%%%%
\node (b) at (0,-2) {$\bullet$};
\node (b1) at (-1.7, -1.75) {res-faithfulness};
%%%%%%%%%%
\path [->,line width=0.5mm] (a) edge (b);
%%%%%%%%%%
\node (c) at (5, -7) {$\bullet$};
\node (c1) at (6.5, -7) {ori-faithfulness};
%%%%%%%%%%
\path [->, line width=0.5mm, bend left] (b) edge (c);
%%%%%%%%%%
\node (d) at (-4,-3) {$\bullet$};
\node (d1) at (-5.5,-3) {adj-faithfulness};
%%%%%%%%%%
\path [->,line width=0.5mm] (b) edge (d);
%%%%%%%%%%
\node (e) at (0,-4) {$\bullet$};
\node (e1) at (1.1,-4) {u-frugality};
%%%%%%%%%%
\path [->,line width=0.5mm] (b) edge (e);
%%%%%%%%%%
\node (f) at (0, -6) {$\bullet$};
\node (f1) at (1, -6) {frugality};
%%%%%%%%%%
\path [->,line width=0.5mm] (d) edge (f);
\path [->,line width=0.5mm] (e) edge (f);
%%%%%%%%%%
\node (g) at (0,-8) {$\bullet$};
\node (g1) at (1.3,-8) {P-minimality};
%%%%%%%%%%
\path [->,line width=0.5mm] (f) edge (g);
%%%%%%%%%%
\node (h) at (0,-10) {$\bullet$};
\node (h1) at (1.6,-10) {SGS-minimality};
%%%%%%%%%%
\path [->,line width=0.5mm] (g) edge (h);
%%%%%%%%%%
\node (i) at (-4,-4) {$\bullet$};
\node (i1) at (-5.5,-4) {tri-faithfulness};
%%%%%%%%%%
\path [->, line width=0.5mm] (d) edge (i);
%%%%%%%%%%
\node (j) at (0,-12) {$\bullet$};
\node (j1) at (-1,-12) {Markov};
\node (empty) at (-1, -12.5) {};
%%%%%%%%%%
\path [->,line width=0.5mm] (h) edge (j);
\path [->,line width=0.5mm] (i) edge (j);
\path [->, line width=0.5mm, bend left] (c) edge (j);
%%%%%%%%%%
\draw[dotted] (-7,-5) -- (9, -5);
\node (pt) at (7.2, -4.7) {\color{gray}{\textit{not always realizable}}};
\node (pt) at (7.5, -5.3) {\color{gray}{\textit{always realizable}}};
\end{tikzpicture}
\caption{\textit{Logical hierarchy of the eleven structural causal razors}}
\label{fig:Hier}
\end{figure}
%%%%%%%%%%%%%%%%%%%%

We end this section by noting the possibility of extending this comparative analysis. First, as mentioned earlier, we can incorporate the recently proposed causal razors to extend the list. Second, pairwise comparisons are not limited to the study of subset relations. \cite{Zhalama2019asp} introduced the relation of \textit{conservative weakening} to study how a weaker causal razor can be reasonably assumed when the stronger causal razor is satisfied.\footnote{Using our language, if \textit{razor}$_1$ is strictly stronger than \textit{razor}$_2$, and $\mathbb{r}_1(\Prob) = \mathbb{r}_2(\Prob)$ for any joint probability distribution $\Prob$ that \textit{razor}$_1$ is satisfied, then \textit{razor}$_2$ is a conservative weakening of \textit{razor}$_1$. For instance, frugality is a conservative weakening of u-frugality.} Thirdly, causal razors can be studied \textit{triple-wise}. One instance is the result in \citep{zhang2013comparison} that the conjunction of SGS-minimality and tri-faithfulness entails P-minimality. Hence, interested readers can elongate our comparative analysis above to an even fuller extent. In the next section, we stretch our current picture by introducing a causal razor of a different kind: \textit{parameter minimality}.\\

%%%%%%%%%%%%%%%%%%%%%%%%%%%%%%%%%%%%%%%%%%%%%%%%%%%%%%%%%%%%%%%%%%%%%%
%%%%%%%%%%%%%%%%%%%%%%%%%%%%%%%%%%%%%%%%%%%%%%%%%%%%%%%%%%%%%%%%%%%%%%
%%%%%%%%%%%%%%%%%%%%%%%%%%%%%%%%%%%%%%%%%%%%%%%%%%%%%%%%%%%%%%%%%%%%%%
\section{Parameter Minimality}
\label{sec:paramM}
%%%%%%%%%%%%%%%%%%%%%%%%%%%%%%%%%%%%%%%%%%%%%%%%%%%%%%%%%%%%%%%%%%%%%%
%%%%%%%%%%%%%%%%%%%%%%%%%%%%%%%%%%%%%%%%%%%%%%%%%%%%%%%%%%%%%%%%%%%%%%
%%%%%%%%%%%%%%%%%%%%%%%%%%%%%%%%%%%%%%%%%%%%%%%%%%%%%%%%%%%%%%%%%%%%%%

This section particularly concerns a pair of causal razors defined in terms of the number of parameters of a causal model. Readers can quickly realize that the task of defining them is misaligned with the vague classification of causal razors in \textbf{Section \ref{sec:common_rz}}. First, they are not purely parametric since we need to make use of the graphical features of a given DAG $\G$ to compute $|\param(\G)|$. On the other hand, they are not purely structural in that one needs to specify the parametric assumption of the joint probability distribution to obtain $\param(\G)$. This semi-structural property might plausibly explain why parameter minimality has not received much attention in the literature.

Nevertheless, the number of parameters is a major metric in measuring statistical complexity of a model in most, if not all, modeling methods in statistics and computer science. Furthermore, this concept is tightly connected to causal search algorithms. For example, a \textit{consistent scoring criterion}, as in \citep{haughton1988choice} and will be further discussed in the next section, is popularly utilized by researchers to evaluate how a DAG hypothesis fits a given observational dataset. In short, a consistent scoring criterion (e.g., \textit{Bayesian information criterion} in \citep{schwarz1978estimating}) demands that Markovian DAGs with fewer parameters should receive a higher score. So, it is not hard to see that causal razors defined by the number of parameters do have some reasonable ground. Without further ado, below are the formal definitions of parameter minimality and its unique variant.\\

%%%%%%%%%%%%%%%%%%%%
\noindent\fbox{\parbox{\textwidth}{
\begin{definition}
\label{def:ParamMin}
(Parameter minimality) For any joint probability distribution $\Prob$,
\begin{enumerate}
    \ib [class] $\ParamM(\Prob) := \{\G \in \CMC(\Prob): \neg \exists \G' \in \CMC(\Prob)$ s.t. $|\param(\G')| < |\param(\G)|\}$;
    \ib [property] a DAG $\G$ is param-minimal (to $\Prob$) if $\G \in \ParamM(\Prob)$;
    \ib [razor] $(\G^*, \Prob)$ satisfies the causal param-minimality assumption if $\G^* \in \ParamM(\Prob)$.
\end{enumerate}
\end{definition}}}
\,\\
%%%%%%%%%%%%%%%%%%%%

%%%%%%%%%%%%%%%%%%%%
\noindent\fbox{\parbox{\textwidth}{
\begin{definition}
\label{def:u-ParamMin}
(Unique parameter minimality) For any joint probability distribution $\Prob$,
\begin{enumerate}
    \ib [class] $\uParamM(\Prob) := $
    $\begin{cases}
    \ParamM(\Prob) & \text{if all DAGs in $\ParamM(\Prob)$ belong to the same MEC}\\
    \varnothing & \text{otherwise; }
    \end{cases}$
    \ib [property] a DAG $\G$ is u-param-minimal (to $\Prob$) if $\G \in \uParamM(\Prob)$;
    \ib [razor] $(\G^*, \Prob)$ satisfies the causal u-param-minimality assumption if $\G^* \in \uParamM(\Prob)$.
\end{enumerate}
\end{definition}}}
\,\\
%%%%%%%%%%%%%%%%%%%%

With a closer look, readers can easily observe the affinity of these two causal razors with frugality and u-frugality defined in \textbf{Section \ref{sec:common_rz}}. They are similar in every aspect except that the pair above is defined in terms of number of parameters instead of number of edges. For linear Gaussian causal models, the difference between the two notions dissipates due to the equality $|\param(\G)| = |\E(\G)|$ for any DAG $\G$ . But the equality fails in the context of multinomial causal models.

By focusing only on multinomial causal models, how do param-minimality and its unique variant relate to the other causal razors discussed in \textbf{Section \ref{sec:common_rz}}? We first show some positive claims regarding the subset relations of their \textit{[class]}'s. In particular, we are going to utilize the famous result in \citep{Chickering1995} and \citep{chickering2002optimal} related to \textit{covered edge reversals} to show that $\ParamM(\Prob) \subseteq \Pm(\Prob)$ for any joint multinomial distribution $\Prob$.

%%%%%%%%%%%%%%%%%%%%
\begin{theorem}
\label{thm:ParamM_subset_PM}
For any joint multinomial distribution $\Prob$, $\ParamM(\Prob) \subseteq \Pm(\Prob)$ holds.
\end{theorem}
\begin{pf}
See \textbf{Appendix \ref{app:ParamM_Pm}}.
\end{pf}\\
%%%%%%%%%%%%%%%%%%%%

By $\CFC(\Prob) = \uPm(\Prob)$ in \textbf{Theorem \ref{CFC-uPm}}, one can easily derive that $\CFC(\Prob) \subseteq \uParamM(\Prob)$. Nonetheless, a more interesting question is whether the stronger claim $\resF(\Prob) \subseteq \uParamM(\Prob)$ holds. We prove the truth of this claim by utilizing the concept \textit{parameterizing sets} from \citep{Hemmecke_imset}. In short, we show that a restricted-faithful DAG must have a minimal set of parameterizing sets, and it necessitates that the number of parameters for that DAG is the smallest over all Markovian DAGs.\\

%%%%%%%%%%%%%%%%%%%%
\begin{theorem}
\label{thm:resF_subset_uParamM}
For any joint multinomial distribution $\Prob$, $\resF(\Prob) \subseteq \uParamM(\Prob)$ holds.
\end{theorem}
\begin{pf}
See \textbf{Appendix \ref{app:ResF_uParamM}}.
\end{pf}\\
%%%%%%%%%%%%%%%%%%%%

%%%%%%%%%%%%%%%%%%%%
\begin{corollary}
\label{coro:rz_sb_MN}
For any joint multinomial distribution $\Prob$, the following statement is true:
\begin{align*}
    \CFC(\Prob) \subseteq \resF(\Prob) = \adjF(\Prob) \cap \oriF(\Prob) \subseteq \uParamM(\Prob) \subseteq \ParamM(\Prob) \subseteq \Pm(\Prob) \subseteq \SGS(\Prob) \subseteq \CMC(\Prob).
\end{align*}
\end{corollary}
\,\\
\vspace{-0.5cm}
%%%%%%%%%%%%%%%%%%%%

So much for the positive results. We now turn to some counterexamples of each subset relation $\mathbb{r}_1(\cdot) \subseteq \mathbb{r}_2(\cdot)$ where one of $\mathbb{r}_1$ and $\mathbb{r}_2$ is the \textit{[class]} of (u-)param-minimality. Each counterexample is from the last section where no parametric assumption was involved. Due to the current context, we need to specify how the underlying multinomial distribution can be constructed parametrically. To simplify our exposition, we leave all construction details in \textbf{Appendix \ref{app:construction}}.\\

%%%%%%%%%%%%%%%%%%%%
\begin{example}
\label{ex:adjF_not_paramM}
There exists a joint multinomial distribution $\Prob$ such that $\adjF(\Prob) \setminus \ParamM(\Prob) \neq \varnothing$.
\end{example}
\noindent [Same as \textbf{Example \ref{ex:adjF_not_uFr}} with \text{Figure} \ref{fig:adjF_not_uFr}. See \textbf{Appendix \ref{app:ex_adjF_not_paramM}} for details.]
\vspace{0.2cm}

\noindent First, $\G_1 \in \adjF(\Prob)$ clearly holds. Given that $\mb{V}$ is ranged from $\la2, 3, 2\ra$, we have $|\param(\G_0)| = 8 > 10 = |\param(\G_1)|$ such that $\G_1 \notin \ParamM(\Prob)$.\hfill $\square$\\
%%%%%%%%%%%%%%%%%%%%

%%%%%%%%%%%%%%%%%%%%
\begin{example}
\label{ex:ParamM_not_uParamM}
There exists a joint multinomial distribution $\Prob$ such that
\begin{enumerate}
    \item[(a)] $\ParamM(\Prob) \setminus \uParamM(\Prob) \neq \varnothing$;
    \item[(b)] $\triF(\Prob) \setminus \uParamM(\Prob) \neq \varnothing$;
    \item[(c)] $\oriF(\Prob) \setminus \uParamM(\Prob) \neq \varnothing$.
\end{enumerate}
\end{example}
\noindent [Same as \textbf{Example \ref{ex:Fr_not_uFr}} with \text{Figure} \ref{fig:Fr_not_uFr}. See \textbf{Appendix \ref{app:ex_ParamM_not_uParamM}} for details.]
\vspace{0.2cm}

\noindent As specified in the construction, $\mb{V}$ is ranged from $\la 2, 2, 2, 3\ra$. We have $|\param(\G^*)| = |\param(\G')| = 13$. Readers can verify that $\G^*, \G' \in \ParamM(\Prob)$. Thus, $\uParamM(\Prob) = \varnothing$. As explained in \textbf{Example \ref{ex:Fr_not_uFr}}, we also have $\G^* \in \oriF(\Prob) \cap \triF(\Prob)$.\hfill $\square$\\
%%%%%%%%%%%%%%%%%%%%

%%%%%%%%%%%%%%%%%%%%
\begin{corollary}
u-param-minimality is not always realizable.\\
\end{corollary}
%%%%%%%%%%%%%%%%%%%%

%%%%%%%%%%%%%%%%%%%%
\begin{example}
\label{ex:uParam_not_oriF}
There exists a joint multinomial distribution $\Prob$ such that 
\begin{enumerate}
    \item[(a)] $\uParamM(\Prob) \setminus \oriF(\Prob) \neq \varnothing$;
    \item[(b)] $\Pm(\Prob) \setminus \ParamM(\Prob) \neq \varnothing$.
\end{enumerate}
\end{example}
\noindent [Same as \textbf{Example \ref{ex:uFr_not_oriF}} with \text{Figure} \ref{fig:uFr_not_oriF}. See \textbf{Appendix \ref{app:ex_uParam_not_oriF}} for details.]
\vspace{0.2cm}

\noindent Consider the case that every variable in $\mb{V}$ is binary. Readers can verify that $\G^* \in \uParamM(\Prob)$ (where $|\param(\G^*)| = 9)$ and $\G' \notin \ParamM(\Prob)$ (where $|\param(\G')| = 13$). See \textbf{Example \ref{ex:uFr_not_oriF}} for an explanation of $\G^* \notin \oriF(\Prob)$ and $\G' \in \Pm(\Prob)$. \hfill $\square$\\
%%%%%%%%%%%%%%%%%%%%

%%%%%%%%%%%%%%%%%%%%
\begin{example}
\label{ex:uFr_not_uParamM}
There exists a joint multinomial distribution $\Prob$ such that
\begin{enumerate}
\item[(a)] $\uParamM(\Prob) \setminus \triF(\Prob) \neq \varnothing$;
\item[(b)] $\uFr(\Prob) \setminus \ParamM(\Prob) \neq \varnothing$;
\item[(c)] $\uParamM(\Prob) \setminus \Fr(\Prob) \neq \varnothing$.
\end{enumerate}
\end{example}
\noindent [Similar to \textbf{Example \ref{ex:uFr_not_triF}}. See \textbf{Appendix \ref{app:ex_uFr_not_uParamM}} for details.]\\
\vspace{-0.2cm}

\noindent Consider $\G_0$ and $\G_1$ in \text{Figure} \ref{fig:uFr_not_uParamM} that are both Markovian to $\Prob$. As specified by the construction,$\mb{V}$ is ranged from $\la 2, 2, 2, 2, 3\ra$. As explained in \textbf{Example \ref{ex:uFr_not_triF}}, we know that $\G_0 \in \uFr(\Prob)$. However, we have $|\param(\G_0)| = 37 < 35 = |\param(\G_1)|$. Indeed, readers can verify that $\uParamM(\Prob) = \{\G_1\}$. Hence, $\G_0 \in \uFr(\Prob) \setminus \ParamM(\Prob)$ whereas $\G_1 \in \uParamM(\Prob) \setminus \Fr(\Prob)$. Lastly, $\G_1 \notin \triF(\Prob)$ because of the triangle $(3, 5, 4)$  in $\G'$ and $\la X_3, X_4\ra \in \Psi_{\G_1, \Prob}$.\hfill $\square$\\
%%%%%%%%%%%%%%%%%%%%

%%%%%%%%%%%%%%%%%%%%
\begin{figure}[h]
\centering
\subfloat{
\begin{tikzpicture}
\node (X1) at (0.0,1.0) {$1$};
\node (X2) at (0.9510565162951535,0.30901699437494745) {$2$};
\node (X3) at (0.5877852522924732,-0.8090169943749473) {$3$};
\node (X4) at (-0.587785252292473,-0.8090169943749476) {$4$};
\node (X5) at (-0.9510565162951536,0.30901699437494723) {$5$};
\node (label) at (1.7, -0.4) {$\G_0$};
\path [->,line width=0.5mm] (X1) edge (X5);
\path [->,line width=0.5mm] (X2) edge (X5);
\path [->,line width=0.5mm] (X3) edge (X5);
\path [->,line width=0.5mm] (X4) edge (X5);
\path [->,line width=0.5mm] (X1) edge (X4);
\end{tikzpicture}}
\hspace{3cm}
\subfloat{
\begin{tikzpicture}
\node (X1) at (0.0,1.0) {$1$};
\node (X2) at (0.9510565162951535,0.30901699437494745) {$2$};
\node (X3) at (0.5877852522924732,-0.8090169943749473) {$3$};
\node (X4) at (-0.587785252292473,-0.8090169943749476) {$4$};
\node (X5) at (-0.9510565162951536,0.30901699437494723) {$5$};
\node (label) at (1.7, -0.4) {$\G_1$};
\path [->,line width=0.5mm] (X1) edge (X4);
\path [->,line width=0.5mm] (X2) edge (X4);
\path [->,line width=0.5mm] (X2) edge (X5);
\path [->,line width=0.5mm] (X3) edge (X4);
\path [->,line width=0.5mm] (X3) edge (X5);
\path [->,line width=0.5mm] (X5) edge (X4);
\end{tikzpicture}}
\caption{$\G_0$ is u-frugal but not param-minimal, whereas $\G_1$ is u-param-minimal but not frugal.}
\label{fig:uFr_not_uParamM}
\end{figure}
%%%%%%%%%%%%%%%%%%%%

Table \ref{tab:rz_pairwise_all} and \ref{fig:Hier_MN} extend the results in \textbf{Section \ref{sec:rz_hier}} by incorporating param-minimality and u-param-minimality in the multinomial context. Particularly, the disagreement between (u-)frugality and (u-)param-minimality uncovers a quandary in causal search algorithms: should one count edges or parameters to \textit{score} a DAG? This methodological problem will be studied in the next section.\\

%%%%%%%%%%%%%%%%%%%%
\begin{table}[p]
\centering
\resizebox{\textwidth}{!}{%
\begin{tabular}{|c|c|c|c|c|c|c|c|c|c|c|c|c|}
\hline
& $\CFC/\uPm$ & $\resF$ & $\adjF$ & $\oriF$ & $\uFr$ & $\Fr$ & $\uParamM$ & $\ParamM$ & $\Pm$ & $\SGS$ & $\triF$ & $\CMC$ 
\\\hline
$\CFC/\uPm$ 
& \cellblk
& \textbf{\ref{thm:rz_sb}} \cellcyan
& \textbf{\ref{thm:rz_sb}} \cellcyan
& \textbf{\ref{thm:rz_sb}} \cellcyan
& \textbf{\ref{thm:rz_sb}} \cellcyan
& \textbf{\ref{thm:rz_sb}} \cellcyan
& \textbf{\ref{coro:rz_sb_MN}} \cellcyan
& \textbf{\ref{coro:rz_sb_MN}} \cellcyan
& \textbf{\ref{thm:rz_sb}} \cellcyan
& \textbf{\ref{thm:rz_sb}} \cellcyan
& \textbf{\ref{thm:rz_sb}} \cellcyan
& \textbf{\ref{def:faithful}} \cellcyan
\\\hline
$\resF$ 
& \textbf{\ref{ex:resF_not_CFC}} \cellmag
& \cellblk
& \textbf{\ref{thm:rz_sb}} \cellcyan
& \textbf{\ref{thm:rz_sb}} \cellcyan
& \textbf{\ref{thm:rz_sb}} \cellcyan
& \textbf{\ref{thm:rz_sb}} \cellcyan
& \textbf{\ref{coro:rz_sb_MN}} \cellcyan
& \textbf{\ref{coro:rz_sb_MN}} \cellcyan
& \textbf{\ref{thm:rz_sb}} \cellcyan
& \textbf{\ref{thm:rz_sb}} \cellcyan
& \textbf{\ref{thm:rz_sb}} \cellcyan
& \textbf{\ref{def:res_faith}} \cellcyan
\\\hline
$\adjF$ 
& \textbf{\ref{ex:uFr_not_oriF}}(b)$^\dagger$ \cellmag
& \textbf{\ref{ex:uFr_not_oriF}}(b)$^\dagger$ \cellmag
& \cellblk
& \textbf{\ref{ex:uFr_not_oriF}}(b) \cellmag
& \textbf{\ref{ex:adjF_not_uFr}} \cellmag
& \textbf{\ref{thm:rz_sb}} \cellcyan
& \textbf{\ref{ex:adjF_not_paramM}}(a)$^\dagger$ \cellmag
& \textbf{\ref{ex:adjF_not_paramM}}(a) \cellmag
& \textbf{\ref{thm:rz_sb}}\cellcyan
& \textbf{\ref{thm:rz_sb}}\cellcyan
& \textbf{\ref{thm:rz_sb}}\cellcyan
& \textbf{\ref{def:adj_faith}}\cellcyan
\\\hline
$\oriF$ 
& \textbf{\ref{ex:Fr_not_uFr}}(e)$^\dagger$ \cellmag
& \textbf{\ref{ex:Fr_not_uFr}}(e)$^\dagger$ \cellmag
& \textbf{\ref{ex:Fr_not_uFr}}(e)$^\dagger$ \cellmag
& \cellblk
& \textbf{\ref{ex:Fr_not_uFr}}(c) \cellmag
& \textbf{\ref{ex:oriF_triF_not_SGSM}}(a)$^\dagger$ \cellmag
& \textbf{\ref{ex:oriF_triF_not_SGSM}}(a)$^\dagger$ \cellmag
& \textbf{\ref{ex:oriF_triF_not_SGSM}}(a)$^\dagger$ \cellmag
& \textbf{\ref{ex:oriF_triF_not_SGSM}}(a)$^\dagger$ \cellmag
& \textbf{\ref{ex:oriF_triF_not_SGSM}}(a) \cellmag
& \textbf{\ref{ex:Fr_not_uFr}}(e) \cellmag
& \textbf{\ref{def:ori_faith}} \cellcyan
\\\hline
$\uFr$
& \textbf{\ref{ex:uFr_not_triF}}(a)$^\dagger$ \cellmag
& \textbf{\ref{ex:uFr_not_triF}}(a)$^\dagger$ \cellmag
& \textbf{\ref{ex:uFr_not_triF}}(a)$^\dagger$ \cellmag
& \textbf{\ref{ex:uFr_not_oriF}}(a) \cellmag
& \cellblk
& \textbf{\ref{def:u-frugal}} \cellcyan
& \textbf{\ref{ex:uFr_not_uParamM}}(b)$^\dagger$ \cellmag
& \textbf{\ref{ex:uFr_not_uParamM}}(b) \cellmag
& \textbf{\ref{thm:rz_sb}} \cellcyan
& \textbf{\ref{thm:rz_sb}} \cellcyan
& \textbf{\ref{ex:uFr_not_triF}}(a) \cellmag
& \textbf{\ref{def:u-frugal}} \cellcyan
\\\hline
$\Fr$
& \textbf{\ref{ex:Fr_not_uFr}}(a)$^\dagger$ \cellmag
& \textbf{\ref{ex:Fr_not_uFr}}(a)$^\dagger$ \cellmag
& \textbf{\ref{ex:uFr_not_triF}}(a)$^\dagger$ \cellmag
& \textbf{\ref{ex:uFr_not_oriF}}(a)$^\dagger$ \cellmag
& \textbf{\ref{ex:Fr_not_uFr}}(a) \cellmag
& \cellblk
& \textbf{\ref{ex:uFr_not_uParamM}}(b)$^\dagger$ \cellmag
& \textbf{\ref{ex:uFr_not_uParamM}}(b)$^\dagger$ \cellmag
& \textbf{\ref{thm:rz_sb}} \cellcyan
& \textbf{\ref{thm:rz_sb}} \cellcyan
& \textbf{\ref{ex:uFr_not_triF}}(a)$^\dagger$ \cellmag
& \textbf{\ref{def:frugal}} \cellcyan
\\\hline
$\uParamM$ 
& \textbf{\ref{ex:uFr_not_uParamM}}(a)$^\dagger$ \cellmag
& \textbf{\ref{ex:uFr_not_uParamM}}(a)$^\dagger$ \cellmag
& \textbf{\ref{ex:uFr_not_uParamM}}(a)$^\dagger$ \cellmag
& \textbf{\ref{ex:uParam_not_oriF}}(a) \cellmag
& \textbf{\ref{ex:uFr_not_uParamM}}(c)$^\dagger$ \cellmag
& \textbf{\ref{ex:uFr_not_uParamM}}(c) \cellmag
& \cellblk
& \textbf{\ref{def:u-ParamMin}} \cellcyan
& \textbf{\ref{coro:rz_sb_MN}} \cellcyan
& \textbf{\ref{coro:rz_sb_MN}} \cellcyan
& \textbf{\ref{ex:uFr_not_uParamM}}(a) \cellmag
& \textbf{\ref{def:u-ParamMin}} \cellcyan
\\\hline
$\ParamM$ 
& \textbf{\ref{ex:uFr_not_uParamM}}(a)$^\dagger$ \cellmag
& \textbf{\ref{ex:uFr_not_uParamM}}(a)$^\dagger$ \cellmag
& \textbf{\ref{ex:uFr_not_uParamM}}(a)$^\dagger$ \cellmag
& \textbf{\ref{ex:uParam_not_oriF}}(a) \cellmag
& \textbf{\ref{ex:uFr_not_uParamM}}(c)$^\dagger$ \cellmag
& \textbf{\ref{ex:uFr_not_uParamM}}(c)$^\dagger$ \cellmag
& \textbf{\ref{ex:ParamM_not_uParamM}}(a) \cellmag
& \cellblk
& \textbf{\ref{coro:rz_sb_MN}} \cellcyan
& \textbf{\ref{coro:rz_sb_MN}} \cellcyan
& \textbf{\ref{ex:uFr_not_uParamM}}(a)$^\dagger$ \cellmag
& \textbf{\ref{def:ParamMin}} \cellcyan
\\\hline
$\Pm$ 
& \textbf{\ref{ex:Fr_not_uFr}}(a)$^\dagger$ \cellmag
& \textbf{\ref{ex:Fr_not_uFr}}(a)$^\dagger$ \cellmag
& \textbf{\ref{ex:uFr_not_triF}}(a)$^\dagger$ \cellmag
& \textbf{\ref{ex:uFr_not_oriF}}(a)$^\dagger$ \cellmag
& \textbf{\ref{ex:Fr_not_uFr}}(a)$^\dagger$ \cellmag
& \textbf{\ref{ex:uFr_not_oriF}}(c) \cellmag
& \textbf{\ref{ex:uParam_not_oriF}}(b)$^\dagger$ \cellmag
& \textbf{\ref{ex:uParam_not_oriF}}(b) \cellmag
& \cellblk
& \textbf{\ref{thm:rz_sb}} \cellcyan
& \textbf{\ref{ex:uFr_not_triF}}$^\dagger$ \cellmag
& \textbf{\ref{def:P-minimal}} \cellcyan
\\\hline
$\SGS$ 
& \textbf{\ref{ex:Fr_not_uFr}}(a)$^\dagger$ \cellmag
& \textbf{\ref{ex:Fr_not_uFr}}(a)$^\dagger$ \cellmag
& \textbf{\ref{ex:uFr_not_triF}}(a)$^\dagger$ \cellmag
& \textbf{\ref{ex:uFr_not_oriF}}(a)$^\dagger$ \cellmag
& \textbf{\ref{ex:Fr_not_uFr}}(a)$^\dagger$ \cellmag
& \textbf{\ref{ex:uFr_not_oriF}}(c)$^\dagger$ \cellmag
& \textbf{\ref{ex:uParam_not_oriF}}(b)$^\dagger$ \cellmag
& \textbf{\ref{ex:uParam_not_oriF}}(b)$^\dagger$ \cellmag
& \textbf{\ref{ex:CMC_not_SGSM}}(b) \cellmag
& \cellblk
& \textbf{\ref{ex:uFr_not_triF}}(a)$^\dagger$ \cellmag
& \textbf{\ref{def:SGS-minimal}} \cellcyan
\\\hline
$\triF$ 
& \textbf{\ref{ex:Fr_not_uFr}}(b)$^\dagger$ \cellmag
& \textbf{\ref{ex:Fr_not_uFr}}(b)$^\dagger$ \cellmag
& \textbf{\ref{ex:Fr_not_uFr}}(d) \cellmag
& \textbf{\ref{ex:uFr_not_oriF}}(b)$^\dagger$ \cellmag
& \textbf{\ref{ex:Fr_not_uFr}}(b) \cellmag
& \textbf{\ref{ex:oriF_triF_not_SGSM}}(b)$^\dagger$ \cellmag
& \textbf{\ref{ex:ParamM_not_uParamM}}(b) \cellmag
& \textbf{\ref{ex:oriF_triF_not_SGSM}}(b)$^\dagger$ \cellmag
& \textbf{\ref{ex:oriF_triF_not_SGSM}}(b)$^\dagger$ \cellmag
& \textbf{\ref{ex:oriF_triF_not_SGSM}}(b) \cellmag
& \cellblk
& \textbf{\ref{def:tri_faith}} \cellcyan
\\\hline
$\CMC$ 
& \textbf{\ref{ex:Fr_not_uFr}}(a)$^\dagger$ \cellmag
& \textbf{\ref{ex:Fr_not_uFr}}(a)$^\dagger$ \cellmag
& \textbf{\ref{ex:uFr_not_triF}}(a)$^\dagger$ \cellmag
& \textbf{\ref{ex:uFr_not_oriF}}(a)$^\dagger$ \cellmag
& \textbf{\ref{ex:Fr_not_uFr}}(a)$^\dagger$ \cellmag
& \textbf{\ref{ex:uFr_not_oriF}}(c)$^\dagger$ \cellmag
& \textbf{\ref{ex:uParam_not_oriF}}(b)$^\dagger$ \cellmag
& \textbf{\ref{ex:uParam_not_oriF}}(b)$^\dagger$ \cellmag
& \textbf{\ref{ex:uFr_not_oriF}}(a)$^\dagger$ \cellmag
& \textbf{\ref{ex:CMC_not_SGSM}}(a) \cellmag
& \textbf{\ref{ex:uFr_not_triF}}(a)$^\dagger$ \cellmag
& \cellblk
\\\hline
\end{tabular}}
    \caption{\textit{Pairwise comparisons of [class]'s from thirteen causal razors.} $\ParamM$ and $\uParamM$ are defined for multinomial causal models. See the caption of \textsc{Table} \ref{tab:rz_pairwise_stru} for the coloring convention. The superscript $\dagger$ indicates an application of \textbf{Fact \ref{fact:dagger}} with \textbf{Corollary \ref{coro:rz_sb_MN}} or \textbf{Theorem \ref{thm:rz_sb}}.}
    \label{tab:rz_pairwise_all}
\end{table}

%%%%%%%%%%%%%%%%%%%%
\begin{figure}[p]
\centering
\begin{tikzpicture}
%%%%%%%%%%
\node (a) at (0,0) {$\bullet$};
\draw [decorate,
	decoration = {brace}] (-0.5,0.5) --  (-0.5,-0.5);
\node(a1) at (-1.7,0.25) {faithfulness};
\node(a2) at (-1.9,-0.25) {u-P-minimality};
%%%%%%%%%%
\node (b) at (0,-2) {$\bullet$};
\node (b1) at (-1.7, -1.75) {res-faithfulness};
%%%%%%%%%%
\path [->,line width=0.5mm] (a) edge (b);
%%%%%%%%%%
\node (c) at (5, -6.4) {$\bullet$};
\node (c1) at (6.5, -6.4) {ori-faithfulness};
%%%%%%%%%%
\path [->, line width=0.5mm, bend left] (b) edge (c);
%%%%%%%%%%
\node (d) at (-5,-3) {$\bullet$};
\node (d1) at (-6.5,-3) {adj-faithfulness};
%%%%%%%%%%
\path [->,line width=0.5mm] (b) edge (d);
%%%%%%%%%%
\node (e) at (-3,-4) {$\bullet$};
\node (e1) at (-1.9,-4.1) {u-frugality};
%%%%%%%%%%
\path [->,line width=0.5mm] (b) edge (e);
%%%%%%%%%%
\node (f) at (-3, -6) {$\bullet$};
\node (f1) at (-2, -5.9) {frugality};
%%%%%%%%%%
\node (up) at (0,-4) {$\bullet$};
\node (up1) at (1.85,-4.1) {u-param-minimality};
%%%%%%%%%%
\path [->,line width=0.5mm] (b) edge (up);
%%%%%%%%%%
\node (p) at (0, -6) {$\bullet$};
\node (p1) at (1.7, -5.9) {param-minimality};
%%%%%%%%%%
\path [->,line width=0.5mm] (up) edge (p);
%%%%%%%%%%
\path [->,line width=0.5mm] (d) edge (f);
\path [->,line width=0.5mm] (e) edge (f);
%%%%%%%%%%
\node (g) at (0,-8) {$\bullet$};
\node (g1) at (1.3,-8) {P-minimality};
%%%%%%%%%%
\path [->,line width=0.5mm] (f) edge (g);
\path [->,line width=0.5mm] (p) edge (g);
%%%%%%%%%%
\node (h) at (0,-10) {$\bullet$};
\node (h1) at (1.6,-10) {SGS-minimality};
%%%%%%%%%%
\path [->,line width=0.5mm] (g) edge (h);
%%%%%%%%%%
\node (i) at (-5,-4.5) {$\bullet$};
\node (i1) at (-6.5,-4.5) {tri-faithfulness};
%%%%%%%%%%
\path [->, line width=0.5mm] (d) edge (i);
%%%%%%%%%%
\node (j) at (0,-12) {$\bullet$};
\node (j1) at (-1,-12) {Markov};
\node (empty) at (-1, -12.5) {};
%%%%%%%%%%
\path [->,line width=0.5mm] (h) edge (j);
\path [->,line width=0.5mm] (i) edge (j);
\path [->, line width=0.5mm, bend left] (c) edge (j);
%%%%%%%%%%
\draw[dotted] (-8.5,-5) -- (9, -5);
\node (pt) at (6.7, -4.7) {\color{gray}{\textit{not always realizable}}};
\node (pt) at (7, -5.3) {\color{gray}{\textit{always realizable}}};
\end{tikzpicture}
\caption{\textit{Logical hierarchy of causal razors with (u-)param-minimality defined for multinomial causal models.}}
\label{fig:Hier_MN}
\end{figure}
%%%%%%%%%%%%%%%%%%%%

%%%%%%%%%%%%%%%%%%%%%%%%%%%%%%%%%%%%%%%%%%%%%%%%%%%%%%%%%%%%%%%%%%%%%%
%%%%%%%%%%%%%%%%%%%%%%%%%%%%%%%%%%%%%%%%%%%%%%%%%%%%%%%%%%%%%%%%%%%%%%
%%%%%%%%%%%%%%%%%%%%%%%%%%%%%%%%%%%%%%%%%%%%%%%%%%%%%%%%%%%%%%%%%%%%%%
\section{Algorithmic Implications}
\label{sec:alg_imp}
%%%%%%%%%%%%%%%%%%%%%%%%%%%%%%%%%%%%%%%%%%%%%%%%%%%%%%%%%%%%%%%%%%%%%%
%%%%%%%%%%%%%%%%%%%%%%%%%%%%%%%%%%%%%%%%%%%%%%%%%%%%%%%%%%%%%%%%%%%%%%
%%%%%%%%%%%%%%%%%%%%%%%%%%%%%%%%%%%%%%%%%%%%%%%%%%%%%%%%%%%%%%%%%%%%%%

The comparative analysis presented in the last section demonstrated how the logical strength of numerous causal razors can be systematically arranged into a logical hierarchy. Certain readers, nevertheless, may dismiss this logical project as purely abstract which yields no substantial implication to developers or users of causal search algorithms. This section demonstrates how this viewpoint can be repudiated. 

First, recall \textbf{Theorem \ref{CFC-uPm}} proven by \cite{GRaSP} that CFC and u-P-minimality are logically equivalent. This logical discovery entails that certain causal search algorithms inevitably face a problem of \textit{P-minimal local optimum}. More specifically, when there is a detectable violation of CFC (i.e., $\CFC(\Prob) = \varnothing$), there must be at least two P-minimal DAGs that are not Markov equivalent. Suppose that $\G_0$ is a P-minimal DAG that is particularly favorable by a pre-selected criterion (e.g., frugality). In that case, if a causal search algorithm \textit{merely} aims at a certain P-minimal DAG (or the MEC of a P-minimal DAG), it may return a P-minimal DAG $\G_1$ (or $\MEC(\G_1)$) which is sub-optimal according to the pre-selected criterion (e.g., $\G_1$ is denser than $\G_0$). In this case, the algorithm is said to be trapped by a P-minimal local optimum. 

One instance of such algorithm is the \textit{TSP algorithm} recently proposed by \citet{solus2021consistency}. In short, TSP performs a greedy search (using the Chickering algorithm in \citep{chickering2002optimal}) from a DAG induced by an arbitrarily chosen initial permutation of the vertices. They argued that TSP can correctly identify the sparsest DAG even without CFC. Unfortunately, their claim is false because the initial permutation chosen can induce a sub-optimal P-minimal DAG (i.e., not the sparsest). In other words, CFC is \textit{necessary} for the correctness of TSP.\footnote{See \citep{GRaSP} for a more detailed analysis of how faithfulness is necessary and sufficient for the correctness of TSP.} This verifies that a careful comparative analysis of causal razors can assist developers of causal search algorithms to identify the limitation of their work.

In addition, the equivalence between CFC and u-P-minimality implicates a knotty question for researchers in causal discovery. Suppose that P-minimality, as suggested by \citet{zhang2013comparison}, is a reasonably safe assumption to be made about the true causal mechanism. Again, when CFC fails detectably, we are left with at least two P-minimal DAGs that are not Markov equivalent. So, is there any principled reason that renders a (set of) P-minimal DAG(s) preferred to the rest?

One natural approach to tackling this problem is to appeal to a causal razor stronger than P-minimality and potentially shrink the hypothesis space. This approach can be pictorially understood as \textit{ascending the hierarchy} in Figure \ref{fig:Hier_MN}. The figure indicates two candidates that are immediately stronger than P-minimality: frugality and param-minimality. As proven in the last section, though the two principles always resort to the same set of DAGs for linear Gaussian models, they can diverge for multinomial models. Recall \textbf{Example \ref{ex:uFr_not_uParamM}} (with Figure \ref{fig:uFr_not_uParamM}) where both $\G_0$ and $\G_1$ are P-minimal. Frugality prefers $\G_0$ for its sparsity, whereas param-minimality prefers $\G_1$ for its fewest parameters.

The logical independence between the two causal razors unveils a dilemma when implementing causal search algorithms, particular the \textit{score-based} species which aim to identify the best-scoring DAG. Given an observational dataset $\D$, a score-based causal search algorithm with a pre-selected \textit{scoring criterion} aims to return a DAG, or a class of DAGs, that has the highest score with respect to $\mc{D}$. As readers can expect, a scoring criterion respecting frugality (i.e., by preferring sparser Markovian models) can make a different judgment compared to one that respects param-minimality (i.e., by preferring Markovian models with fewer parameters). Thus, the theoretical choice between the two causal razors is now reduced to an algorithmic decision. This further verifies that the logical analysis of causal razors is not purely abstract but also gives rise to a practical dilemma. 

Let us delve deeper into the discussion of scoring criteria. First, we assume that every concerned observational dataset $\D$ henceforth consists of $n$ independent and identically distributed (i.i.d.) samples over a set of variables $\mb{V}$ following a joint probability distribution $\Prob$. We sometimes write $\D_\Prob$ to signify that $\D$ is sampled from $\Prob$. Given that this section is primarily studied in the context of large sample theory, we assume that $n$ approaches infinity for the time being.

One scoring criterion that aligns with the essence of frugality is the \textit{negative edge count} that is used to establish the different varieties of the SP algorithm in \cite{raskutti2018learning} and \cite{solus2021consistency}.  Formally speaking, given an observational dataset $\D_\Prob$ over $\mb{V}$, the negative edge count of a DAG $\G \in \DAG(\mb{v})$ is:\footnote{The original definition of negative edge count in \cite{solus2021consistency} concerns an independence model $\I(\Prob)$ instead of a dataset with the distribution $\Prob$. However, this difference in formulation draws no loss of generality due to our assumption that $\D_\Prob$ follows the distribution $\Prob$ asymptotically.}
\begin{align*}
    \mt{NEC}(\G, \D_\Prob):= \begin{cases}
    \,-|\E(\G)| & \text{ if } \G \in \CMC(\Prob)\\
    \,-\infty & \textit{ otherwise.}
    \end{cases}
\end{align*}
Evidently, $\mt{NEC}$ is in line with frugality where the sparsest DAGs must have the highest score.

On the other hand, one of the most popular scoring criteria in the literature is the \textit{Bayesian information criterion} (BIC), originated by \cite{schwarz1978estimating}, which aims to approximate the \textit{marginal log-likelihood} of a DAG.
%%%%%%%%%%%%%%%%%%%%
\begin{definition}
\label{def:BIC}
(BIC score) Given an observational dataset $\D$ with $n$ i.i.d. observations from a joint probability distribution $\Prob$ over $\mb{V}$ that belongs to a curved exponential family\footnote{See \citep{Kass:1437490} for an in-depth analysis of curved exponential families.}, for $X \in \mb{V}$ and $\mb{S} \subseteq \mb{V} \setminus \{X\}$,
\begin{align}
    \mt{BIC}(X, \mb{S}, \D_\Prob) = 2\,\ell(\hat{\bs\theta}_{X\,|\,\mb{S}}\,;\,\D_\Prob) - c \, |\hat{\bs\theta}_{X\,|\,\mb{S}}| \log(n)
\label{eq:BIC_variable}
\end{align}
where $\ell$ is the log-likelihood function, $\hat{\bs\theta}_{X\,|\,\mb{S}}$ is the maximum likelihood estimator (MLE) of $X$ conditioning on $\mb{S}$, and $c > 0$ is the multiplier for the parameter penalty (which is set to be 1 unless specified otherwise). Also, BIC is a decomposable scoring function such that the score of $\G \in \DAG(\mb{v})$, denoted $\mt{BIC}(\G, \D_\Prob)$, satisfies the following:
\begin{align}
    \mt{BIC}(\G, \D_\Prob) &= \sum_{i \in \mb{v}} \mt{BIC}(X_i, \mb{X}_{\Pa(i, \G)}, \D_\Prob)\nonumber\\
    &= \sum_{i \in \mb{v}} 2\,\ell(\hat{\bs\theta}_{X_i\,|\,\mb{X}_{\Pa(i, \G)}}; \D_\Prob) - c\,|\hat{\bs\theta}_{X_i\,|\,\mb{X}_{\Pa(i, \G)}}| \log (n)\nonumber\\ 
    &= 2\,\ell(\hat{\bs\theta}_\G; \D_\Prob) - c\,|\hat{\bs\theta}_\G| \log(n) 
\label{eq:BIC_DAG}
\end{align}
where $\hat{\bs\theta}_\G$ is the set of estimated independent parameters of $\G$ and $\ell(\hat{\bs\theta}_\G; \D_\Prob)$ is the maximum log-likelihood estimator (MLLE) of $\G$.
\end{definition}
%%%%%%%%%%%%%%%%%%%%

\noindent By this formulation, when considering $\mt{BIC}(X_i, \mb{X}_{\Pa(i, \G)}, \D_\Prob)$ for any vertex $i \in \mb{v}$ and any $\G \in \DAG(\mb{v})$, the MLE $\hat{\bs\theta}_{X_i\,|\,\mb{X}_{\Pa(i, \G)}}$ is the estimation of $\bs\theta_{i, \G}$, that is, the set of parameters needed to compute $\Prob(X_i\,|\,\mb{X}_{\Pa(i, \G)})$. Accordingly, $\hat{\bs\theta}_\G$ is the estimation of $\bs\theta_\G$ by assuming that the parameters in the latter are independent of one another.

As proven by \citet{haughton1988choice}, a nice feature that BIC possesses is that it is \textit{consistent} for distributions belonging to the curved exponential families, including linear Gaussian and multinomial distributions.\footnote{See \cite{Kass:1437490} for a more detailed investigation of curved exponential family of statistical models.}

%%%%%%%%%%%%%%%%%%%%
\begin{definition}
\label{def:consistent_score}
(Consistent scoring criterion) Given an observational dataset $\D$ with $n$ i.i.d. observations from a joint probability distribution $\Prob$ over $\mb{V}$, a scoring criterion $\mt{score}$ is consistent if the following two properties hold for any $\G, \mc{H} \in \DAG(\mb{v})$ in the large sample limit of $n$:
\begin{enumerate}
    \item[(a)] if $\mc{G} \in \CMC(\Prob)$ and $\mc{H} \notin \CMC(\Prob)$, then $\mt{score}(\mc{G}, \mc{D}) > \mt{score}(\mc{H}, \mc{D})$;
    \item[(b)] if $\G, \mc{H} \in \CMC(\Prob)$, and $|\param(\G)| < |\param(\mc{H})|$, then $\mt{score}(\mc{G}, \mc{D}) > \mt{score}(\mc{H}, \mc{D})$.
\end{enumerate}
\end{definition}
%%%%%%%%%%%%%%%%%%%%

%%%%%%%%%%%%%%%%%%%%
\begin{theorem}
\label{thm:BIC_consistent}
\citep{haughton1988choice, chickering2002optimal} BIC is a consistent scoring criterion for joint probability distributions belonging to the curved exponential families.
\end{theorem}
%%%%%%%%%%%%%%%%%%%%

\noindent The second condition in \textbf{Definition \ref{def:consistent_score}} aligns well with the concept of param-minimality. When a DAG $\G$ is param-minimal, all consistent scoring criteria, including BIC, will unanimously judge that $\G$ has the highest score over all DAGs. The prominent \textit{Greedy Equivalence Search} (GES) algorithm from \cite{chickering2002optimal} is a classical score-based algorithm that utilizes BIC.

One can easily verify that $\mt{NEC}$ satisfies only the first condition of \textbf{Definition \ref{def:consistent_score}} but not the second. Consider \textbf{Example \ref{ex:uFr_not_uParamM}} again. We have $\mt{NEC}(\G_0, \D_\Prob) = -5 > -6 = \mt{NEC}(\G_1, \D_\Prob)$ while $|\param(\G_0)| = 37 > 35 = |\param(\G_1)|$ (where $\Prob$ is a multinomial distribution as constructed in \textbf{Appendix \ref{app:ex_uFr_not_uParamM}}. This demonstrates that negative edge count is not a consistent scoring criterion (for multinomial distributions). 

Now imagine that you are going to deploy a score-based algorithm on an observational dataset $\mc{D}_\Prob$ where $\ParamM(\Prob) \neq \Fr(\Prob)$ like that in \textbf{Example \ref{ex:uFr_not_uParamM}}. Should you select $\mt{NEC}$ or $\mt{BIC}$ as your scoring criterion used by the algorithm? Your answer to this question reflects your preference for a specific methodological causal razor. For instance, due to the wide usage of BIC, it is unsurprising to see statisticians and computer scientists inclined to adopt param-minimality. Generally speaking, a param-minimal hypothesis can avoid the problem of overfitting in finite samples better than the non-param-minimal ones.\footnote{As argued by \citet{MacKay_1992}, a complex hypothesis (i.e., with more parameters) tends to have a lower marginal likelihood and thus is easier to overfit the training data compared to a simpler hypothesis.} Following this line of thought, a consistent scoring criterion is preferred to its inconsistent alternatives. Also, perhaps edge count can be interpreted as a shorthand for counting parameters in some special cases like linear Gaussian models. But this shorthand is not generally applicable, like for multinomial models.

Still, the argument above merely suggests that a consistent scoring criterion should be preferred \textit{when the desideratum is to solve a specific problem} (e.g., overfitting). It does not imply that $\mt{NEC}$ is methodologically inferior to a consistent scoring criterion in nature. Indeed, it is not hard to propose a different desideratum that favors $\mt{NEC}$ over $\mt{BIC}$. For example, counting edges is clearly much more computationally efficient than computing $\mt{BIC}$. When the computational resource is lacking, researchers can justifiably embrace frugality as a methodological principle and use $\mt{NEC}$.

In addition, how about the \textit{non-parametric} cases where the underlying joint distribution of the data cannot be represented parametrically? In those cases, the distribution may not belong to the exponential curved family and we have no consistency result like \textbf{Theorem \ref{thm:BIC_consistent}}. Still, one can employ non-parametric CI tests (e.g., KCIT from \citep{Strobl_nonparametric_CI_test}) to construct a DAG, and then use $\mt{NEC}$ to score DAGs. So $\mt{NEC}$ can be viewed as a more general scoring criterion that does not commit to any parametric assumption.

Incidentally, one may suggest looking at the dilemma from a \textit{descriptive} perspective by inquiring how the simplicity of causal hypotheses is perceived by human beings psychologically. This may hopefully shed some light on which principle comes with a greater \textit{naturalistic} flavor. Multiple experimental research in psychology, including \citep{Children_node_simplicity} and \citep{Abductive_node_simplicity}, find positive evidence that people identify simplicity as the number of causes invoked in a causal explanation.\footnote{In addition, \citet{pacer_ockham} compare the \textit{node simplicity} (i.e., the number of causes involved in an explanation) with \textit{root simplicity} (i.e., the number of causes left unexplained in an explanation), and observe that explanatory preferences track the latter.} When expressed more formally, counting causes can be reduced to counting edges in DAGs. On the contrary, \citet{Bayesian_Razor_of_People} provided evidence that people's judgments of simplicity in causal explanations are consistent with the behavior of penalizing free parameters in a hypothesis. Apparently, we have limited descriptive results to solve the methodological dilemma.

Lastly, from an \textit{epistemic} point of view, one can question whether frugality or param-minimality is more likely to converge to the truth. \cite{forster2020frugal} provides an argument for frugality by arguing that frugal DAGs entail the greatest number of \textit{basic CIs}.\footnote{The set of basic CIs entailed by a DAG can be easily obtained by the \textit{ordered Markov condition}. In short, given a topological order $\pi$ of $\G$, the set of basic CIs of $\G$ is $\{\la X_i, X_j \mid \mb{X}_{\Pa(i, \G)}\ra: i \in \mb{v} \text{ and } j \text{ precedes } i \text{ in } \pi \text{ and } j \notin \Pa(i, \G)\}.$ See \citep{lauritzenlectures} for a more formal definition of the condition.} However, their argument implicitly assumes an equal weighting of all basic CIs. This assumption is questionable in the multinomial context because each CI can be decomposed into \textit{probabilistic equalities}, and different CIs can be associated with a different number of probabilistic equalities.\footnote{For instance, suppose that $X_1, X_2$, and $X_3$ are all binary. Then $\la X_1, X_2 \mid \{X_3\}\ra \in \I(\Prob)$ can be further decomposed into two probabilistic equalities: $\Prob(X_1 = 0 \mid X_2 = 0, X_3 = 0) = \Prob(X_1 = 0 \mid X_2 = 1, X_3 = 0)$ and $\Prob(X_1 = 0 \mid X_2 = 0, X_3 = 1) = \Prob(X_1 = 0 \mid X_2 = 1, X_3 = 1)$. On the contrary, there is only one probabilistic equality associated with the marginal independence $\la X_1, X_2\ra \in \I(\Prob)$: $\Prob(X_1 = 0\mid X_2=0) = \Prob(X_1 = 0 \mid X_1 = 1)$.} \cite{MyPhD} showed that a param-minimal DAG must be associated with the greatest number of \textit{basic probabilistic equalities} (i.e., probabilistic equalities of the basic CIs). On the other hand, the author also provided a \textit{Bayesian} justification for the principle of param-minimality. 

Instead of targeting an absolute judgment, this work embraces a rather flexible methodological attitude on the preference of scoring criteria. The interest and desideratum at stake can vastly influence the choice of an appropriate scoring criterion on a case-by-case basis. But, most importantly, algorithm developers and users should be aware of the fact that their choices are subtly tied to a methodological preference of causal razors. A more philosophical comparison between frugality and param-minimality will be explored in a future work. 

\section{Discussions}
\label{sec:discuss}
%%%%%%%%%%%%%%%%%%%%%%%%%%%%%%%%%%%%%%%%%%%%%%%%%%%%%%%%%%%%%%%%%%%%%%
%%%%%%%%%%%%%%%%%%%%%%%%%%%%%%%%%%%%%%%%%%%%%%%%%%%%%%%%%%%%%%%%%%%%%%
%%%%%%%%%%%%%%%%%%%%%%%%%%%%%%%%%%%%%%%%%%%%%%%%%%%%%%%%%%%%%%%%%%%%%%

Causal discovery cannot be done without making assumptions on how the true causal mechanism relates to the underlying joint probability distribution. We studied these assumptions under the generic term \textit{causal razors} in this work. Many causal razors discussed in the literature are structural in the sense that they are defined independent of any algorithmic or parametric assumption, including the causal faithfulness condition and the causal frugality assumption. We initiate a different kind of causal razors that are not purely structural, namely causal parameter minimality condition. We provided a comprehensive logical comparisons of thirteen causal razors in the context of multinomial causal models, and observed that frugality and parameter minimality are logically independent despite their similarity. This raises algorithmic concerns on which causal razor should be assumed, methodologically or ontologically, when performing causal discovery.

Below we briefly lay out a number of directions to stretch the study of causal razors. First, we can extend the list of structural causal razors. Recent works like \citep{Ng_local_A_star} and \citep{Zhalama2019asp} have proposed some new candidates. Also, in light of the use of parameterizing sets in \textbf{Appendix \ref{app:ResF_uParamM}}, one can define a structural causal razor restricted to the set-theoretic minimal class of parameterizing sets. Stimulated by the proof in \textbf{Lemma \ref{lem:imset_crucial_lemma}}, we conjecture that this new causal razor logically lies between restricted faithfulness and parameter minimality for multinomial causal models. 

Second, we did not discuss any algorithmic causal razor in this work. The ESP assumption and GRaSP-razors in \cite{solus2021consistency} and \cite{GRaSP} respectively discuss how these algorithmic causal razors are strictly weaker than faithfulness. It will be interesting to see how they relate to the ones surveyed in this work. Also, we have only inspected parameter minimality (and its unique variant) in the context of linear Gaussian and multinomial causal models. It is unclear how parameter-minimality and frugality are related under other parametric assumptions (e.g., conditional Gaussian distributions). 

Third, the choice of causal razors to be assumed can be assessed with computational concerns in practice. Even though it is always safer to assume weaker, weakening faithfulness is accompanied by an algorithmic trade-off. Consider GES which assumes faithfulness. Despite the frequent violations of faithfulness, GES is favored by researchers for its computational efficiency and scalability. As experimented in \cite{Joe_GES_1M}, some version of GES can even scale up to a million variables. By assuming u-frugality, contrarily, the SP algorithm from \cite{raskutti2018learning} can hardly handle cases with more than 10 variables. Greedy versions of SP have been pursued by \cite{GRaSP} and \cite{solus2021consistency}, but their scalability is nowhere close to that of GES. This demonstrates how practical concerns can outweigh theoretical correctness when researchers decide which causal razor to be assumed. To abridge the gap between these two factors, a fast and scalable weaker-than-faithfulness algorithm is highly desirable. 

Fourth, the comparative analysis of causal razors in this work can be studied in terms of simulations. For example, how often do we observe almost violations of, say, frugality and parameter minimality (and their unique variants)? If parameter minimality and frugality rarely disagree at the sample level, the choice of which causal razor to be assumed, and the choice between negative edge count and a consistent scoring criterion, will be unsubstantial when implementing causal search algorithms on finite samples. To answer this question, a comprehensive simulation study is required to verify how frequent a causal razor can be almost violated.

Lastly, all causal razors discussed in this work are defined over DAGs such that the analysis is subsumed under the framework of \textit{causal sufficiency}, that is, no latent variables in the true causal model. However, by replacing the d-separations with \textit{m-separations} in \citep{Richardson_MAGs}, all of our discussed structural causal razors can be defined in terms of \textit{maximal ancestral graphs} (MAGs) that handle latent variables. By adopting the technique of parameterizing acyclic directed mixed graphs (ADMGs), one can also effectively enumerate the number of parameters in a MAG (in the multinomial context). Accordingly, it is reasonable to expect that all of our logical results can be extended to the causally insufficient framework.

% Lastly, the problem of biased parametric preference for parameter minimality remained an open question. To recall, in the three-vertex case, an unshielded collider always have more parameters than an unshielded non-collider. If both DAGs are Markovian to the joint multinomial distribution, parameter minimality will prefer the non-collider. We suspect this problem can be re-framed as a more general philosophical-statistical discussion. For example, a Bayesian would prefer the non-collider, whereas a frequentist interested in a restricted from of uniform consistency will prefer neither causal models. But this theoretical discussion awaits a different occasion. 
\bibliography{ref.bib} 
\newpage
\appendix
% \addappheadtotoc
% \begin{group}
% \let\clearpage\relax
%%%%%%%%%%%%%%%%%%%%%%%%%%%%%%%%%%%%%%%%%%%%%%%%%%%%%%%%%%%%%%%%%%%%%%
%%%%%%%%%%%%%%%%%%%%%%%%%%%%%%%%%%%%%%%%%%%%%%%%%%%%%%%%%%%%%%%%%%%%%%
%%%%%%%%%%%%%%%%%%%%%%%%%%%%%%%%%%%%%%%%%%%%%%%%%%%%%%%%%%%%%%%%%%%%%%
\section{Graphoid Axioms}
\label{app:graphoid}
%%%%%%%%%%%%%%%%%%%%%%%%%%%%%%%%%%%%%%%%%%%%%%%%%%%%%%%%%%%%%%%%%%%%%%
%%%%%%%%%%%%%%%%%%%%%%%%%%%%%%%%%%%%%%%%%%%%%%%%%%%%%%%%%%%%%%%%%%%%%%
Consider any pairwise disjoint sets of variables $\mb{W}, \mb{X}, \mb{Y},$ and $\mb{Z}$.
\begin{align*}
\mb{X} \CI \mb{Y} \,|\,\mb{Z} &\,\,\Rightarrow\,\, \mb{Y} \CI \mb{X} \,|\,\mb{Z} & (\textit{symmetry})\\
\mb{X} \CI \mb{Y} \cup \mb{W} \,|\,\mb{Z} &\,\,\Rightarrow\,\, \mb{X} \CI \mb{Y} \,|\,\mb{Z}  &
(\textit{decomposition})\\
\mb{X} \CI \mb{Y} \cup \mb{W} \,|\,\mb{Z} &\,\,\Rightarrow\,\, \mb{X} \CI \mb{Y} \,|\,\mb{Z} \cup \mb{W} &
(\textit{weak union})\\
(\mb{X} \CI \mb{Y}\,|\,\mb{Z}) \wedge (\mb{X} \CI \mb{W}\,|\,\mb{Z} \cup \mb{Y}) &\,\,\Rightarrow\,\, \mb{X} \CI \mb{Y}\cup \mb{W}\,|\,\mb{Z} &(\textit{contraction})\\
(\mb{X} \CI \mb{Y}\,|\,\mb{Z} \cup \mb{W}) \wedge (\mb{X} \CI \mb{W}\,|\,\mb{Z} \cup \mb{Y}) &\,\,\Rightarrow\,\, \mb{X} \CI \mb{Y}\cup \mb{W}\,|\,\mb{Z} &(\textit{intersection})\\
(\mb{X} \CI \mb{Y}\,|\,\mb{Z}) \wedge (\mb{X} \CI \mb{W}\,|\,\mb{Z}) &\,\,\Rightarrow\,\, \mb{X} \CI \mb{Y}\cup \mb{W}\,|\,\mb{Z} & (\textit{composition})
\end{align*}

\noindent A distribution $\Prob$ is a \textit{semigraphoid} if $\I(\Prob)$ is closed under \textit{symmetry}, \textit{decomposition}, \textit{weak union}, and \textit{contraction}. A semigraphoid $\Prob$ is a \textit{graphoid} if $\I(\Prob)$ is closed under \textit{intersection}. A graphoid $\Prob$ is \textit{compositional} if $\I(\Prob)$ is closed under \textit{composition}. Every joint probability distribution $\Prob$ is a semigraphoid. Every strictly positive distribution is a graphoid, and every linear Gaussian distribution is a compositional graphoid. See \cite[Chapter 2]{Studeny_book} for a more comprehensive study of graphoid axioms. In addition, applications of \textit{symmetry} in our upcoming proofs will be performed implicitly.
%%%%%%%%%%%%%%%%%%%%%%%%%%%%%%%%%%%%%%%%%%%%%%%%%%%%%%%%%%%%%%%%%%%%%%
%%%%%%%%%%%%%%%%%%%%%%%%%%%%%%%%%%%%%%%%%%%%%%%%%%%%%%%%%%%%%%%%%%%%%%
%%%%%%%%%%%%%%%%%%%%%%%%%%%%%%%%%%%%%%%%%%%%%%%%%%%%%%%%%%%%%%%%%%%%%%
\section{Proof of Theorem 4.2}
\label{app:CFC_uPM}
%%%%%%%%%%%%%%%%%%%%%%%%%%%%%%%%%%%%%%%%%%%%%%%%%%%%%%%%%%%%%%%%%%%%%%
%%%%%%%%%%%%%%%%%%%%%%%%%%%%%%%%%%%%%%%%%%%%%%%%%%%%%%%%%%%%%%%%%%%%%%
\begin{lemma}
\label{CMC_single_CI}
\cite{GRaSP} Given a joint probability distribution $\Prob$ over $\mb{V}$, for any $\la X_i, X_j\,|\,\mb{X}_{\mb{s}} \ra \in \I(\Prob)$, there exists $\G \in \DAG(\mb{v})$ s.t. $\I(\G) = \{\la X_i, X_j\,|\,\mb{X}_{\mb{s}} \ra\}$. 
\end{lemma}
\begin{pf}
Suppose that $\mb{V}$ consists of $m$ variables such that $\mb{v} = \{1,...,m\}$ where $m \geq 2$. When $m = 2$, an empty DAG will suffice to prove the lemma. Now consider the case where $m \geq 3$. Suppose that $\la X_i, X_j\,|\,\mb{X}_\mb{s}\ra \in \I(\Prob)$ for some distinct $i, j \in \mb{v}$ and $\mb{s} \subseteq \mb{v} \setminus \{i, j\}$. Without loss of generality, we relabel the indices such that $i = 1$, $j = k + 2$ and $\mb{s} = \la 2,..., k+1\ra$ where $\mb{v}\setminus\mb{s} = \la k+3,..., m \ra$. Now consider the following algorithm which aims to construct the desired $\G \in \DAG(\mb{v})$.
\begin{algorithm}
\DontPrintSemicolon
$\G \ot \text{a complete undirected graph over } \mb{v}$\;
remove the adjacency $(1, k+2) \in \SK(\G)$\;
\ForEach{$(j, k) \in \SK(\G)$}{
    \If{$j < k$}{
        orient $j \to k$ in $\G$
    }
}
return $\G$    
\end{algorithm}

\noindent Line 3 to 5 guarantee that $\mathcal{G}$ is a DAG. This is because all edges are directed and pointing from lower indices to higher indices such that no directed cycle can be induced. Finally, $1 \perp_\G k+2\,|\,\mb{s}$ holds since all directed paths from $1$ to $k+2$ either contain a non-collider $i \in \mb{s}$ or contain a collider $i \notin \mb{s}$. Therefore, $\I(\G) = \{\la X_1, X_{k+2}\,|\,\mb{X}_\mb{s})\}$ because no other d-separation relations hold in $\mathcal{G}$.
\end{pf}\\

\noindent \textbf{Theorem \ref{CFC-uPm}.} \textit{\cite{GRaSP} For any joint probability distribution $\Prob$, $\CFC(\Prob) = \uPm(\Prob)$.}
\begin{pf}
$[\subseteq]$ Suppose that $\G \in \CFC(\Prob)$. It follows from \textbf{Theorem \ref{thm:rz_sb}}(a) that $\G \in \Pm(\Prob)$. For any $\G' \in \CMC(\Prob)$, if $\I(\G') \subset \I(\G)$, then $\G' \notin \Pm(\Prob)$. Hence, if $\G' \in \Pm(\Prob)$, then $\I(\G') = \I(\G)$. Hence, $\G \in \uPm(\Prob)$.

$[\supseteq$] Suppose that $\G \notin \CFC(\Prob)$. Given that $\uPm(\Prob) \subseteq \Pm(\Prob)$ by \textbf{Definition \ref{def:P-minimal}}, we have $\G \notin \uPm(\Prob)$ immediately if $\G \notin \Pm(\Prob)$. Consider the case where $\G \in \Pm(\Prob)$. It follows from $\G \notin \CFC(\Prob)$ that $\Psi_{\G, \Prob} \neq \varnothing$. Consider any unfaithful CI $\psi \in \Psi_{\G, \Prob}$. By \textbf{Lemma \ref{CMC_single_CI}}, we can construct a DAG $\G'$ such that $\I(\G') = \{\psi\}$. Consequently, there exists $\G'' \in \Pm(\Prob)$ such that $\I(\G') \subseteq \I(\G'') \subseteq \I(\Prob)$. From $\psi \in \I(\G'')$ but $\psi \notin \I(\G)$, it follows that $\G'' \notin \MEC(\G)$. Therefore, we have $\G \notin \uPm(\Prob)$ from $\G, \G'' \in \Pm(\Prob)$.
\end{pf}\\
%%%%%%%%%%%%%%%%%%%%%%%%%%%%%%%%%%%%%%%%%%%%%%%%%%%%%%%%%%%%%%%%%%%%%%
%%%%%%%%%%%%%%%%%%%%%%%%%%%%%%%%%%%%%%%%%%%%%%%%%%%%%%%%%%%%%%%%%%%%%%
%%%%%%%%%%%%%%%%%%%%%%%%%%%%%%%%%%%%%%%%%%%%%%%%%%%%%%%%%%%%%%%%%%%%%%
\section{Proof of Theorem 5.3}
\label{app:ParamM_Pm}
%%%%%%%%%%%%%%%%%%%%%%%%%%%%%%%%%%%%%%%%%%%%%%%%%%%%%%%%%%%%%%%%%%%%%%
%%%%%%%%%%%%%%%%%%%%%%%%%%%%%%%%%%%%%%%%%%%%%%%%%%%%%%%%%%%%%%%%%%%%%%

We are going to make use of the famous results in \citep{Chickering1995} and \citep{chickering2002optimal} related to \textit{covered edge reversals} to show that $\ParamM(\Prob) \subseteq \Pm(\Prob)$ for any joint multinomial distribution $\Prob$. To be precise, for any DAG $\G$, a directed edge $j \to k \in \E(\G)$ is a \textit{covered edge} if $\Pa(j, \G) = \Pa(k, \G) \setminus \{j\}$.\\

%%%%%%%%%%%%%%%%%%%%
\begin{theorem}
\label{thm:covered_MEC_chain}
\cite{Chickering1995} Given a set of variables $\mb{V}$, consider any pair of DAGs $\G, \mc{H} \in \DAG(\mb{v})$ where $\mc{H} \in \MEC(\G)$, and for which there are $k$ edges in $\G$ that have opposite orientation in $\mc{H}$. Then there exists a sequence of $k$ distinct covered edge reversals in $\G$ s.t. $\G$ becomes $\mc{H}$ after all reversals.\\
\end{theorem}
%%%%%%%%%%%%%%%%%%%%

%%%%%%%%%%%%%%%%%%%%
\begin{theorem}
\label{thm:Chickering_seq} 
\cite{chickering2002optimal} Given a set of variables $\mb{V}$, for every pair of DAGs $\G, \mc{H} \in \mt{DAG}(\mb{v})$, if $\I(\mc{H}) \subseteq \I(\G)$, there exists a sequence of DAGs, call it a \textit{Chickering sequence} $\la \mc{H} = \G_1, \G_2, ..., \G_k = \G\ra$ (from $\mc{H}$ to $\G$) s.t. $\I(\G_{i}) \subseteq \I(\G_{i+1})$ and $\G_{i+1}$ is obtained from $\G_i$ by either reversing a covered edge or deleting a directed edge for each $1 \leq i < k$.\\
\end{theorem}
%%%%%%%%%%%%%%%%%%%%

%%%%%%%%%%%%%%%%%%%%
\begin{lemma}
\label{lem:MEC_same_params}
Given a joint multinomial distribution $\Prob$ over $\mb{V}$, consider any pair of DAGs $\G, \mc{G}' \in \DAG(\mb{v})$ where $\mc{G}' \in \MEC(\G)$. Then $\param(\G) = \param(\mc{G}')$.
\end{lemma}
\begin{pf}
Due to \textbf{Theorem \ref{thm:covered_MEC_chain}}, it suffices to show that the number of parameters remains unchanged after the reversal of a covered edge. Suppose that $\G$ and $\G'$ in $\DAG(\mb{v})$ differ by reversing exactly one covered edge such that $\E(\G') = (\E(\G) \setminus \{i \to j\}) \cup \{j \to i\}$ for some $i, j \in \mb{v}$. Further denote $\mt{r}(i), \mt{r}(j)$, and $\mt{r}(\Pa(i, \G))$ as $a, b$, and $c$ respectively. Thus, we have
\begin{align*}
    |\theta_{i, \G}| = (a - 1)c, \hspace{1cm} |\theta_{j, \G}| = (b - 1)ac, \hspace{1cm} |\theta_{i, \G'}| = (a - 1)bc, \hspace{1cm} |\theta_{j, \G'}| = (b - 1)c.
\end{align*}
Notice that $|\theta_{l, \G}| = |\theta_{l, \G'}|$ for each $l \in \mb{v}\setminus \{i, j\}$. By (\ref{eq:BN_param_eq_DAG}), we have:
\begin{align*}
    |\param(\G)| - |\param(\G')| =\,& (|\theta_{i, \G}| + |\theta_{j, \G}|) - (|\theta_{i, \G'}| + |\theta_{j, \G'}|)\\
    =\,& \Big((a-1)c + (b-1)ac \Big) - \Big((a-1)bc + (b-1)c\Big)\\
    =\, & c\,\Big((a-1) + (b-1)a - (a-1)b - (b-1)\Big)\\
    =\, & c\,\Big((a-1)(1-b) + (b-1)(a-1)\Big)\\
    =\, & 0.
\end{align*}
\end{pf}\\
%%%%%%%%%%%%%%%%%%%%

%%%%%%%%%%%%%%%%%%%%
\noindent \textbf{Theorem \ref{thm:ParamM_subset_PM}.} \textit{For any joint multinomial distribution $\Prob$, $\ParamM(\Prob) \subseteq \Pm(\Prob)$ holds.}\\
\vspace{-0.2cm}

\begin{pf}
By reductio, suppose that $\G \in \ParamM(\Prob)$ but $\G \notin \Pm(\Prob)$. So there exists $\G' \in \Pm(\Prob)$ such that $\I(\G) \subset \I(\G') \subseteq \I(\Prob)$. By \textbf{Theorem \ref{thm:Chickering_seq}}, there exists a Chickering sequence to obtain $\G'$ from $\G$. By \textbf{Lemma \ref{lem:MEC_same_params}}, covered edge reversals preserve the number of parameters. On the other hand, edge deletion is obviously an operation that decreases the number of parameters. Given that $\G'$ and $\G$ do not belong to the same MEC, there must be at least one operation of edge deletion to obtain $\G'$ from $\G$. In other words, we have $|\param(\G')| < |\param(\G)|$. Contradiction arises with $\G \in \ParamM(\Prob)$.\end{pf}\\
%%%%%%%%%%%%%%%%%%%%
%%%%%%%%%%%%%%%%%%%%%%%%%%%%%%%%%%%%%%%%%%%%%%%%%%%%%%%%%%%%%%%%%%%%%%
%%%%%%%%%%%%%%%%%%%%%%%%%%%%%%%%%%%%%%%%%%%%%%%%%%%%%%%%%%%%%%%%%%%%%%
%%%%%%%%%%%%%%%%%%%%%%%%%%%%%%%%%%%%%%%%%%%%%%%%%%%%%%%%%%%%%%%%%%%%%%
\section{Proof of Theorem 5.10}
\label{app:ResF_uParamM}
%%%%%%%%%%%%%%%%%%%%%%%%%%%%%%%%%%%%%%%%%%%%%%%%%%%%%%%%%%%%%%%%%%%%%%
%%%%%%%%%%%%%%%%%%%%%%%%%%%%%%%%%%%%%%%%%%%%%%%%%%%%%%%%%%%%%%%%%%%%%%
First, we introduce some concepts facilitating the proof of $\resF(\Prob) \subseteq \uParamM(\Prob)$ for every joint multinomial distribution $\Prob$. In particular, we borrow the concept \textit{parameterizing sets} that has been used to parameterize \textit{acyclic directed mixed graphs} (ADMGs) in \cite{Evans2014}. Given that DAGs are a special class of ADMGs, the concept borrowed can easily be applied to our subject matter.\footnote{I owe a huge debt of gratitude to Bryan Andrews for pointing out this approach and also the relevant literature including \cite{Hemmecke_imset} and \cite{Bryan_imset}. The main proof of the mentioned claim was also sketched by him.}\\ 

%%%%%%%%%%%%%%%%%%%%
\begin{definition}
\label{def:imset_def}
\cite{Studeny_book} Consider any DAG $\G \in \DAG(\mb{v})$. The characteristic imset $\mf{c}_\G: \mathbb{P}(\mb{v})\setminus\{\varnothing\} \to \{0, 1\}$ is defined as follows\footnote{$\mathbb{P}(\mb{X})$ is the powerset of $\mb{X}$ for any set $\mb{X}$.}: 
\begin{align*}
    \mf{c}_\G(\mb{s}) := \begin{cases}
    \,1 & \text{ if }\,\, \{\la X_i, X_j\,|\,\mb{X}_\mb{k}\ra \in \I(\G): \mb{s} \setminus \mb{k} = \{i, j\}\ra\} = \varnothing\\
    \,0 & \text{ otherwise }
    \end{cases}
\end{align*}
for any $\mb{s} \in \mathbb{P}(\mb{v})$ where $\mb{s} \neq \varnothing$.\\
\end{definition}
%%%%%%%%%%%%%%%%%%%%

\noindent Denote the \textit{class of parameterizing sets} of $\G$ as $ \mf{S}(\G) = \{\mb{s} \in \mathbb{P}(\mb{v})\setminus \{\varnothing\}: \mf{c}_\G(\mb{s}) = 1\}$. In other words, $\mb{s}$ is a parameterizing set if and only if the characteristic imset evaluated at $\mb{s}$ is 1. Below is a nice graphical feature of parameterizing sets.\\

%%%%%%%%%%%%%%%%%%%%
\begin{lemma} 
\label{lem:imset_parent}\cite{Hemmecke_imset, Bryan_imset} Consider any DAG $\G \in \DAG(\mb{v})$ and any $\mb{s} \in \mathbb{P}(\mb{v})$. Then $\mf{c}_\G(\mb{s}) = 1$ if and only if there exists $i \in \mb{s}$ s.t. $\mb{s}\setminus \{i\} \subseteq \Pa(i, \G)$.\\
\end{lemma}
%%%%%%%%%%%%%%%%%%%%

\noindent Observe that the existence in the last lemma is necessarily unique. That is, each $\mb{s} \in \mf{S}(\G)$ contains a unique vertex $i$ such that $\mb{s} \setminus \{i\} \subseteq \Pa(i, \G)$ holds.\footnote{By reductio, suppose that $i, j \in \mb{s}$ s.t. $\mb{s} \setminus \{i\} \subseteq \Pa(i, \G)$ and $\mb{s} \setminus \{j\} \subseteq \Pa(j, \G)$. Then $j \in \Pa(i, \G)$ and $i \in \Pa(j, \G)$. Contradiction.} Hence, we can partition $\mf{S}(\G)$ by vertices. Define $\mf{S}(i, \G)$ as the class of parameterizing sets of vertex $i \in \mb{v}$ in $\G \in \DAG(\mb{v})$ such that 
\begin{align}
\label{eq:parameterizing_set_vertex}
    \mf{S}(i, \G) := \{\mb{s} \in \mf{S}(\G): i \in \mb{s} \text{ and } \mb{s} \setminus \{i\} \subseteq \Pa(i, \G)\}.
\end{align}
As a consequence, we have $\mf{S}(\G) = \bigcup_{i \in \mb{v}}\mf{S}(i, \G)$, and $\mf{S}(i, \G) \cap \mf{S}(j, \G) = \varnothing$ for any distinct $i, j \in \mb{v}$.

Below is a crucial result on how the number of parameters of a DAG in a multinomial causal model, as expressed in (\ref{eq:BN_param_eq_DAG}), can be characterized in an alternative manner.\\

%%%%%%%%%%%%%%%%%%%%
\begin{lemma}
\label{lem:param_count}
Consider any DAG $\G \in \DAG(\mb{v})$ of a multinomial causal model and any $i \in \mb{v}$. Then 
\begin{align}
\label{eq:param_count}
    \sum_{\mb{s} \in \mf{S}(i, \G)} \prod_{j \in \mb{s}} (\mt{r}(j) - 1) = (\mt{r}(i) - 1) \times \mt{r}(\Pa(i, \G)).
\end{align}
\end{lemma}
\begin{pf}
Consider the simplest case where $\Pa(\G, i) = \varnothing$. Then $\mf{S}(i, \G) = \{\{i\}\}$ and hence the LHS of (\ref{eq:param_count}) is $\mt{r}(i) - 1$ which is equivalent to the RHS (since $\mt{r}(\varnothing) = 1$).

In the following, without loss of generality, we write $i$ as 0. Let $|\Pa(0, \G)| = k \leq |\mb{v}| - 1$ such that $\Pa(0, \G) = \{1,..., k\}$. To obtain $\mf{S}(0, \G)$, notice the bijection between $\mf{S}(0, \G)$ and $\mathbb{P}(\Pa(0, \G))$ such that each $\mb{s} \in \mf{S}(0, \G)$ can be uniquely identified by $\mb{s}' \cup \{0\}$ for some $\mb{s}' \in \mathbb{P}(\Pa(0, \G))$. 

Now, we want to show (\ref{eq:param_count}) by an induction on $k$. To begin with, consider the base case where $k = 1$ such that $\Pa(0, \G) = \{1\}$ and $\mf{S}(0, \G) = \{\{0\}, \{0, 1\}\}$. The LHS of (\ref{eq:param_count}) becomes 
\begin{align*}
    \sum_{\mb{s} \in \mf{S}(0, \G)} \prod_{j \in \mb{s}} (\mt{r}(j) - 1) = (\mt{r}(0)-1) + (\mt{r}(0) - 1) (\mt{r}(1)-1) = (\mt{r}(0) - 1)\, \mt{r}(1)
\end{align*}
where $\mt{r}(1) = \mt{r}(\{1\}) = \mt{r}(\Pa(0, \G))$. Thus, the base case is proven. Consider the inductive hypothesis that (\ref{eq:param_count}) holds for some arbitrary $k-1 \leq |\mb{v}|-2$ for DAG $\G$ such that $\Pa(0, \G) = \{1, ..., k-1\}$. Then we construct $\G'$ by adding the directed edge $k \to 0$ to $\G$ such that $\Pa(0, \G') = \{1, ..., k-1, k\}$. We want to show that (\ref{eq:param_count}) holds for $\G'$ and the vertex 0. 

Observe that $\mf{S}(0,\G) \subset \mf{S}(0,\G')$ where the former has a size of $2^{k-1}$ and the latter of $2^k$. Also, there is a bijection between $\mf{S}(0,\G') \setminus \mf{S}(0,\G)$ and $\mf{S}(0,\G)$ such that each $\mb{s}' \in \mf{S}(0, \G')\setminus \mf{S}(0, \G)$ can be uniquely identified by $\mb{s} \cup \{k\}$ for some $\mb{s} \in \mf{S}(0,\G)$. This observation yields the following.
\begin{align}
\label{eq:parameterizing_lemma_eq}
    \sum_{\mb{s} \in \mf{S}(0,\G')} \prod_{j \in \mb{s}} (\mt{r}(j) - 1) &= \sum_{\mb{s} \in \mf{S}(0,\G)} \prod_{j \in \mb{s}} (\mt{r}(j) - 1) + \sum_{\mb{s} \in \mf{S}(0,\G') \setminus \mf{S}(0,\G)}  \prod_{j \in \mb{s}} (\mt{r}(j) - 1)
\end{align}
Consider the second term on the RHS of (\ref{eq:parameterizing_lemma_eq}) in particular.
\begin{align*}
    \sum_{\mb{s} \in \mf{S}(0, \G') \setminus \mf{S}(0,\G)}  \prod_{j \in \mb{s}} (\mt{r}(j) - 1)
    & = \sum_{\mb{s} \in \mf{S}(0, \G') \setminus \mf{S}(0,\G)} \prod_{j \in \mb{s}\setminus \{k\}} (\mt{r}(j) - 1) (\mt{r}(k) - 1)\\
    & = (\mt{r}(k) - 1) \times \sum_{\mb{s} \in \mf{S}(0, \G') \setminus \mf{S}(0,\G)} \prod_{j \in \mb{s}\setminus \{k\}} (\mt{r}(j) - 1)\\
    & = (\mt{r}(k) - 1) \times \sum_{\mb{s} \in \mf{S}(0,\G)} \prod_{j \in \mb{s}} (\mt{r}(j) - 1)\\
    &= (\mt{r}(k) - 1) \times (\mt{r}(0) - 1) \times \mt{r}(\Pa(0, \G))
\end{align*}
where the last equality is obtained by the inductive hypothesis. Finally, by plugging the above into (\ref{eq:parameterizing_lemma_eq}):
\begin{align*}
    \sum_{\mb{s} \in \mf{S}(0,\G')} \prod_{j \in \mb{s}} (\mt{r}(j) - 1) &= \sum_{\mb{s} \in \mf{S}(0,\G)} \prod_{j \in \mb{s}} (\mt{r}(j) - 1) + \sum_{\mb{s} \in \mf{S}(0,\G') \setminus \mf{S}(0,\G)}  \prod_{j \in \mb{s}} (\mt{r}(j) - 1)\\
    &= \bigg((\mt{r}(0) - 1) \times \mt{r}(\Pa(0, \G))\bigg) + \bigg((\mt{r}(k) - 1) \times (\mt{r}(0) - 1) \times \mt{r}(\Pa(0, \G))\bigg)\\
    &= (\mt{r}(0) - 1) \times \mt{r}(\Pa(0, \G)) \times \big(1 + (\mt{r}(k) - 1)\big)\\
    &= (\mt{r}(0) - 1) \times \mt{r}(\{1,...,k-1\}) \times \mt{r}(k)\\
    &= (\mt{r}(0) - 1) \times \mt{r}(\{1,...,k-1, k\})\\
    &= (\mt{r}(0) - 1) \times \mt{r}(\Pa(0, \G')).
\end{align*}
\end{pf}\\
%%%%%%%%%%%%%%%%%%%%

%%%%%%%%%%%%%%%%%%%%
\begin{corollary}
\label{coro:MN_param_new}
Consider any DAG $\G \in \DAG(\mb{v})$ of a multinomial causal model. Then
\begin{align}
\label{eq:MN_param_new}
    |\param(\G)| = \sum_{i \in \mb{v}} \sum_{\mb{s} \in \mf{S}(i,, \G)} \prod_{j \in \mb{s}} (\mt{r}(j) - 1) = \sum_{\mb{s} \in \mf{S}(\G)} \prod_{j \in \mb{s}} (\mt{r}(j) - 1).
\end{align}
\end{corollary}
\,\\
%%%%%%%%%%%%%%%%%%%%

Now we show a crucial lemma stating that the class of parameterizing sets in a restricted-faithful DAG is necessarily a subset of that in any Markovian DAG. As a consequence, this claim entails our desired result that $\resF(\Prob) \subseteq \uParamM(\Prob)$ for any joint multinomial distribution $\Prob$.\\

%%%%%%%%%%%%%%%%%%%%
\begin{lemma}
\label{lem:imset_crucial_lemma}
For any joint multinomial distribution $\Prob$, if $\G \in \resF(\Prob)$, then $\mf{S}(\G) \subseteq \mf{S}(\G')$ for any $\G' \in \CMC(\Prob)$.
\end{lemma}
%%%%%%%%%%%%%%%%%%%%
\begin{pf}
First, following from $\G \in \adjF(\Prob)$, we have $\SK(\G) \subseteq \SK(\G')$ for any $\G' \in \CMC(\Prob)$. Now we prove the lemma by considering the cardinality of each $\mb{s} \in \mf{S}(\G)$ (where $|\mb{s}| \geq 1$ as required by \textbf{Definition \ref{def:imset_def}}). When $|\mb{s}| = 1$, we have $\mb{s} \in \mf{S}(\G')$ trivially. Consider $|\mb{s}| = 2$ where $\mb{s} = \{i, j\}$ for some distinct $i, j \in \mb{v}$. Notice that $\mb{s}$ corresponds to the adjacency between $i$ and $j$ in $\G$. Since $\SK(\G) \subseteq \SK(\G')$, $i$ and $j$ must also be adjacent in $\G'$ and thus $\{i, j\} = \mb{s} \in \mf{S}(\G')$.

Consider $|\mb{s}| = 3$ where $\mb{s} = \{i, j, k\}$. Given that $\mb{s} \in \mf{S}(\G)$, it follows from \textbf{Lemma \ref{lem:imset_parent}} that $\{i, j, k\}$ forms either an unshielded collider or a triangle in $\G$ (because a triangle must be a shielded collider to avoid any cycle in a DAG). From $\G \in \resF(\Prob)$, $\{i, j, k\}$ must also form either an unshielded collider or a triangle in $\G'$. Hence, $\{i, j, k\} = \mb{s} \in \mf{S}(\G')$ by \textbf{Lemma \ref{lem:imset_parent}}.

Lastly, consider $|\mb{s}| > 3$. By \textbf{Lemma \ref{lem:imset_parent}}, there must exist $i \in \mb{s}$ such that $\mb{s} \setminus \{i\} \subseteq \Pa(i, \G)$. Accordingly, for every pair of distinct $j, k \in \mb{s} \setminus \{i\}$, we have $\{i, j\}, \{i, k\}, \{i, j, k\} \in \mf{S}(\G)$. Following the above, we have $\{i, j\}, \{i, k\}, \{i, j, k\} \in \mf{S}(\G')$ as well. It entails that $\la X_i, X_j\,|\,\mb{X}_\mb{l} \cup \{X_k\}\ra, \la X_i, X_k\,|\,\mb{X}_\mb{l} \cup \{X_j\}\ra, \la X_j, X_k\,|\,\mb{X}_\mb{l} \cup \{X_i\}\ra \notin \I(\G')$ for every $\mb{l} \subseteq \mb{v}$ where $\mb{s} \setminus \mb{l} = \{i, j, k\}$. By \textbf{Definition \ref{def:imset_def}}, it follows that $\mf{c}_{\G'}(\mb{s}) = 1$ and therefore $\mb{s} \in \mf{S}(\G')$.
\end{pf}\\

%%%%%%%%%%%%%%%%%%%%
\noindent \textbf{Theorem \ref{thm:resF_subset_uParamM}.} \textit{For any joint multinomial distribution $\Prob$, $\resF(\Prob) \subseteq \uParamM(\Prob)$ holds.}\\
\vspace{-0.2cm}

\begin{pf}
Consider $\G \in \resF(\Prob) = \adjF(\Prob) \cap \oriF(\Prob)$. From $\G \in \adjF(\Prob)$, if a DAG $\G' \in \DAG(\mb{v})$ misses any of $\G$'s adjacencies, $\G'$ is not Markovian (to $\Prob$). This observation implies three results. First, no DAG sparser than $\G$ can be Markovian. Second, for any Markovian $\G'$ which is strictly denser to $\G$, the set of adjacencies in $\G'$ must be a proper superset of that in $\G$ (i.e., $\SK(\G) \subset \SK(\G')$). Lastly, those Markovian DAGs that are equally sparse as $\G$ must have the same skeleton as $\G$. The last point, together with $\G \in \oriF(\Prob)$, entails that all those equally sparse DAGs are in the same MEC as $\G$, and thus having the same number of parameters according to \textbf{Lemma \ref{lem:MEC_same_params}}. In other words, to prove that $\G \in \uParamM(\Prob)$, it suffices to prove that $|\param(\G)| < |\param(\G')|$ for any $\G' \in \CMC(\Prob)$ satisfying $\SK(\G) \subset \SK(\G')$.

Now consider any of such $\G'$. By \textbf{Lemma \ref{lem:imset_crucial_lemma}}, we have $\mf{S}(\G) \subseteq \mf{S}(\G')$. However, notice that there must exist $i, j \in \mb{v}$ such that $(i, j) \in \SK(\G') \setminus \SK(\G)$. It follows that $\{i, j\} \in \mf{S}(\G')$ but $\{i, j\} \notin \mf{S}(\G)$. Consequently, we have $\mf{S}(\G) \subset \mf{S}(\G')$. Therefore, by (\ref{eq:MN_param_new}) in \textbf{Corollary \ref{coro:MN_param_new}}, $|\param(\G)| < |\param(\G')|$.
\end{pf}\\
%%%%%%%%%%%%%%%%%%%%
%%%%%%%%%%%%%%%%%%%%%%%%%%%%%%%%%%%%%%%%%%%%%%%%%%%%%%%%%%%%%%%%%%%%%%
%%%%%%%%%%%%%%%%%%%%%%%%%%%%%%%%%%%%%%%%%%%%%%%%%%%%%%%%%%%%%%%%%%%%%%
%%%%%%%%%%%%%%%%%%%%%%%%%%%%%%%%%%%%%%%%%%%%%%%%%%%%%%%%%%%%%%%%%%%%%%
\section{Constructing the Examples}
\label{app:construction}
%%%%%%%%%%%%%%%%%%%%%%%%%%%%%%%%%%%%%%%%%%%%%%%%%%%%%%%%%%%%%%%%%%%%%%
%%%%%%%%%%%%%%%%%%%%%%%%%%%%%%%%%%%%%%%%%%%%%%%%%%%%%%%%%%%%%%%%%%%%%%

This section discusses how the examples in \textbf{Section \ref{sec:paramM}} can be simulated constructively. Recall that the claim in each of these examples is the existence of a joint multinomial distribution $\headcirc\Prob$ (where the overhead $\circ$ signifies that it is the joint probability distribution we aim to prove its existence) satisfying certain conditions (e.g., $\adjF(\headcirc\Prob) \setminus \ParamM(\headcirc\Prob) \neq \varnothing$). Also observe that each of them concerns a particular DAG $\G \in \CMC(\headcirc\Prob)$ such that $\I(\headcirc\Prob) \setminus \I(\G) = \Psi_{\G, \headcirc\Prob} \neq \varnothing$ (i.e., $\G$ is not faithful to $\headcirc\Prob$). To show the existence of such $\headcirc\Prob$, we construct a joint multinomial distribution $\Prob$ such that $\I(\Prob)$ satisfies the conditions in question. 
The construction can be divided into two steps. First, we obtain $\I(\G)$ by d-separation, and introduce the set of CIs in $\Psi_{\G, \headcirc\Prob}$ by a specific assignment of probabilities such that $\I(\G) \cupdot \Psi_{\G, \headcirc\Prob} \subseteq \I(\Prob)$. The second step is to ensure that the mentioned assignment does not introduce any extra CI in $\I(\Prob)$. Once both steps are done, we obtain $\I(\Prob) = \I(\G) \cup \Psi_{\G, \headcirc\Prob} = \I(\headcirc\Prob)$ as desired. Since the second step involves tedious but straightforward arithmetic derivations, we will not go through the details except in \textbf{Example \ref{ex:adjF_not_paramM}} for an illustration.

Now we introduce some necessary notations for our constructions. Consider any joint multinomial distribution $\Prob$ over $\mb{V}$. Next, given that the DAG $\G$ in question is Markov to $\Prob$, we can represent the joint multinomial distribution $\Prob$ by the Markov factorization:
\begin{align*}
    \Prob(\mb{V}) = \prod_{i \in \mb{v}} \Prob(X_i\,|\,\mb{X}_{\Pa(i, \G)}). \tag{\ref{eq:Markov_factorization}}
\end{align*}
Note that each $\Prob(X_i\,|\,\mb{X}_{\Pa(i, \G)})$ is determined by the set of parameters $\bs\theta_{i, \G}$. We express each $\bs\theta_{i, \G}$ in a tabular form (e.g., the three tables in \textbf{Example \ref{ex:adjF_not_paramM}}) and call them $\bs\theta$-tables. The table for $\bs\theta_{i, \G}$ has the dimension of $\mt{r}(\Pa(i, \G)) \times \mt{r}(i)$ where each row corresponds to the value assignment of $\mb{X}_{\Pa(i, \G)}$, and each column to the value of $X_i$. Each cell is the probability of $X_i$ taking the value specified by the column conditioned on the value assignment of $\mb{X}_{\Pa(i, \G)}$ specified by the row. Given that each row must sum to 1, cells in the last column are determined by the rest of the cells in the same row. For this reason, the cells in the last column are not parameters needed to compute $\Prob(X_i\,|\,\mb{X}_{\Pa(i, \G)})$ and thus are marked as {\color{gray}\text{gray}}.\\

%%%%%%%%%%%%%%%%%%%%%%%%%%%%%%%%%%%%%%%%%%%%%%%%%%%%%%%%%%%%%%%%%%%%%%
%%%%%%%%%%%%%%%%%%%%%%%%%%%%%%%%%%%%%%%%%%%%%%%%%%%%%%%%%%%%%%%%%%%%%%
\subsection{Example \ref{ex:adjF_not_paramM}}
%%%%%%%%%%%%%%%%%%%%%%%%%%%%%%%%%%%%%%%%%%%%%%%%%%%%%%%%%%%%%%%%%%%%%%
%%%%%%%%%%%%%%%%%%%%%%%%%%%%%%%%%%%%%%%%%%%%%%%%%%%%%%%%%%%%%%%%%%%%%%
\label{app:ex_adjF_not_paramM}
\begin{figure}[h]
    \centering
    \begin{tikzpicture}
    \node(V1) at (0,0) {$1$};
    \node(V2) at (1.5,1.5) {$2$};
    \node(V3) at (3,0) {$3$};
    \node(G) at (3.75, 1.5) {$\G_0$};
    \path[->, line width=0.5mm] (V1) edge (V2);
    \path[->, line width=0.5mm] (V2) edge (V3);
    \end{tikzpicture}
    \caption{$\G_0$ from Figure \ref{fig:adjF_not_uFr}}
    \label{fig:app:ex_adjF_not_paramM}
\end{figure}
\noindent Given that $\mb{V} = \{X_1, X_2, X_3\}$, consider $\mt{r}(1) = \mt{r}(3) = 2$ and $\mt{r}(2) = 3$. We want to construct a multinomial distribution $\Prob$ such that $\I(\Prob) = \{\la X_1, X_3\,|\,\{X_2\}\ra, \la X_1, X_3\ra\} = \I(\headcirc\Prob)$. First, we start with $\G_0$ in Figure \ref{fig:app:ex_adjF_not_paramM} where $\I(\G_0) = \{\la X_1, X_3\,|\,\{X_2\}\ra\}$. Below are the $\bs\theta$-tables for the three variables.\\
\vspace{0.1cm}

\begin{minipage}{.45\linewidth}
\begin{flushleft}
\begin{tabular}{|c|c|c|}
     \multicolumn{3}{c}{$\bs\theta_{1, \G_0}$}\\
     \hline
     $i$ & 0 & 1\\
     \hline
     $\Prob(X_1 = i)$ & $a = 0.5$ & \cellgray0.5 \\ 
     \hline
\end{tabular}\\
\vspace{0.2cm}
\begin{tabular}{|c|c|c|c|}
     \multicolumn{4}{c}{$\bs\theta_{2, \G_0}$}\\
     \hline
     $i$ & 0 & 1 & 2\\
     \hline
     $\Prob(X_2 = i\,|\,X_1=0)$ & $b_0 = 0.2$ & $c_0 = 0.2$ & 0\cellgray.6\\ 
     \hline
     $\Prob(X_2 = i\,|\,X_1=1)$ & $b_1 = 0.3$ & $c_1 = 0.3$ & \cellgray0.4\\ 
     \hline
\end{tabular}
\end{flushleft}
\end{minipage}
\begin{minipage}{.5\linewidth}
\centering
\begin{tabular}{|c|c|c|}
     \multicolumn{3}{c}{$\bs\theta_{3, \G_0}$}\\
     \hline
     $i$ & 0 & 1\\
     \hline
     $\Prob(X_3 = i\,|\,X_2=0)$ & $d_0=0.2$ & \cellgray0.8\\ 
     \hline
     $\Prob(X_3 = i\,|\,X_2=1)$ & $d_1=0.4$ &\cellgray 0.6\\ 
     \hline
     $\Prob(X_3 = i\,|\,X_2=2)$ & $d_2=0.3$ & \cellgray0.7\\ 
     \hline
\end{tabular}
\end{minipage}\\

\vspace{0.2cm}
\noindent To verify that $\la X_1, X_3\ra \in \I(\Prob)$, it suffices to show that $\Prob(X_3 = 0\,|\,X_1 = 0) = \Prob(X_3 = 0\,|\,X_1 = 1)$. Consider any $i \in \{0, 1\}$.

\begin{align*}
    \Prob(X_3 = 0\,|\,X_1 = i) =& \sum_{j=0}^2\,\Prob(X_3 = 0, X_2 = j\,|\,X_1 = i) = \sum_{j=0}^2\,\frac{\Prob(X_1=i, X_2=j, X_3=0)}{\Prob(X_1=i)}\\
    =& \sum_{j=0}^2\,\frac{\Prob(X_1=i)\,\Prob(X_2=j\,|\,X_1=i)\,\Prob(X_3 = 0\,|\,X_2 = j)}{\Prob(X_1 = i)}\\
    =& \sum_{j=0}^2\,\Prob(X_3=0\,|\,X_2=j)\,\Prob(X_2=j\,|\,X_1=i)\\
    =& \,d_0 b_i + d_1 c_i + d_2 (1- b_i - c_i)\\
    \Prob(X_3 = 0\,|\,X_1=0) =& (0.2 \times 0.2) + (0.4 \times 0.2) + (0.3 \times 0.6) = 0.3\\
    \Prob(X_3 = 0\,|\,X_1=1) =& (0.2 \times 0.3) + (0.4 \times 0.3) + (0.3 \times 0.4) = 0.3
\end{align*}
and hence $\la X_1, X_3\ra \in \I(\Prob)$.

On the other hand, we show arithmetically that there exists no CI in $\I(\Prob)$ except $\la X_1, X_3\ra$ and $\la X_1, X_3\,|\,\{X_2\}\ra$. By ignoring symmetric pairs, there are a total of 6 possible CIs in $\I(\mb{V})$. $\la X_1, X_3\ra, \la X_1, X_3\,|\,\{X_2\}\ra \in \I(\Prob)$ have been shown above. For $\la X_1, X_2\ra, \la X_2, X_3\ra \in \I(\mb{V})$, readers can verify that they are not in $\I(\Prob)$ simply by reading off the tables for $\bs\theta_{2, \G_0}$ and $\bs\theta_{3, \G_0}$ respectively. For $\la X_1, X_2\,|\,\{X_3\}\ra \in \I(\mb{V})$, it suffices to show that $\Prob(X_1=0\,|\,X_2=0, X_3=0) \neq \Prob(X_1=0\,|\,X_2=1, X_3=0)$. For any $j \in \{0, 1, 2\}$,

\begin{align*}
    \Prob(X_1 = 0\,|\,X_2=j, X_3=0) &= \frac{\Prob(X_1=0, X_2=j, X_3=0)}{\Prob(X_2=j, X_3=0)}\\
    &= \frac{\Prob(X_1=0)\,\Prob(X_2=j\,|\,X_1=0)\,\Prob(X_3=0\,|\,X_2=j)}{\Prob(X_2=j)\,\Prob(X_3=0\,|\,X_2=j)}\\
    &= \frac{\Prob(X_1=0)\,\Prob(X_2=j\,|\,X_1=0)}{\Prob(X_2=j)}\\
    &= \frac{\Prob(X_1=0)\,\Prob(X_2=j\,|\,X_1=0)}{\Prob(X_1=0, X_2=j) + \Prob(X_1=1, X_2=j)}\\
    &= \frac{\Prob(X_1=0)\,\Prob(X_2=j\,|\,X_1=0)}{\Prob(X_1=0)\,\Prob(X_2=j\,|\,X_1=0) + \Prob(X_1=1)\,\Prob(X_2=j\,|\,X_1 = 1)}\\
    \Prob(X_1=0\,|\,X_2=0, X_3=0) &= \frac{0.5 \times 0.2}{0.5 \times 0.2 + 0.5 \times 0.3} = 0.4\\
    \Prob(X_1=0\,|\,X_2=2, X_3=0) &= \frac{0.5 \times 0.6}{0.5 \times 0.6 + 0.5 \times 0.4} = 0.6
\end{align*}
and thus $\la X_1, X_2\,|\,\{X_3\}\ra \notin \I(\Prob)$. 

Lastly, for $\la X_2, X_3\,|\,\{X_1\} \ra$, consider any $k \in \{0, 1\}$. By making use of $\Prob(X_3=0\,|\,X_1=0)$ obtained above, we have

\begin{align*}
    \Prob(X_2 = 0\,|\,X_1=0, X_3=k) &= \frac{\Prob(X_1=0, X_2=0, X_3=k)}{\Prob(X_1=0, X_3=k)}\\
    &= \frac{\Prob(X_1=0)\,\Prob(X_2=0\,|\,X_1=0)\,\Prob(X_3=k\,|\,X_2=0)}{\Prob(X_1=0)\,\Prob(X_3=k\,|\,X_1=0)}\\
    &= \frac{\Prob(X_2=0\,|\,X_1=0)\,\Prob(X_3=k\,|\,X_2=0)}{\Prob(X_3=k\,|\,X_1=0)}\\
    \Prob(X_2 = 0\,|\,X_1=0, X_3=0) &= \frac{0.2 \times 0.2}{0.3} = 0.133\\
    \Prob(X_2 = 0\,|\,X_1=0, X_3=1) &= \frac{0.2 \times 0.8}{(1-0.3)} = 0.229
\end{align*}
and hence $\la X_2, X_3\,|\,\{X_1\}\ra \notin \I(\Prob)$.\hfill $\square$\\

%%%%%%%%%%%%%%%%%%%%%%%%%%%%%%%%%%%%%%%%%%%%%%%%%%%%%%%%%%%%%%%%%%%%%%
%%%%%%%%%%%%%%%%%%%%%%%%%%%%%%%%%%%%%%%%%%%%%%%%%%%%%%%%%%%%%%%%%%%%%%
\subsection{Example \ref{ex:ParamM_not_uParamM}}
%%%%%%%%%%%%%%%%%%%%%%%%%%%%%%%%%%%%%%%%%%%%%%%%%%%%%%%%%%%%%%%%%%%%%%
%%%%%%%%%%%%%%%%%%%%%%%%%%%%%%%%%%%%%%%%%%%%%%%%%%%%%%%%%%%%%%%%%%%%%%
\label{app:ex_ParamM_not_uParamM}
\begin{figure}[h]
    \centering
    \begin{tikzpicture}
    \node(V1) at (0,2) {$1$};
    \node(V2) at (2,2) {$2$};
    \node(V3) at (2,0) {$3$};
    \node(V4) at (0,0) {$4$};
    \node(G) at (3, 0) {$\G^*$};
    \path[->, line width=0.5mm] (V1) edge (V2);
    \path[->, line width=0.5mm] (V2) edge (V3);
    \path[->, line width=0.5mm] (V3) edge (V4);
    \path[->, line width=0.5mm] (V1) edge (V4);
    \end{tikzpicture}
    \caption{$\G^*$ from Figure \ref{fig:Fr_not_uFr}}
    \label{fig:app:ex_ParamM_not_uParamM}
\end{figure}

\noindent Given that $\mb{V} = \{X_1, X_2, X_3, X_4\}$, consider $\mt{r}(1) =  \mt{r}(2) =\mt{r}(3) = 2$, and $\mt{r}(4)=3$. We want to construct a multinomial distribution $\Prob$ such that $\I(\Prob) = \{\la X_1, X_3,\,|\,\{X_2\}\ra, \la X_2, X_4\,|\,\{X_1,X_3\}\ra, \la X_1, X_4\ra\}$. First, we start with $\G^*$ from Figure \ref{fig:app:ex_ParamM_not_uParamM} where $\I(\G^*) = \{\la X_1, X_3,\,|\,\{X_2\}\ra, \la X_2, X_4\,|\,\{X_1,X_3\}\ra\}$. Below are the $\bs\theta$-tables.\\

\begin{minipage}{.35\linewidth}
\begin{flushleft}
\begin{tabular}{|c|c|c|}
     \multicolumn{3}{c}{$\bs\theta_{1, \G^*}$}\\
     \hline
     $i$ & 0 & 1\\
     \hline
     $\Prob(X_1 = i)$ & $a=0.5$ & \cellgray $0.5$ \\ 
     \hline
\end{tabular}\\
\vspace{0.2cm}
\begin{tabular}{|c|c|c|}
     \multicolumn{3}{c}{$\bs\theta_{2, \G^*}$}\\
     \hline
     $i$ & 0 & 1\\
     \hline
     $\Prob(X_2 = i\,|\,X_1=0)$ & $b_0=0.6$ & \cellgray $0.4$\\ 
     \hline
     $\Prob(X_2 = i\,|\,X_1=1)$ & $b_1=0.1$ & \cellgray $0.9$\\ 
     \hline
\end{tabular}\\
\vspace{0.2cm}
\begin{tabular}{|c|c|c|}
     \multicolumn{3}{c}{$\bs\theta_{3, \G^*}$}\\
     \hline
     $i$ & 0 & 1\\
     \hline
     $\Prob(X_3 = i\,|\,X_2=0)$ & $c_0=0.7$ & \cellgray $0.3$\\ 
     \hline
     $\Prob(X_3 = i\,|\,X_2=1)$ & $c_1=0.2$ &\cellgray $0.8$\\ 
     \hline
\end{tabular}
\end{flushleft}
\end{minipage}
\begin{minipage}{.7\linewidth}
\centering
\begin{tabular}{|c|c|c|c|}
     \multicolumn{4}{c}{$\bs\theta_{4, \G^*}$}\\
     \hline
     $i$ & 0 & 1 & 2\\
     \hline
     $\Prob(X_4 = i\,|\,X_1=0, X_3=0)$ & $d_{00}=0.1$ & $e_{00}=0.1$ & \cellgray $0.8$\\ 
     \hline
     $\Prob(X_4 = i\,|\,X_1=0, X_3=1)$ & $d_{01}=0.05$ & $e_{01}=0.05$ & \cellgray $0.9$\\  
     \hline
     $\Prob(X_4 = i\,|\,X_1=1, X_3=0)$ & $d_{10}=0.1125$ & $e_{10}=0.1125$ & \cellgray $0.775$\\  
     \hline
     $\Prob(X_4 = i\,|\,X_1=1, X_3=1)$ & $d_{11}=0.0625$ & $e_{11}=0.0625$ & \cellgray $0.875$\\ 
     \hline
\end{tabular}
\end{minipage}\\
\,\\

\noindent To verify that $\la X_1, X_4\ra \in \I(\Prob)$, it suffices to show that $\Prob(X_4 = l\,|\,X_1 = 0) = \Prob(X_4 = l\,|\,X_1 = 1)$ for each $l \in \{0, 1\}$. Consider each $i \in \{0, 1\}$.

\begin{align*}
    \Prob(X_4 = 0\,|\,X_1=i) &= \sum_{k \in \{0,1\}} \Prob(X_3=k, X_4=0\,|\,X_1=i) = \sum_{k \in \{0, 1\}} \frac{\Prob(X_1=i, X_3=k, X_4=0)}{\Prob(X_1=i)}\\
    &= \sum_{k \in \{0, 1\}} \frac{\Prob(X_4=0\,|\,X_1=i, X_3=k)\,\Prob(X_1=i, X_3=k)}{\Prob(X_1=i)}\\
    &= \sum_{k \in \{0, 1\}} \frac{\Prob(X_4=0\,|\,X_1=i, X_3=k)\,\sum_{j \in \{0,1\}}\Prob(X_1=i, X_2=j, X_3=k)}{\Prob(X_1=i)}\\
    &= \sum_{k \in \{0, 1\}} \bigg(\frac{\Prob(X_4=0\,|\,X_1=i, X_3=k)}{\Prob(X_1=i)} \sum_{j \in \{0, 1\}} \Prob(X_1=i)\,\Prob(X_2=j\,|\,X_1=i)\,\Prob(X_3=k\,|\,X_2=j)\bigg)\\
    &= \sum_{k \in \{0, 1\}} \bigg(\Prob(X_4=0\,|\,X_1=i, X_3=k) \sum_{j \in \{0, 1\}} \Prob(X_2=j\,|\,X_1=i)\,\Prob(X_3=k\,|\,X_2=j)\bigg)\\
    &=d_{i0}(b_ic_0 + (1-b_i)c_1) + d_{i1}(b_i(1-c_0) + (1-b_i)(1-c_1))\\
    \Prob(X_4=0\,|\,X_1=0) &= \bigg(0.1 \times (0.6 \times 0.7 + 0.4 \times 0.2)\bigg) + \bigg(0.05 \times (0.6 \times 0.3 + 0.4 \times 0.8)\bigg) = 0.075\\
    \Prob(X_4=0\,|\,X_1=1) &= \bigg(0.1125 \times (0.1 \times 0.7 + 0.9 \times 0.2)\bigg) + \bigg(0.0625 \times (0.1 \times 0.3 + 0.9 \times 0.8)\bigg) = 0.075
\end{align*}
The case for $\Prob(X_4=1\,|\,X_1=0) = \Prob(X_4=1\,|\,X_1=1)$ is exactly the same due to the two identical columns in the table for $\bs\theta_{4, \G^*}$. Hence, $\la X_1, X_4\ra \in \I(\Prob)$. We leave the step of checking no extra CI to readers.\hfill $\square$

%%%%%%%%%%%%%%%%%%%%%%%%%%%%%%%%%%%%%%%%%%%%%%%%%%%%%%%%%%%%%%%%%%%%%%
%%%%%%%%%%%%%%%%%%%%%%%%%%%%%%%%%%%%%%%%%%%%%%%%%%%%%%%%%%%%%%%%%%%%%%
\subsection{Example \ref{ex:uParam_not_oriF}}
%%%%%%%%%%%%%%%%%%%%%%%%%%%%%%%%%%%%%%%%%%%%%%%%%%%%%%%%%%%%%%%%%%%%%%
%%%%%%%%%%%%%%%%%%%%%%%%%%%%%%%%%%%%%%%%%%%%%%%%%%%%%%%%%%%%%%%%%%%%%%
\label{app:ex_uParam_not_oriF}
\begin{figure}[h]
    \centering
    \subfloat{
    \begin{tikzpicture}
    \node(V1) at (0,2) {$1$};
    \node(V2) at (2,2) {$2$};
    \node(V3) at (2,0) {$4$};
    \node(V4) at (0,0) {$3$};
    \node(G) at (3, 0) {$\G^*$};
    \path[->, line width=0.5mm] (V1) edge (V2);
    \path[->, line width=0.5mm] (V2) edge (V3);
    \path[->, line width=0.5mm] (V4) edge (V3);
    \path[->, line width=0.5mm] (V1) edge (V4);
    \end{tikzpicture}}
    \caption{$\G^*$ from Figure \ref{fig:uFr_not_oriF}}
    \label{fig:app:ex_uParam_not_oriF}
\end{figure}

\noindent Consider $\mb{V} = \{X_1, X_2, X_3, X_4\}$ where all variables are binary. We want to construct a multinomial distribution $\Prob$ such that $\I(\Prob) = \{\la X_1, X_4\,|\,\{X_2, X_3\}\ra, \la X_2, X_3\,|\,\{X_1\}\ra, \la X_1, X_4\ra\}$. First, we start with $\G^*$ in Figure \ref{fig:app:ex_uParam_not_oriF} where $\I(\G^*) = \{ \la X_1, X_4\,|\,\{X_2, X_3\}\ra, \la X_2, X_3\,|\,\{X_1\}\ra\}$. Below are the four $\bs\theta$-tables.\\

\begin{minipage}{.4\linewidth}
\begin{flushleft}
\begin{tabular}{|c|c|c|}
     \multicolumn{3}{c}{$\bs\theta_{1, \G^*}$}\\
     \hline
     $i$ & 0 & 1\\
     \hline
     $\Prob(X_1 = i)$ & $a = 0.5$ & \cellgray $0.5$ \\ 
     \hline
\end{tabular}\\
\vspace{0.2cm}
\begin{tabular}{|c|c|c|}
     \multicolumn{3}{c}{$\bs\theta_{3, \G^*}$}\\
     \hline
     $i$ & 0 & 1\\
     \hline
     $\Prob(X_3 = i\,|\,X_1=0)$ & $c_0 = 0.4$ & \cellgray $0.6$\\ 
     \hline
     $\Prob(X_3 = i\,|\,X_1=1)$ & $c_1 = 0.3$ &\cellgray $0.7$\\ 
     \hline
\end{tabular}
\end{flushleft}
\end{minipage}
\begin{minipage}{.6\linewidth}
\begin{flushleft}
\begin{tabular}{|c|c|c|}
     \multicolumn{3}{c}{$\bs\theta_{2, \G^*}$}\\
     \hline
     $i$ & 0 & 1\\
     \hline
     $\Prob(X_2 = i\,|\,X_1=0)$ & $b_0=0.3$ & \cellgray $0.7$\\ 
     \hline
     $\Prob(X_2 = i\,|\,X_1=1)$ & $b_1=0.4$ & \cellgray $0.6$\\ 
     \hline
\end{tabular}\\
\vspace{0.2cm}
\begin{tabular}{|c|c|c|}
     \multicolumn{3}{c}{$\bs\theta_{4, \G^*}$}\\
     \hline
     $i$ & 0 & 1\\
     \hline
     $\Prob(X_4 = i\,|\,X_2=0, X_3=0)$ & $d_{00}=0.1$  & \cellgray $0.9$\\ 
     \hline
     $\Prob(X_4 = i\,|\,X_2=0, X_3=1)$ & $d_{01}=0.7$ & \cellgray $0.3$\\  
     \hline
     $\Prob(X_4 = i\,|\,X_2=1, X_3=0)$ & $d_{10}=0.7$ & \cellgray $0.3$\\  
     \hline
     $\Prob(X_4 = i\,|\,X_2=1, X_3=1)$ & $d_{11}=0.8$ & \cellgray $0.2$\\ 
     \hline
\end{tabular}
\end{flushleft}
\end{minipage}\\
\vspace{0.2cm}

\noindent To verify that $\la X_1, X_4\ra \in \I(\Prob)$, it suffices to show that $\Prob(X_4 = 0\,|\,X_1 = 0) = \Prob(X_4 = 0\,|\,X_1 = 1)$. Consider any $i \in \{0, 1\}$.
\begin{align*}
    \Prob(X_4 = 0\,|\,X_1=i) &= \sum_{j \in \{0,1\}} \Prob(X_2=j, X_4=0\,|\,X_1=i)\\ 
    &= \sum_{j \in \{0, 1\}}\sum_{k \in \{0, 1\}} \Prob(X_2=j, X_3=k, X_4=0\,|\,X_1=i)\\
    &= \sum_{j \in \{0, 1\}} \sum_{k \in \{0, 1\}} \frac{\Prob(X_1=i, X_2=j, X_3=k, X_4=0)}{\Prob(X_1=i)}\\
    &= \sum_{j \in \{0, 1\}} \sum_{k \in \{0, 1\}} \frac{\Prob(X_1=i)\,\Prob(X_2=j\,|\,X_1=i)\,\Prob(X_3=k\,|\,X_1=i)\,\Prob(X_4=0\,|\,X_2=j, X_3=k)}{\Prob(X_1=i)}\\
    &= \sum_{j \in \{0, 1\}} \Prob(X_2=j\,|\,X_1=i) \sum_{k \in \{0, 1\}} \,\Prob(X_3=k\,|\,X_1=i)\,\Prob(X_4=0\,|\,X_2=j, X_3=k)\\
    &= b_i(c_i d_{00} + (1-c_i)d_{01}) + (1-b_i)(c_i d_{10} + (1-c_i) d_{11})\\
    &= b_i(c_i(d_{00}-d_{01}) + d_{01}) + (1-b_i)(c_i(d_{10}-d_{11}) + d_{11})\\
    \Prob(X_4=0\,|\,X_1=0) &= 0.3\times(0.4\times(0.1-0.7) + 0.7) + (1-0.3)\times(0.4\times(0.7-0.8) + 0.8) = 0.67\\
    \Prob(X_4=0\,|\,X_1=1) &= 0.4\times(0.3\times (0.1-0.7) + 0.7) + (1-0.4)\times(0.3\times(0.7-0.8) + 0.8) = 0.67
\end{align*}
and hence $\la X_1, X_4\ra \in \I(\Prob)$. We leave the step of checking no extra CI to readers.\hfill $\square$

%%%%%%%%%%%%%%%%%%%%%%%%%%%%%%%%%%%%%%%%%%%%%%%%%%%%%%%%%%%%%%%%%%%%%%
%%%%%%%%%%%%%%%%%%%%%%%%%%%%%%%%%%%%%%%%%%%%%%%%%%%%%%%%%%%%%%%%%%%%%%
\subsection{Example \ref{ex:uFr_not_uParamM}}
%%%%%%%%%%%%%%%%%%%%%%%%%%%%%%%%%%%%%%%%%%%%%%%%%%%%%%%%%%%%%%%%%%%%%%
%%%%%%%%%%%%%%%%%%%%%%%%%%%%%%%%%%%%%%%%%%%%%%%%%%%%%%%%%%%%%%%%%%%%%%
\label{app:ex_uFr_not_uParamM}
\begin{figure}[htbp]
    \centering
    \begin{tikzpicture}
    \node (X1) at (0.0,1.0) {$1$};
    \node (X2) at (0.9510565162951535,0.30901699437494745) {$2$};
    \node (X3) at (0.5877852522924732,-0.8090169943749473) {$3$};
    \node (X4) at (-0.587785252292473,-0.8090169943749476) {$4$};
    \node (X5) at (-0.9510565162951536,0.30901699437494723) {$5$};
    \node (label) at (2, -0.4) {$\G_0$};
    \path [->,line width=0.4mm] (X1) edge (X5);
    \path [->,line width=0.4mm] (X2) edge (X5);
    \path [->,line width=0.4mm] (X3) edge (X5);
    \path [->,line width=0.4mm] (X4) edge (X5);
    \path [->,line width=0.4mm] (X1) edge (X4);
    \end{tikzpicture}
    \caption{$\G_0$ from \textsc{Figure} \ref{fig:uFr_not_triF}}
    \label{fig:app:ex:uParam_not_triF}
\end{figure}

\noindent Consider that $\mb{V} = \{X_1, X_2, X_3, X_4, X_5\}$ where the first four variables are binary and the last is ternary. From $\G_0$ in \textsc{Figure} \ref{fig:app:ex:uParam_not_triF}, below shows all the 20 CIs in $\I(\G_0)$.
\begin{align*}
    \I(\G_0) = &
\begin{Bmatrix}
    \la X_1, X_2\ra, &
    \la X_1, X_3\ra, & 
    \la X_2, X_3\ra, \\
    \la X_2, X_4\ra, &
    \la X_3, X_4\ra, &
    \la X_1, X_2\,|\,\{X_3\}\ra, \\
    \la X_1, X_2\,|\,\{X_4\}\ra, &
    \la X_1, X_3\,|\,\{X_2\}\ra, &
    \la X_2, X_3\,|\,\{X_1\}\ra, \\
    \la X_1, X_3\,|\,\{X_4\}\ra, &
    \la X_2, X_3\,|\,\{X_4\}\ra, &
    \la X_2, X_4\,|\,\{X_1\}\ra, \\ 
    \la X_2, X_4\,|\,\{X_3\}\ra, &
    \la X_3, X_4\,|\,\{X_1\}\ra, &
    \la X_3, X_4\,|\,\{X_2\}\ra, \\
    \la X_1, X_2\,|\,\{X_3, X_4\}\ra, &
    \la X_1, X_3\,|\,\{X_2, X_4\}\ra, & 
    \la X_2, X_3\,|\,\{X_1, X_4\}\ra, \\
    \la X_2, X_4\,|\,\{X_1, X_3\}\ra, &
    \la X_3, X_4\,|\,\{X_1, X_2\}\ra
\end{Bmatrix}
\end{align*}
Below is the set of unfaithful CIs.
\begin{align*}
    \Psi_{\G_0, \mathring\Prob} = &
\begin{Bmatrix}
    \psi_1: \la X_1, X_5\ra, &
    \psi_2: \la X_1, X_2\,|\,\{X_5\}\ra, \\ 
    \psi_3: \la X_1, X_3\,|\,\{X_5\}\ra, &
    \psi_4: \la X_1, X_5\,|\,\{X_2\}\ra, \\
    \psi_5: \la X_1, X_5\,|\,\{X_3\}\ra, &
    \psi_6: \la X_1, X_2\,|\,\{X_3, X_5\}\ra, \\
    \psi_7: \la X_1, X_3\,|\,\{X_2, X_5\}\ra, &
    \psi_8: \la X_1, X_5\,|\,\{X_2, X_3\}\ra
\end{Bmatrix}.
\end{align*}

Unlike the previous examples where we only need to induce one CI parametrically, we need to ensure that the eight CIs in $\Psi_{\G_0, \mathring\Prob}$ are satisfied. Here are the five $\bs\theta$-tables.\\ 

\begin{minipage}{.4\linewidth}
\begin{flushleft}
\begin{tabular}{|c|c|c|}
     \multicolumn{3}{c}{$\bs\theta_{1, \G_0}$}\\
     \hline
     $i$ & 0 & 1\\
     \hline
     $\Prob(X_1 = i)$ & $a = 0.5$ & \cellgray $0.5$ \\ 
     \hline
\end{tabular}\\
\vspace{0.2cm}
\begin{tabular}{|c|c|c|}
     \multicolumn{3}{c}{$\bs\theta_{3, \G_0}$}\\
     \hline
     $i$ & 0 & 1\\
     \hline
     $\Prob(X_3 = i)$ & $c=0.5$ & \cellgray $0.5$\\ 
     \hline
\end{tabular}
\end{flushleft}
\end{minipage}
\begin{minipage}{.5\linewidth}
\begin{flushleft}
\begin{tabular}{|c|c|c|}
     \multicolumn{3}{c}{$\bs\theta_{2, \G_0}$}\\
     \hline
     $i$ & 0 & 1\\
     \hline
     $\Prob(X_2 = i)$ & $b=0.5$ & \cellgray $0.5$\\ 
     \hline
\end{tabular}\\
\vspace{0.2cm}
\begin{tabular}{|c|c|c|}
     \multicolumn{3}{c}{$\bs\theta_{4, \G_0}$}\\
     \hline
     $i$ & 0 & 1\\
     \hline
     $\Prob(X_4 = i\,|\,X_1=0)$ & $d_0=0.4$ & \cellgray $0.6$\\ 
     \hline
     $\Prob(X_4 = i\,|\,X_1=1)$ & $d_1=0.8$ & \cellgray $0.2$\\ 
     \hline
\end{tabular}
\end{flushleft}
\end{minipage}

\vspace{0.2cm}
\begin{tabular}{|c|c|c|c|}
\multicolumn{4}{c}{$\bs\theta_{5, \G_0}$}\\
\hline
$i$ & 0 & 1 & 2\\
\hline
$\Prob(X_5 = i\,|\,X_1=0, X_2=0, X_3=0, X_4=0)$ & $e_{0000}=0.4$ & $f_{0000}=0.4$ & \cellgray $0.2$\\ 
\hline
$\Prob(X_5 = i\,|\,X_1=0, X_2=0, X_3=0, X_4=1)$ & $e_{0001}=0.15$ & $f_{0001}=0.15$ & \cellgray $0.7$\\
\hline
$\Prob(X_5 = i\,|\,X_1=0, X_2=0, X_3=1, X_4=0)$ & $e_{0010}=0.2$ & $f_{0010}=0.2$ & \cellgray $0.6$\\  
\hline
$\Prob(X_5 = i\,|\,X_1=0, X_2=0, X_3=1, X_4=1)$ & $e_{0011}=0.2$ & $f_{0011}=0.2$ & \cellgray $0.6$\\ 
\hline
$\Prob(X_5 = i\,|\,X_1=0, X_2=1, X_3=0, X_4=0)$ & $e_{0100}=0.3$ & $f_{0100}=0.3$ & \cellgray $0.4$\\ 
\hline
$\Prob(X_5 = i\,|\,X_1=0, X_2=1, X_3=0, X_4=1)$ & $e_{0101}=0.08$ & $f_{0101}=0.08$ & \cellgray $0.84$\\
\hline
$\Prob(X_5 = i\,|\,X_1=0, X_2=1, X_3=1, X_4=0)$ & $e_{0110}=0.1$ & $f_{0110}=0.1$ & \cellgray $0.8$\\  
\hline
$\Prob(X_5 = i\,|\,X_1=0, X_2=1, X_3=1, X_4=1)$ & $e_{0111}=0.3$ & $f_{0111}=0.3$ & \cellgray $0.4$\\ 
\hline
$\Prob(X_5 = i\,|\,X_1=1, X_2=0, X_3=0, X_4=0)$ & $e_{1000}=0.2$ & $f_{1000}=0.2$ & \cellgray $0.6$\\ 
\hline
$\Prob(X_5 = i\,|\,X_1=1, X_2=0, X_3=0, X_4=1)$ & $e_{1001}=0.45$ & $f_{1001}=0.45$ & \cellgray $0.1$\\
\hline
$\Prob(X_5 = i\,|\,X_1=1, X_2=0, X_3=1, X_4=0)$ & $e_{1010}=0.15$ & $f_{1010}=0.15$ & \cellgray $0.7$\\
\hline
$\Prob(X_5 = i\,|\,X_1=1, X_2=0, X_3=1, X_4=1)$ & $e_{1011}=0.4$ & $f_{1011}=0.4$ & \cellgray $0.2$\\ 
\hline
$\Prob(X_5 = i\,|\,X_1=1, X_2=1, X_3=0, X_4=0)$ & $e_{1100}=0.15$ & $f_{1100}=0.15$ & \cellgray $0.7$\\
\hline
$\Prob(X_5 = i\,|\,X_1=1, X_2=1, X_3=0, X_4=1)$ & $e_{1101}=0.24$ & $f_{1101}=0.24$ & \cellgray $0.52$\\
\hline
$\Prob(X_5 = i\,|\,X_1=1, X_2=1, X_3=1, X_4=0)$ & $e_{1110}=0.225$ & $f_{1110}=0.225$ & \cellgray $0.55$\\  
\hline
$\Prob(X_5 = i\,|\,X_1=1, X_2=1, X_3=1, X_4=1)$ & $e_{1111}=0.2$ & $f_{1111}=0.2$ & \cellgray $0.6$\\ 
\hline
\end{tabular}\\
\,\\

\noindent Consider $\psi_8 = \la X_1, X_5\,|\,\{X_2, X_3\}\ra$ in particular. We first want to show that, if $\psi_8 \in \I(\Prob)$, then all other seven unfaithful CIs in $\Psi_{\G_0, \headcirc\Prob}$ are also in $\I(\Prob)$. To do so, observe that $\Prob$ is a strictly positive multinomial distribution as every cell in the five $\bs\theta$-tables is positive. Thus, $\Prob$ is a graphoid. Now we show that each of $\psi_1,...,\psi_7$ can be derived from $\I(\G_0) \cup \{\psi_8\}$ using the graphoid axioms discussed in \textbf{Appendix \ref{app:graphoid}} including intersection. In particular, denote $\phi_1 = \la X_1, X_2\ra$, $\phi_2 = \la X_1, X_2\,|\,\{X_3\}\ra$, and $\phi_3 = \la X_1, X_3\,|\,\{X_2\}\ra$ where $\phi_1, \phi_2, \phi_3 \in \I(\G_0) \subset \I(\Prob)$.\\

\begin{align}
    \underbrace{(X_1 \CI_\Prob X_2\,|\,\{X_3\})}_{\phi_2}\,\wedge\,\underbrace{(X_1 \CI_\Prob X_5\,|\,\{X_2, X_3\})}_{\psi_8} &\Rightarrow \underbrace{X_1 \CI_\Prob \{X_2, X_5\}\,|\,\{X_3\}}_{\text{(i)}} & \text{contraction}\\
    \underbrace{(X_1 \CI_\Prob X_2\,|\,\{X_3\})}_{\phi_2}\,\wedge\,\underbrace{(X_1 \CI_\Prob X_5\,|\,\{X_2, X_3\})}_{\psi_8} &\Rightarrow \underbrace{X_1 \CI_\Prob \{X_2, X_5\}\,|\,\{X_3\}}_{\text{(i)}} & \text{contraction}\\
    &\Rightarrow \underbrace{X_1 \CI_\Prob X_2\,|\,\{X_3, X_5\}}_{\psi_6} & \text{(i), weak union}\\
    &\Rightarrow \underbrace{X_1 \CI_\Prob X_5\,|\,\{X_3\}}_{\psi_5} & \text{(i), decomposition}\\
    \underbrace{(X_1 \CI_\Prob X_3\,|\,\{X_2\})}_{\phi_3}\,\wedge\,\underbrace{(X_1 \CI_\Prob X_5\,|\,\{X_2, X_3\})}_{\psi_8} &\Rightarrow \underbrace{X_1 \CI_\Prob \{X_3, X_5\}\,|\,\{X_2\}}_{\text{(ii)}} & \text{contraction}\\
    &\Rightarrow \underbrace{X_1 \CI_\Prob X_3\,|\,\{X_2, X_5\}}_{\psi_7} & \text{(ii), weak union}\\
    &\Rightarrow \underbrace{X_1 \CI_\Prob X_5\,|\,\{X_2\}}_{\psi_4} & \text{(ii), decomposition}\\
    \underbrace{X_1 \CI_\Prob X_2\,|\,\{X_3, X_5\}}_{\psi_6} \wedge \underbrace{X_1 \CI_\Prob X_3\,|\,\{X_2, X_5\}}_{\psi_7} &\Rightarrow \underbrace{X_1 \CI_\Prob \{X_2, X_3\}\,|\,\{X_5\}}_{\text{(iii)}} & \text{intersection}\\
    &\Rightarrow \underbrace{X_1 \CI_\Prob X_2\,|\,\{X_5\}}_{\psi_2} & \text{(iii), decomposition}\\
    &\Rightarrow \underbrace{X_1 \CI_\Prob X_3\,|\,\{X_5\}}_{\psi_3} & \text{(iii), decomposition}\\
    \underbrace{(X_1 \CI_\Prob X_2)}_{\phi_1} \wedge \underbrace{(X_1, X_5\,|\,\{X_2\})}_{\psi_4} &\Rightarrow \underbrace{X_1 \CI_\Prob \{X_2, X_5\}}_{\text{(iv)}} & \text{contraction}\\
    &\Rightarrow \underbrace{X_1 \CI_\Prob X_5}_{\psi_1} & \text{(iv), decomposition}
\end{align}

Thus, it suffices to prove that $\psi_8 \in \I(\Prob)$, that is, $\Prob(X_5=0\,|\,X_1=0, X_2=j, X_3=k) = \Prob(X_5=0\,|\,X_1=1, X_2=j, X_3=k)$ for every $j, k \in \{0, 1\}$. Consider any $i, j, k \in \{0, 1\}$.
\begin{align*}
&\,\,\,\,\,\,\,\,\,\,\Prob(X_5=0\,|\,X_1=i, X_2=j, X_3=k)\\
&=\sum_{l \in \{0,1\}} \Prob(X_4 = l, X_5=0\,|\,X_1=i, X_2=j, X_3=k)\\
&=\sum_{l \in \{0,1\}} \frac{\Prob(X_1=i)\,\Prob(X_2=j)\,\Prob(X_3=k)\,\Prob(X_4=l\,|\,X_1=i)\,\Prob(X_5=0\,|\,X_1=i, X_2=j, X_3=k, X_4=l)}{\Prob(X_1=i)\,\Prob(X_2=j)\,\Prob(X_3=k)}\\
&=\sum_{l \in \{0,1\}}\Prob(X_4=l\,|\,X_1=i)\,\Prob(X_5=0\,|\,X_1=i, X_2=j, X_3=k, X_4=l)\\
&=d_{i}e_{ijk0} + (1-d_{i})e_{ijk1}
\end{align*}

Thus, we have:

\begin{align*}
\Prob(X_5=0\,|\,X_1=0, X_2=j, X_3=k) &= 0.4e_{0jk0} + 0.6e_{0jk1}\\
\Prob(X_5=0\,|\,X_1=1, X_2=j, X_3=k) &= 0.8e_{1jk0} + 0.2e_{1jk1}    
\end{align*}

% for every $j, k \in \{0, 1\}$. Now we show that $\Prob(X_5=0\,|\,X_1=0, X_2=j, X_3=k) = \Prob(X_5=0\,|\,X_1=1, X_2=j, X_3=k)$ for every $j, k \in \{0, 1\}$.

\begin{align*}
    \Prob(X_5=0\,|\,X_1=0, X_2=0, X_3=0) &= 0.4 \times 0.4 + 0.6 \times 0.15 = 0.25\\
    \Prob(X_5=0\,|\,X_1=1, X_2=0, X_3=0) &= 0.8 \times 0.2 + 0.2 \times 0.45 = 0.25\\
    \Prob(X_5=0\,|\,X_1=0, X_2=0, X_3=1) &= 0.4 \times 0.2 + 0.6 \times 0.2 = 0.2\\
    \Prob(X_5=0\,|\,X_1=1, X_2=0, X_3=1) &= 0.8 \times 0.15 + 0.2 \times 0.4 = 0.2\\
    \Prob(X_5=0\,|\,X_1=0, X_2=1, X_3=0) &= 0.4 \times 0.3 + 0.6 \times 0.08 = 0.168\\
    \Prob(X_5=0\,|\,X_1=1, X_2=1, X_3=0) &= 0.8 \times 0.15 + 0.2 \times 0.24 = 0.168\\
    \Prob(X_5=0\,|\,X_1=0, X_2=1, X_3=1) &= 0.4 \times 0.1 + 0.6 \times 0.3 = 0.22\\
    \Prob(X_5=0\,|\,X_1=1, X_2=1, X_3=1) &= 0.8 \times 0.225 + 0.2 \times 0.2 = 0.22
\end{align*}

The case for $\Prob(X_5=1\,|\,X_1=0, X_2=j, X_3=k) = \Prob(X_5=1\,|\,X_1=1, X_2=j, X_3=k)$ for every $j, k\in \{0, 1\}$ is exactly the same due to the two identical columns in the table of $\bs\theta_{5, \G_0}$. Hence, we have $\psi_8 = \la X_1, X_5\,|\,\{X_2, X_3\}\ra \in \I(\Prob)$.

% \noindent Hence, $\Psi_{\G_0} \cup \I(\G^*) \subseteq \I(\Prob)$. We leave the step of checking no extra CI to readers. \hfill $\square$
%%%%%%%%%%%%%%%%%%%%%%%%%%%%%%%%%%%%%%%%%%%%%%%%%%%%%%%%%%%%%%%%%%%%%%
% \end{group}

\end{document}